\documentclass[11pt]{article} % For LaTeX2e
\usepackage{url}
\usepackage{smile}
\usepackage{graphicx} % more modern
\usepackage{algorithm}
\usepackage{algorithmic}
\usepackage{todonotes}
\usepackage{epstopdf}
\usepackage{wrapfig}
\usepackage[colorlinks, linkcolor=blue, anchorcolor=blue, citecolor=blue]{hyperref}
\usepackage[margin=1in]{geometry}
\usepackage[normalem]{ulem}
\usepackage[export]{adjustbox}% http://ctan.org/pkg/adjustbox
\usepackage{mathtools, cuted}
\usepackage{natbib}
\usepackage{enumerate}
\usepackage{microtype}
\usepackage{graphicx}
\usepackage{subcaption}
\usepackage{booktabs} % for professional tables
\usepackage{wrapfig}
\usepackage{amsmath,amsfonts,bm}

\def\eqref#1{equation~\ref{#1}}
\def\1{\bm{1}}

\DeclareMathAlphabet{\mathsfit}{\encodingdefault}{\sfdefault}{m}{sl}
\SetMathAlphabet{\mathsfit}{bold}{\encodingdefault}{\sfdefault}{bx}{n}

\DeclareMathOperator*{\argmax}{arg\,max}
\DeclareMathOperator*{\argmin}{arg\,min}

\usepackage{kpfonts}
\DeclareMathAlphabet{\mathsf}{OT1}{cmss}{m}{n}
\SetMathAlphabet{\mathsf}{bold}{OT1}{cmss}{bx}{n}

\providecommand{\norm}[1]{\|#1\|}

\begin{document}

\title{\huge A Hypergradient Approach to Robust Regression without Correspondence}

\author{Yujia Xie\thanks{Equal contributions.}, Yixiu Mao\footnotemark[1], Simiao Zuo, Hongteng Xu,\\ Xiaojing Ye, Tuo Zhao, Hongyuan Zha\thanks{Yujia Xie, Simiao Zuo, Tuo Zhao are affiliated with Georgia Institute of Technology. Emails: {\{\tt xieyujia, simiaozuo, tourzhao\}@gatech.edu.} Yixiu Mao is affiliated with Shanghai Jiao Tong University. Email: {\tt 956986044myx@gmail.com}. Hongteng Xu is affiliated with Duke University. Email: {\tt  hongteng.xu@duke.edu}. Xiaojing Ye is affiliated with Georgia State University. Email: {\tt  xye@gsu.edu}. Hongyuan Zha is affiliated with School of Data Science at Shenzhen Research Institute of Big Data and Shenzhen Institute of Artificial Intelligence and Robotics for Society. On leave from Georgia Institute of Technology. Email: {\tt  zhahy@cuhk.edu.cn}. }}
\date{}

\maketitle

%!TEX root = ../robot_tech.tex

\begin{abstract}
We consider a variant of regression problem, where the correspondence between input and output data is not available. Such shuffled data is commonly observed in many real world problems. Taking flow cytometry as an example, the measuring instruments may not be able to maintain the correspondence between the samples and the measurements. Due to the combinatorial nature of the problem, most existing methods are only applicable when the sample size is small, and limited to linear regression models. To overcome such bottlenecks, we propose a new computational framework --- ROBOT --- for the shuffled regression problem, which is applicable to large data and complex nonlinear models. Specifically, we reformulate the regression without correspondence as a continuous optimization problem. Then by exploiting the interaction between the regression model and the data correspondence, we develop a hypergradient approach based on differentiable programming techniques. Such a hypergradient approach essentially views the data correspondence as an operator of the regression, and therefore allows us to find a better descent direction for the model parameter by differentiating through the data correspondence. ROBOT can be further extended to the inexact correspondence setting, where there may not be an exact alignment between the input and output data. Thorough numerical experiments show that ROBOT achieves better performance than existing methods in both linear and nonlinear regression tasks, including real-world applications such as {\it flow cytometry} and {\it multi-object tracking}.  
\end{abstract}

% , where correspondence is parameterized as the optimal solution to an entropic optimal transport problem. 
% Such a formulation enables end-to-end training with efficient first order optimization algorithms, such that our framework can handle large-scale data and complex models. 
% Moreover, it naturally takes into account the interaction between the correspondence and the regression model in the training, leading to an effective optimization.

% For example, datasets collected from multiple platforms are anonymized so that they cannot be matched due to privacy constraints. with the same participators; the participator identities of the platforms are typically not accessible due to logistics or privacy issues.
%!TEX root = ../robot_tech.tex
% \vspace{-0.15in}
\section{Introduction}
%\vspace{-0.15in}

Regression analysis has been widely used in various machine learning applications to infer the  
the relationship between an explanatory random variable (i.e., the input) $X\in \mathbb{R}^d$ and a response random variable (i.e., the output) $Y\in \mathbb{R}^o$ \citep{stanton2001galton}. 
In the classical setting, regression is used on labeled datasets that contain paired samples $\{x_i, y_i\}_{i=1}^n$, where $x_i$, $y_i$ are realizations of $X$, $Y$, respectively. 

Unfortunately, such an input-output correspondence is not always available in some applications. One example is flow cytometry, which is a physical experiment for measuring properties of cells, e.g., affinity to a particular target \citep{abid2018stochastic}. Through this process, cells are suspended in a fluid and injected into the flow cytometer, where measurements are taken using the scattering of a laser. However, the instruments are unable to differentiate the cells passing through the laser, such that the correspondence between the cell proprieties (i.e., the measurements) and the cells is unknown. This prevents us from analyzing the relationship between the instruments and the measurements using classical regression analysis, due to the missing correspondence. Another example is multi-object tracking, where we need to infer the motion of objects given consecutive frames in a video. This requires us to find the correspondence between the objects in the current frame and those in the next frame.

The two examples above can be formulated as a shuffled regression problem. Specifically, we consider a multivariate regression model
\begin{align*}
Y = f\left(X, Z; w\right)+\varepsilon,
\end{align*}
where $X\in\RR^d,~Z\in\RR^{e}$ are two input vectors, $Y\in\RR^{o}$ is an output vector, $f:\RR^{d+e}\rightarrow\RR^{o}$ is the unknown regression model with parameters $w$ and $\varepsilon$ is the random noise independent on $X$ and $Z$. When we sample realizations from such a regression model, the correspondence between $(X,Y)$ and $Z$ is not available. Accordingly, we collect two datasets $\cD_1=\{x_i, y_i\}_{i=1}^n$ and $\cD_2=\{z_j\}_{j=1}^n$, and there exists a permutation $\pi^*$ such that $(x_i, z_{\pi(i)})$ corresponds to $y_i$ in the regression model. Our goal is to recover the unknown model parameter $w$. Existing literature also refer to the shuffled regression problem as \textit{unlabeled sensing}, \textit{homomorphic sensing}, and \textit{regression with an unknown permutation} \citep{unnikrishnan2018unlabeled}.
Throughout the rest of the paper, we refer to it as \textit{Regression WithOut Correspondence} (RWOC).

A natural choice of the objective for RWOC is to minimize the sum of squared residuals with respect to the regression model parameter $w$ up to the permutation $\pi(\cdot)$ over the training data, i.e.,
\begin{align}\label{eq:ls}
 \min_{w, \pi} \cL(w, \pi) =  \sum_{i=1}^n \norm{y_i-f \left(x
   _i, z_{\pi(i)}; w \right)}_2^2.
\end{align}
Existing works on RWOC mostly focus on theoretical properties of the global optima to \eqref{eq:ls} for estimating $w$ and $\pi$ \citep{pananjady2016linear, pananjady2017linear, abid2017linear, Elhami2017unlabeled, hsu2017linear, unnikrishnan2018unlabeled, pmlr-v97-tsakiris19a, zhang2020optimal}. The development of practical algorithms, however, falls far behind from the following three aspects:

\noindent $\bullet$ Most of the works are only applicable to linear regression models.

\noindent $\bullet$ Some of the existing algorithms are of very high computational complexity, and can only handle small number of data points in low dimensions \citep{Elhami2017unlabeled,pananjady2017denoising, tsakiris2018algebraic, peng2020linear}. Other algorithms choose to optimize with respect to $w$ and $\pi$ in an alteranting manner, e.g., alternating minimization in \citet{abid2017linear}. However, as there exists a strong interaction between $w$ and $\pi$, the optimization landscape of \eqref{eq:ls} is ill-conditioned. Therefore, these algorithms are not effective and often get stuck in local optima.
%fail to consider the interaction between the two parameters $w$ and $\pi(\cdot)$ (see Section \ref{sec:method}). For example, \citet{abid2017linear} considers the two parameters separately and applies alternating minimization with respect to them. Due to the strong interaction between $w$ and $\pi(\cdot)$, the optimization landscape is usually ill-conditioned. Therefore, these algorithms are not effective, e.g., they get stuck in local optima easily.

% Existing algorithms often suffer from significant suboptimality due to nonconvexity. For example, \citet{abid2017linear} applies alternating minimization with respect to the two parameters $w$ and $\pi(\cdot)$. However, such an algorithm does not consider the interaction of the two parameters, and therefore get stuck in local optima easily.

\noindent $\bullet$  Most of the works only consider the case where there exists an exact one-to-one correspondence between $\cD_1$ and $\cD_2$. For many more scenarios, however, these two datasets are not necessarily well aligned. For example, consider $\cD_1$ and $\cD_2$ collected from two separate databases, where the users overlap, but are not identical. As a result, there exists only partial one-to-one correspondence. A similar situation also happens to multiple-object tracking: Some objects may leave the scene in one frame, and new objects may enter the scene in subsequent frames. Therefore, not all objects in different frames can be perfectly matched. The RWOC problem with partial correspondence is known as robust-RWOC, or rRWOC \citep{varol2019robust}, and is much less studied in existing literature.

To address these concerns, we propose a new computational framework -- ROBOT (\underline{R}egression with\underline{O}ut correspondence using \underline{B}ilevel \underline{O}ptimiza\underline{T}ion). Specifically, we propose to formulate the regression without correspondence as a continuous optimization problem. Then by exploiting the interaction between the regression model and the data correspondence, we propose to develop a hypergradient approach based on differentiable programming techniques \citep{duchi2008efficient, luise2018differential}. Our hypergradient approach views the data correspondence as an operator of the regression, i.e., for a given $w$, the optimal correspondence is
\begin{align}\label{permutation-subproblem}
\hat{\pi}(w) = \argmin_\pi \cL(w, \pi).
\end{align}
Accordingly, when applying gradient descent to (\ref{eq:ls}), we need to find the gradient with respect to $w$ by differentiating through both the objective function $\cL$ and the data correspondence $\hat{\pi}(w)$. For simplicity, we refer as such a gradient to ``hypergradient''. Note that due to its discrete nature, $\hat{\pi}(w)$ is actually not continuous in $w$. Therefore, such a hypergradient does not exist. To address this issue, we further propose to construct a smooth approximation of $\hat{\pi}(w)$ by adding an additional regularizer to \eqref{permutation-subproblem}, and then we replace $\hat{\pi}(w)$ with our proposed smooth replacement when computing the hyper gradient of $w$. Moreover, we also propose an efficient and scalable implementation of  hypergradient computation based on simple first order algorithms and implicit differentiation, which outperforms conventional automatic differentiation in terms of time and memory cost.

ROBOT can also be extended to the robust RWOC problem, where $\cD_1$ and $\cD_2$ are not necessarily exactly aligned, i.e., some data points in $\cD_1$ may not correspond to any data point in $\cD_2$. Specifically, we relax the constraints on the permutation $\pi(\cdot)$ \citep{liero2018optimal} to automatically match related data points and ignore the unrelated ones.
% during training without sacrificing any computational efficiency.
%  we formulate the regression problem as a bilevel optimization problem, where the lower-level optimization aims to find the optimal correspondence $S^*$ between $(x_1, y)$ and $x_2$, and the upper-level optimization aims to find the optimal regression model $f(\cdot; w^*)$.  

% We formulate the lower-level optimization problem as the Optimal Transport (OT) problem to solve for the optimal correspondence. We further extend the OT problem to Robust OT (ROT) problem, that particularly considers the case when the data in $\cD_1$ and $\cD_2$ is not a one-to-one correspondence. 
 
% The upper-level optimization is solved using gradient-based optimization methods, which requires to compute the gradient of the loss with respect to $w$. By the chain rule, we need to solve the gradient of the optimal correspondence with respect to $w$.
% In order to do that, we relaxed the lower-level optimization problem to be differentiable, i.e., the optimal correspondence is differentiable with respect to $w$, by adding entropy regularization to the optimal transport problem.

At last, we conduct thorough numerical experiments to demonstrate the effectiveness of ROBOT. For RWOC (i.e., exact correspondence), we use several synthetic regression datasets and a real gated flow cytometry dataset, and we show that ROBOT outperforms baseline methods by significant margins. For robust RWOC (i.e., inexact correspondence), we consider a vision-based multiple-object tracking task, and then we show that ROBOT also achieves significant improvement over baseline methods.

% These works slightly improve the computational efficiency, but are nevertheless impractical.

\noindent {\bf Notations}. 
% We use upper letter to represent matrices, and lower letter to represent vectors and functions. 
% Denote $\PP(\cdot)$ the probability measure, i.e., $\PP(\cA)$ is the probability of set $\cA$. 
Let $\norm{\cdot}_2$ denote  the $\ell_2$ norm of vectors, $\langle \cdot, \cdot \rangle$ the inner product of matrices, i.e., $\langle A,B \rangle = \sum_{i,j}A_{ij}B_{ij}$ for matrices $A$ and $B$. $a_{i:j}$ 
are the entries from index $i$ to index $j$ of vector $a$.
Let $\bm{1}_n$ denote an $n$-dimensional vector of all ones.
Denote $\frac{d(\cdot)}{d(\cdot)}$ the gradient of scalars, and $\nabla_{(\cdot)}(\cdot)$ the Jacobian of tensors. We denote $[v_1,v_2]$ the concatenation of two vectors $v_1$ and $v_2$. $\cN(\mu,\sigma^2)$ is the Gaussian distribution with mean $\mu$ and variance $\sigma^2$.
% We then extend the framework to the vision-based Multiple Object Tracking (MOT) task. The goal of MOT is to anticipate the future motion of some objects, e.g., pedestrians. 

% We formulate the lower level optimization problem is an Optimal Transport (OT) problem. OT aims to find the optimal way to transport the mass from one distribution to another distribution, which is widely used to find the alignment between two datasets by viewing the two datasets as realizations of the two distribution. In our case, the two datasets are $\{x_1^{(i)}, y^{(i)}\}_{i=1}^n$ and $\{x_2^{(j)}\}_{j=1}^m$.

% The upper-level optimization is solved using gradient-based optimization methods, which requires to compute the gradient of the loss with respect to $w$. By the chain rule, we need to solve the gradient of the optimal correspondence with respect to $w$.
% In order to do that, we relaxed the lower-level optimization problem to be differentiable, i.e., the optimal correspondence is differentiable with respect to $w$, by adding entropy regularization to the optimal transport problem. 

%{\color{blue} Note: Changed the dimension of $Z$ from $m$ to $e$, since $m$ is used later.}

%!TEX root = ../robot_tech.tex 

\section{ROBOT: A Hypergradient Approach for RWOC}

\label{sec:method}
% \subsection{Problem Statement}

We develop our hypergradient approach for RWOC. Specifically, we first introduce a continuous formulation equivalent to (\ref{eq:ls}), and then propose a smooth bi-level relaxation with an efficient hypergradient descent algorithm.

\subsection{Equivalent Continuous Formulation}

We propose a continuous optimization problem equivalent to (\ref{eq:ls}). Specifically, we rewrite an equivalent form of (\ref{eq:ls}) as follows,
\begin{align}\label{eq:ls-ot}
{\min_{w}\min_{S\in\RR^{n\times n}} \cL(w,S) = \langle C(w),S\rangle\quad\textrm{subject to}~S\in\cP,}
\end{align}
where $\cP$ denotes the set of all $n \times n$ permutation matrices, $C(w)\in\mathbb{R}^{n\times n}$ is the loss matrix with $$C_{ij}(w) = \norm{y_i-f\left(x_{i}, z_j;w \right)}_2^2.$$
Note that we can relax $\cS\in\cP$, which is the discrete feasible set of the inner minimization problem of (\ref{eq:ls-ot}), to a convex set, without affecting the optimality, as suggested by the next theorem.
\begin{proposition}\label{them:rwoc-relax}
Given any $a\in \RR^n$ and $b\in \RR^m$, we define $$\Pi(a, b)=\{A\in\RR^{n\times m}: A\bm{1}_m = a, A^\top\bm{1}_n = b, A_{ij} \geq 0\}.$$ The optimal solution to the inner discrete minimization problem of (\ref{eq:ls-ot}) is also the optimal solution to the following continuous optimization problem,
\begin{align} \label{eq:inner}
    {\min_{S\in\RR^{n\times n}} \langle C(w),S\rangle,  ~~\textrm{s.t.}~S\in\Pi(\bm{1}_n, \bm{1}_n).}
\end{align}
\end{proposition}
This is a direct corollary of the Birkhoff-von Neumann theorem \citep{birkhoff1946three, von1953certain}, and please refer to Appendix \ref{sec:appendix_connection} for more details. Theorem \ref{them:rwoc-relax} allows us to replace $\cP$ in (\ref{eq:ls-ot}) with $\Pi(\bm{1}_n, \bm{1}_n)$, which is also known as  the Birkhoff polytope\footnote{This is a common practice in integer programming \citep{marcus1959diagonals}.}\citep{ziegler2012lectures}. Accordingly, we obtain the following continuous formulation,
\begin{align}\label{eq:cont-robot}
{ \min_{w}\min_{S\in\RR^{n\times n}} \langle C(w),S\rangle\quad\textrm{subject to}~S\in\Pi(\bm{1}_n, \bm{1}_n).}
\end{align}

\begin{remark}
In general, \eqref{eq:ls-ot} can be solved by linear programming algorithms  \citep{dantzig1998linear}. 
\end{remark}

\subsection{Conventional Wisdom: Alternating Minimization}

Conventional wisdom for solving (\ref{eq:cont-robot}) suggests to use alternating minimization (AM, \citet{abid2017linear}). Specifically, at the $k$-th iteration, we first update $S$ by solving
\begin{align*}
{ S^{(k)} = \argmin_{S\in\Pi(\bm{1}_n, \bm{1}_n)}\cL(w^{(k-1)},S),}
\end{align*}
and then given $S^{(k)}$, we update $w$ using gradient descent or exact minimization, i.e., 
\begin{align*}
w^{(k)} = w^{(k-1)} - \eta\nabla_w\cL(w^{(k-1)}, S^{(k)}).%\quad\textrm{or}\quad w^{(k)} = \argmin_{w}\cL(w, \pi^{(k)}).
\end{align*}
However, AM works poorly for solving (\ref{eq:cont-robot}) in practice. This is because $w$ and $S$ have a strong interaction throughout the iterations: A slight change to $w$ may lead to significant change to $S$. Therefore, the optimization landscape is ill-conditioned, and AM can easily get stuck at local optima.

\subsection{Smooth Bi-level Relaxation}

To tackle the aforementioned computational challenge, we propose a hypergradient approach, which can better handle the interaction between $w$ and $S$. Specifically, we first relax (\ref{eq:cont-robot}) to a smooth bi-level optimization problem, and then we solve the relaxed bi-level optimization problem using the hypergradient descent algorithm.

We rewrite (\ref{eq:cont-robot}) as a smoothed bi-level optimization problem, 
\begin{equation} \label{eq:sinkhorn}
{\min_{w} \cF_{\epsilon}(w)=\langle C(w), S^*_\epsilon(w) \rangle, ~\textrm{subject~to}~S^*_\epsilon(w) = \argmin_{S \in \Pi(\textbf{1}_n,\textbf{1}_n)} \langle C(w),S\rangle +  \epsilon H(S), }
\end{equation}
where $H(S) = \langle \log S, S \rangle$ is the entropy of $S$. The regularizer $H(S)$ in (\ref{eq:sinkhorn}) alleviates the sensitivity of $S^*(w)$ to $w$. Note that if without such a regularizer, we solve
\begin{align}\label{nonsmooth-OT}
{S^*(w) = \argmin_{S \in \Pi(\textbf{1}_n,\textbf{1}_n)} \langle C(w),S\rangle.}
\end{align}
The resulting $S^*(w)$ can be discontinuous in $w$. This is because $S^*(w)$ is the optimal solution of a linear optimization problem, and usually lies on a vertex of $\Pi(\textbf{1}_n, \textbf{1}_n)$. This means that if we change $w$, $S^*(w)$ either stays the same or jumps to another vertex of $\Pi(\textbf{1}_n, \textbf{1}_n)$. The jump makes $S^*(w)$ highly sensitive to $w$. To alleviate this issue, we propose to smooth $S^*(w)$ by adding an entropy regularizer to the lower level problem. The entropy regularizer enforces $S^*_\epsilon(w)$ to stay in the interior of $\Pi(\textbf{1}_n, \textbf{1}_n)$, and $S^*_\epsilon(w)$ changes smoothly with respect to $w$, as suggested by the following theorem.
\begin{theorem}
For any $\epsilon>0$, $S^*_\epsilon(w)$ is differentiable, if the cost $C(w)$ is differentiable with respect to $w$. Consequently, the objective $\cF_{\epsilon}(w)=\langle C(w), S^*_\epsilon(w) \rangle$ is also differentiable. 
\end{theorem}
\vspace{-0.05in}
The proof is deferred to Appendix \ref{sec:appendix_proof}. Note that (\ref{eq:sinkhorn}) provides us a new perspective to interpret the relationship between $w$ and $S$. As can be seen from (\ref{eq:sinkhorn}), $w$ and $S$ have different priorities: $w$ is the parameter of the leader problem, which is of the higher priority; $S$ is the parameter of the follower problem, which is of the lower priority, and can also be viewed as an operator of $w$ -- denoted by $S^*_\epsilon(w)$. Accordingly, when we minimize (\ref{eq:sinkhorn}) with respect to $w$ using gradient descent, we should also differentiate through $S^*_\epsilon$. We refer to such a gradient as ``hypergradient'' defined as follows,
\begin{align*}
\nabla_w \cF_\epsilon (w) = \frac{\partial \cF_\epsilon(w)}{\partial C(w)}\frac{\partial C(w)}{\partial w}  + \frac{\partial \cF_\epsilon (w)}{\partial S_\epsilon^*(w)} \frac{\partial S_\epsilon^*(w)}{\partial w}= \nabla_w\cL(w, S) + \frac{\partial \cF_\epsilon(w)}{\partial S_\epsilon^*(w)}\frac{\partial S_\epsilon^*(w)}{\partial w}.
\end{align*}
We further examine the alternating minimization algorithm from the bi-level optimization perspective: Since $\nabla_w\cL(w^{(k-1)}, S^{(k)})$ is not differentiable through $S^{(k)}$, AM is essentially using an inexact gradient. From a game-theoretic perspective\footnote{The bilevel formulation can be viewed as a Stackelberg game.}, (\ref{eq:sinkhorn}) defines a competition between the leader $w$ and the follower $S$. When using AM, $S$ only reacts to what $w$ has responded. In contrast, when using the hypergradient approach, the leader essentially recognizes the follower's strategy and reacts to what the follower is anticipated to response through $ \frac{\partial \cF_\epsilon(w)}{\partial S_\epsilon^*(w)}\frac{\partial S_\epsilon^*(w)}{\partial w}$. In this way, we can find a better descent direction for $w$.

\begin{remark} 
We use a simple example of quadratic minimization to illustrative why we expect the bilevel optimization formulation in (\ref{eq:sinkhorn}) to enjoy a benign optimization landscape. We consider a quadratic function
\begin{align}\label{qp-exp}
L(a_1,a_2)= a^\top P a + b^\top a,
\end{align}
where $a_1\in \RR^{d_1}$, $a_2\in \RR^{d_2}$, $a=[a_1, a_2]$, $P \in \RR^{(d_1+d_2)\times (d_1+d_2)}$, $b \in \RR^{d_1+d_2}$. Let $P = \rho \bm{1}_{d_1+d_2} \bm{1}_{d_1+d_2}^\top + (1-\rho) I_{d_1+d_2}$, where $I_{d_1+d_2}$ is the identity matrix, and $\rho$ is a constant. We solve the following bilevel optimization problem,
\begin{align}\label{qp-bi-level}
\textstyle{ \min_{a_1} F(a_1) = L(a_1,a_2^*(a_1))\quad\textrm{subject~to}~a_2^*(a_1) = \argmin_{a_2}L(a_1,a_2)+\lambda \norm{a_2}_2^2,}
\end{align}
where $\lambda$ is a regularization coefficient. The next proposition shows that $\nabla^2 F(a_1)$ enjoys a smaller condition number than $\nabla^2_{a_1a_1} L(a_1,a_2)$, which  corresponds to the problem that AM solves.
\begin{proposition}\label{better-cond} Given $F$ defined in (\ref{qp-bi-level}), we have
\begin{align*}
\frac{\lambda_{\max}(\nabla^2 F(a_1))}{\lambda_{\min}(\nabla^2 F(a_1))} = 1+\frac{1-\rho+\lambda}{d_2 \rho -\rho+\lambda +1}\cdot\frac{d_1 \rho}{1-\rho}\quad\textrm{and}\quad \frac{\lambda_{\max}(\nabla^2_{a_1a_1} L(a_1,a_2))}{\lambda_{\min}(\nabla^2_{a_1a_1} L(a_1,a_2))} = 1+\frac{d_1 \rho}{1-\rho}.
\end{align*}
\end{proposition}
The proof is deferred to Appendix \ref{sec:fast_convergence}. As suggested by Proposition \ref{better-cond}, $F(a_1)$ is much better-conditioned than $L(a_1,a_2)$ in terms of $a_1$ for high dimensional settings.
\end{remark}

\subsection{Solving rWOC by Hypergradient Descent}

%% \begin{wrapfigure}{R}{0.4\textwidth}
%%    \begin{minipage}{0.4\textwidth}
%%    \vspace{-20pt}
%      \begin{algorithm}[t]
%       \caption{ROBOT}
%    \label{alg:unroll_sinkhorn}
%        \begin{algorithmic}
%          \STATE Initialize $w$\;
%        \WHILE{not converged}
%          \STATE $p^*, G, q^*=$Sinkhorn($C(w);\epsilon$)
%          \STATE $S^*_{\epsilon, ij} (w) =n p^*_i G_{ij} q^*_j$
%          \STATE Compute $\nabla_w \cL_\epsilon$ using (\ref{eq:update_w})
%          \STATE $w = w -\beta\nabla_w \cL_\epsilon$
%        \ENDWHILE
%        \end{algorithmic}
%      \end{algorithm}
%%    \end{minipage}
%%    % \vspace{-10pt}
%%  \end{wrapfigure}
  
We present how to solve (\ref{eq:sinkhorn}) using our hypergradient approach. Specifically, we compute the ``hypergradient'' of $ \cF_\epsilon(w)$ based on the following theorem.
\begin{theorem} \label{them:gradient}
The gradient of $\cF_{\epsilon}$ with respect to $w$ is
\begin{align}
    \nabla_{w}{\cF_{\epsilon}}(w)  = \frac{1}{\epsilon} \sum_{i,j=1}^{n,n} \left( (1- C_{ij})S^*_{\epsilon,ij} + \sum_{h,\ell=1}^{n,n} C_{h\ell}S^*_{\epsilon,h\ell}P_{hij} + \sum_{h,\ell=1}^{n,n} C_{h\ell}S^*_{\epsilon,h\ell}Q_{\ell ij} \right)\nabla_{w}{ C_{ij}}. \label{eq:update_w}
\end{align}
  where $\displaystyle \begin{bmatrix}
      P\\
      Q
  \end{bmatrix} := 
  \begin{bmatrix}
   - H^{-1} D \\
   \bm{0}
  \end{bmatrix} \quad \text{with} \quad 
      P, Q\in\mathbb{R}^{n\times n \times n}, - H^{-1} D\in \mathbb{R}^{(2n-1)\times n \times n}, \bm{0}\in \mathbb{R}^{1\times n \times n},$
  {\normalsize
\begin{align*}
    & D_{\ell ij} = \dfrac{1}{n\epsilon} \begin{cases}
    \delta_{\ell i} S^*_{\epsilon,ij},\quad  \ell=1, \cdots, n; \\
    \delta_{\ell j} S^*_{\epsilon,ij},\quad  \ell=n+1, \cdots, 2n-1,
    \end{cases} 
    H^{-1} = -{\epsilon}{n} \begin{bmatrix} 
    I_n +  \bar{S^*_{\epsilon}} \mathcal{K}^{-1} \bar{S^*_{\epsilon}}^T  & -  \bar{S^*_{\epsilon}} \mathcal{K}^{-1} \\
    -\mathcal{K}^{-1} \bar{S^*_{\epsilon}} ^T  & \mathcal{K}^{-1}
    \end{bmatrix}, \\
    &\textrm{and}\quad \mathcal{K} = I_{n-1}- \bar{S^*_{\epsilon}}^T  \bar{S^*_{\epsilon}}, \quad \bar{S^*_{\epsilon}} = S^*_{\epsilon, 1:n,1:n-1}.
\end{align*}
  }
\end{theorem}
The proof is deferred to Appendix \ref{sec:appendix_proof}. Theorem \ref{them:gradient} suggests that we first solve the lower level problem in (\ref{eq:sinkhorn}), 
\begin{equation} \label{eq:inner_sinkhorn}
    { S^*_\epsilon = \argmin_{S \in \Pi(\textbf{1}_n,\textbf{1}_n)} \langle C(w),S\rangle +  \epsilon H(S),}
\end{equation}
and then substitute $S^*_\epsilon$ into (\ref{eq:update_w}) to obtain $\nabla_{w}{\cF_{\epsilon}}(w)$. 

Note that the optimization problem in (\ref{eq:inner_sinkhorn}) can be efficiently solved by a variant of Sinkhorn algorithm \citep{cuturi2013sinkhorn,benamou2015iterative}. Specifically, (\ref{eq:inner_sinkhorn}) can be formulated as an entropic optimal transport (EOT) problem \citep{monge1781memoire, kantorovich1960mathematical}, which aims to find the optimal way to transport the mass from a categorical distribution with weight  $\mu = [\mu_1, \dots, \mu_n]^\top$ to another categorical distribution with weight $\nu = [\nu_1, \dots, \nu_m]^\top$,
%The EOT problem considers two discrete distributions $\PP_1$, $\PP_2$ defined on supports $\mathcal{A}=\{a_i\}_{i=1}^n$, $\mathcal{B}=\{b_j\}_{j=1}^m$, respectively. Denote $\PP_1(\{a_i\}) = \mu_i$, $\PP_2(\{b_j\}) = \nu_j$, and let $\mu = [\mu_1, \dots, \mu_n]^\top$, $\nu = [\nu_1, \dots, \nu_m]^\top$. Further denote $M \in \RR^{n \times m}$ the cost matrix with $M_{ij}$ the transport cost. The EOT problem solves the following entropy regularized optimization problem,
\begin{equation}\label{eq:kanto}
\begin{split}
    \Gamma^* &= {\argmin_{\Gamma \in \Pi(\mu, \nu)} } \langle M, \Gamma \rangle + \epsilon H(\Gamma),\\
    \textrm{with}~~\Pi(\mu, \nu)&=\{\Gamma\in\RR^{n\times m}: \Gamma\bm{1}_m = \mu, \Gamma^\top\bm{1}_n = \nu, \Gamma_{ij} \geq 0\},
\end{split}
\end{equation}
where $M \in \RR^{n \times m}$ is the cost matrix with $M_{ij}$ the transport cost. When we set the two categorical distributions as the empirical distribution of $\cD_1$ and $\cD_2$, respectively,
\begin{align*}
M=C(w)\quad\textrm{and}\quad\mu=\nu=\bm{1}_n/n,
\end{align*}
one can verify that (\ref{eq:kanto}) is a scaled lower problem of (\ref{eq:sinkhorn}), and their optimal solutions satisfies $S^*_{\epsilon} = n \Gamma^*$.
Therefore, we can apply Sinkhorn algorithm to solve the EOT problem in \eqref{eq:kanto}: At the $\ell$-th iteration, we take 
% where at the $\ell$-th iteration, we update
 \begin{align*}
 p^{(\ell+1)} = \frac{\mu}{Gq^{(\ell)}}\;\textrm{and}\; q^{(\ell+1)} = \frac{\nu}{G^\top p^{(\ell+1)}}, \;\; \textrm{where}\;\; q^{(0)} = \frac{1}{n}\bm{1}_n\;\textrm{and}\; G_{ij} = \exp\left(\frac{-C_{ij}(w)}{\epsilon}\right),
 \end{align*}
$G \in \RR^{n \times n}$, and the division here is entrywise. Let $p^*$ and $q^*$ denote the stationary points. Then we obtain $S_{\epsilon, ij}^* = n p_i^* G_{ij} q_j^*$. 
\begin{remark}
The Sinkhorn algorithm is iterative and cannot exactly solve (\ref{eq:inner_sinkhorn}) within finite steps. As the Sinkhorn algorithm is very efficient and attains linear convergence, it suffices to well approximate the gradient $\nabla_{w}{\cF_{\epsilon}}(w)$ using the output inexact solution.
\end{remark}

\section{ROBOT for Robust Correspondence}
\label{sec:robust}
%\vspace{-0.05in}

% \vspace{-0.05in}
% \subsection{From Optimal Transport to Relaxed Optimal Transport}
% \vspace{-0.1in}

% \begin{wrapfigure}{r}{5.cm}
% % \vspace{-30pt}
% \centering
% \includegraphics[width=.95\linewidth]{figures/synthetic/robust_demo.png}
% \vspace{-5pt}
% \caption{\label{fig:illu_rot} \textit{Illustration of $\Gamma_{\rm r}^*$, $\bar{{\mu}}^*$, and $\bar{{\nu}}^*$ for robust matching, $\epsilon=10^{-5}$.}}
% \vspace{-10pt}
% \end{wrapfigure}
We next propose a robust version of ROBOT to solve rRWOC \citep{varol2019robust}. Note that in (\ref{eq:sinkhorn}), the constraint $S\in \Pi(\bm{1}_n, \bm{1}_n)$ enforces a one-to-one matching between $\cD_1$ and $\cD_2$. For rRWOC, however, such an exact matching may not exist. For example, we have $n<m$, where $n=|\cD_1|$, $m=|\cD_2|$. Therefore, we need to relax the constraint on $S$.

\begin{figure}[htb!]
    \centering
%    \vspace{-15pt}
    \begin{subfigure}{0.48\linewidth}
        \includegraphics[width=0.99\linewidth]{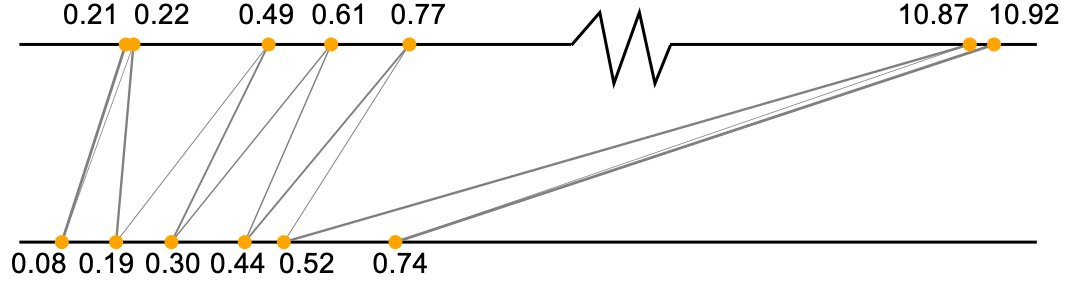}
        \vskip -8pt
        \caption{Original}
    \end{subfigure}
    \begin{subfigure}{0.48\linewidth}
        \includegraphics[width=0.99\linewidth]{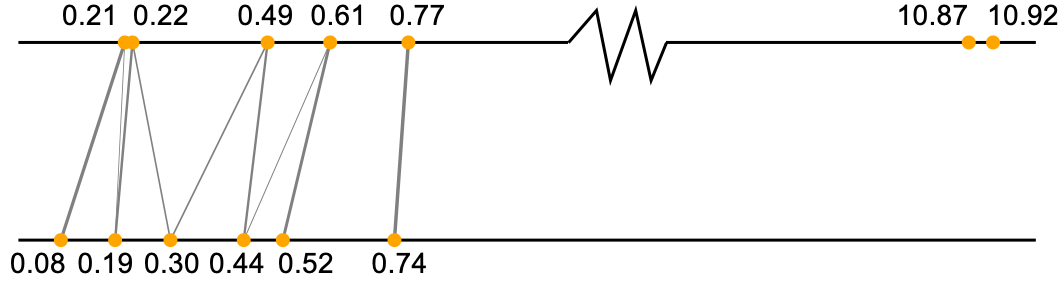}
        \vskip -8pt
        \caption{Robust}
    \end{subfigure}
%    \vspace{-10pt}
    \caption{ \textit{Illustrative example of exact (L) and robust (R) alignments.} The robust alignment can drop potential outliers and only match data points close to each other.}
    \label{fig:illu_rot}
%    \vspace{-15pt}
\end{figure}

Motivated by the connection between (\ref{eq:sinkhorn}) and (\ref{eq:kanto}), we propose to solve the following lower problem\footnote{The idea is inspired by the marginal relaxation of optimal transport, first independently proposed by \citet{kondratyev2016new} and \citet{chizat2018interpolating}, and later developed by \citet{chizat2018unbalanced}  and \citet{liero2018optimal}. \citet{chizat2018scaling} share the same formulation as ours.},
\begin{align}
     (S^*_{\rm r}(w),\bar{{\mu}}^*, \bar{{\nu}}^*)  = &\argmin_{S\in \Pi(\bar{{\mu}}, \bar{{\nu}})}
    \langle C(w), S \rangle + \epsilon H(S), \label{eq:robust_rwoc}\\
    &{\rm subject~to~} \bar{\mu}^\top \bm{1}_n = n, ~ \bar{\nu}^\top \bm{1}_m = m, ~\norm{\bar{{\mu}} - \bm{1}_n}_2^2 \leq \rho_1, ~\norm{\bar{{\nu}}- \bm{1}_m}^2_2 \leq \rho_2,\notag
\end{align}
where $S^*_{\rm r}(w)\in\RR^{n\times m}$ denotes an inexact correspondence between $\cD_1$ and $\cD_2$. As can be seen in (\ref{eq:robust_rwoc}), we relax the marginal constraint $\Pi(\bm{1}, \bm{1})$ in (\ref{eq:sinkhorn}) to $\Pi(\bar{{\mu}}, \bar{{\nu}})$, where $\bar{{\mu}}, \bar{{\nu}}$ are required to not deviate much from $\bm{1}$.  Problem (\ref{eq:robust_rwoc}) relaxes the marginal constraints $\Pi(\bm{1}, \bm{1})$ in the original problem to $\Pi(\bar{{\mu}}, \bar{{\nu}})$, where $\bar{{\mu}}, \bar{{\nu}}$ are picked such that they do not deviate too much from $\bm{1}$. Illustrative examples of the exact and robust alignments are provided in Figure \ref{fig:illu_rot}. 

Computationally, (\ref{eq:robust_rwoc}) can be solved by taking the Sinkhorn iteration and the projected gradient iteration in an alternating manner (See more details in Appendix \ref{sec:appendix_robust_foward}). Given $S^*_{\rm r}(w)$, we solve the upper level optimization in (\ref{eq:sinkhorn}) to obtain $w^*$, i.e.,
\begin{align*}
    & w^* =  \argmin_{w}
    \langle C(w), S^*_{\rm r}(w) \rangle.
\end{align*}
Similar to the previous section, we use a first-order algorithm to solve this problem, and we derive explicit expressions for the update rules. See Appendix \ref{sec:appendix_robust} for details.

%\noindent 

% \subsection{Continuous Case}
% Analogous to the previous unrolling algorithm with NN, we denote
% \begin{align*}
%     & \mathcal{L}_{\rm r1}(p, q;w,\theta) =  \int c([y,x_1], x_2;w) d\pi(y,x_1, x_2; \bar{p}(\theta), \bar{q}(\theta), \theta) + \eta_1 \mathcal{D}_f (p, \bar{p}(\theta)) + \eta_2 \mathcal{D}_f (q, \bar{q}(\theta)) \\
%     & \mathcal{L}_{\rm r2}(p, q;w,\theta) =  \int c([y,x_1], x_2;w) d\pi(y,x_1, x_2; \bar{p}(\theta), \bar{q}(\theta), \theta) + \epsilon \int \log \pi d \pi  + \eta_1 \mathcal{D}_f (p, \bar{p}(\theta)) + \eta_2 \mathcal{D}_f (q, \bar{q}(\theta)).
% \end{align*}
% And the parameter update is
% \begin{align*}
%     & w = w - \eta \frac{d \mathcal{L}_{\rm r1}(p, q;w,\theta^{(K)})}{d w}, \\
%     & \theta = \theta - \eta \frac{d \mathcal{L}_{\rm r2}(p, q;w,\theta)}{d \theta},
% \end{align*}
% where $\theta^{(K)}$ is defined as the output of $K$ steps of gradient descent (or variants of gradient descent) of $\mathcal{L}_{\rm r2}$ on $\theta$.

% \section{Handling the Inconsistency}

% \cite{pananjady2017linear, abid2018least}
% \input{tex_files/mot}
% \input{algorithm}

%!TEX root = ../robot_tech.tex

%\vspace{-0.1in}
\section{Experiment}
%\vspace{-0.05in}

We evaluate ROBOT and ROBOT-robust on both synthetic and real-world datasets, including flow cytometry and multi-object tracking. We first present numerical results and then we provide insights in the discussion section. Experiment details and auxiliary results can be found in Appendix \ref{sec:appendix_exp}.
% In this section we evaluate the proposed computational framework ROBOT using synthetic datasets, and apply it to two real-world applications, i.e., gated flow cytometry and multi-object tracking.

%\vspace{-0.13in}
\subsection{Unlabeled Sensing}
%\vspace{-0.08in}
\label{sec:sec51}
% \subsubsection{Linear Model}
% \vspace{-0.1in}
% To evaluate the generalization ability of the learned regressor, we consider the evaluation of the following two settings: 
% a). In sensing problems, we learn $f(\cdot; w)$ from a batch of permuted data, then make predictions on new incoming data \citep{pmlr-v97-tsakiris19a,rose2014signaling}. We refer this setting as the \textit{sensing setting}.  b). For data collected from multiple sources, we have input features $x^{(i)}_1$ and response $y^{(i)}$  on platform 1, and some other input features $x^{\pi(i)}_2$ on platform 2. Since our goal is the prediction on platform 1, we use all data on platform 2 as the training data, and split the data on platform 1 to be the training data and the test data \citep{unnikrishnan2013anonymizing}. We refer this setting as the \textit{privacy setting}.

\textbf{Data Generation}. We follow the unlabeled sensing setting \citep{pmlr-v97-tsakiris19a} and generate $n=1000$ data points $\{(y_i, z_i)\}_{i=1}^n$, where $z_i \in \mathbb{R}^{e}$. Note here we take $d=0$.
We first generate $z_i, w \sim \mathcal{N}(\bm{0}_{e},\bm{I}_{e})$, and $\varepsilon_i \sim \mathcal{N}(0,\rho_{\rm noise}^2)$. Then we compute $y_i= z_i^\top w + \varepsilon_i$.
% Our model is $y = w^\top x + \epsilon$, where $w, x \sim \mathcal{N}(0,1)$ and $\epsilon \sim \mathcal{N}(0, \rho_{\rm noise}^2)$. 
We randomly permute the order of 50\% of $z_i$ so that we lose the $Z$-to-$Y$ correspondence. We generate the test set in the same way, only without permutation.

% {\bf Data Generation}. Following the standard setting in unlabeled sensing \citep{pmlr-v97-tsakiris19a}, we generate $n$ data samples $\{(y^{(i)}, x_2^{(i)})\}_{i=1}^n$, where we have $x_2^{(i)}\in \mathbb{R}^{d_2}$, $y^{(i)}\in \mathbb{R}$ , and $x_1^{(i)}\in \mathbb{R}$ is omitted, i.e., we adopt $d_1=0$. First, we generate $x_2^{(i)}$ and $y^{(i)} = f(x_2^{(i)};w) + \varepsilon^{(i)}$, where $x_2^{(i)}, w \sim \mathcal{N}(\bm{0}_{d_2},\bm{I}_{d_2})$, and $\varepsilon^{(i)} \sim \mathcal{N}(0,\rho_{\rm noise}^2)$. Then, we randomly permute $50\%$ of the indexes of $\{x_2^{(i)}\}$. We then generate the test dataset with $n$ data items from the same distribution as the training data, but without permutation.

\vspace{10pt}

\noindent \textbf{Baselines and Training}. We consider the following scalable methods:
%\vspace{-0.1in}
\begin{enumerate}
\itemsep0em 
\item \textit{Oracle}: Standard linear regression where no data are permuted. 
% \item \textit{Partial}: Standard linear regression where only the data on platform 1 is considered. 
\item \textit{Least Squares} (\textit{LS}): Standard linear regression, i.e., treating the data as if they are not permuted. 
\item \textit{Alternating Minimization } (\textit{AM}, \cite{abid2017linear}):
% Alternatingly minimize \eqref{eq:ls-ot}.
We iteratively solve the correspondence given $w$, and update $w$ using gradient descent with the correspondence.
% Gradient descent on $\langle C(w), S^* \rangle$ with respect to $w$, where $S^*= \argmin_{S} \langle C(w), S \rangle$. 
% \zha{Need to make alternating clear, otherwise it's the same as ROBOTR!}
\item \textit{Stochastic EM} \citep{abid2018stochastic}: A stochastic EM approach to recover the permutation.
\item \textit{Robust Regression} (\textit{RR}, \cite{Slawski2019Linear, Slawski2019Two}). A two-stage block coordinate descent approach to discard outliers and fit regression models.
% \item \citep{saab2019shuffled}
\item \textit{Random Sample} (\textit{RS}, \cite{varol2019robust}): A random sample consensus (RANSAC) approach to estimate $w$.
% Random sample $d_2$ data points from $\cD_1$ and $\cD_2$, respectively, and fit a regression model. Repeat this for many times and pick the model where the most data points that are reasonably fitted. 
\end{enumerate}
%\vspace{-0.1in}
We initialize AM, EM and ROBOT using the output of RS with multi-start. We adopt a linear model $f(Z;w)=Z^\top w$. Models are evaluated by the relative error on the test set, i.e., error
$=\sum_i (\hat{y}_i-y_i)^2 / \sum_i (y_i-\bar{y})^2$, where $\hat{y}_i$ is the predicted label, and $\bar{y}$ is the mean of $\{y_i\}$.

\begin{figure}[htb!]
    \centering
%    \vspace{-10pt}
    \begin{subfigure}{0.35\linewidth}
        \includegraphics[width=0.95\linewidth]{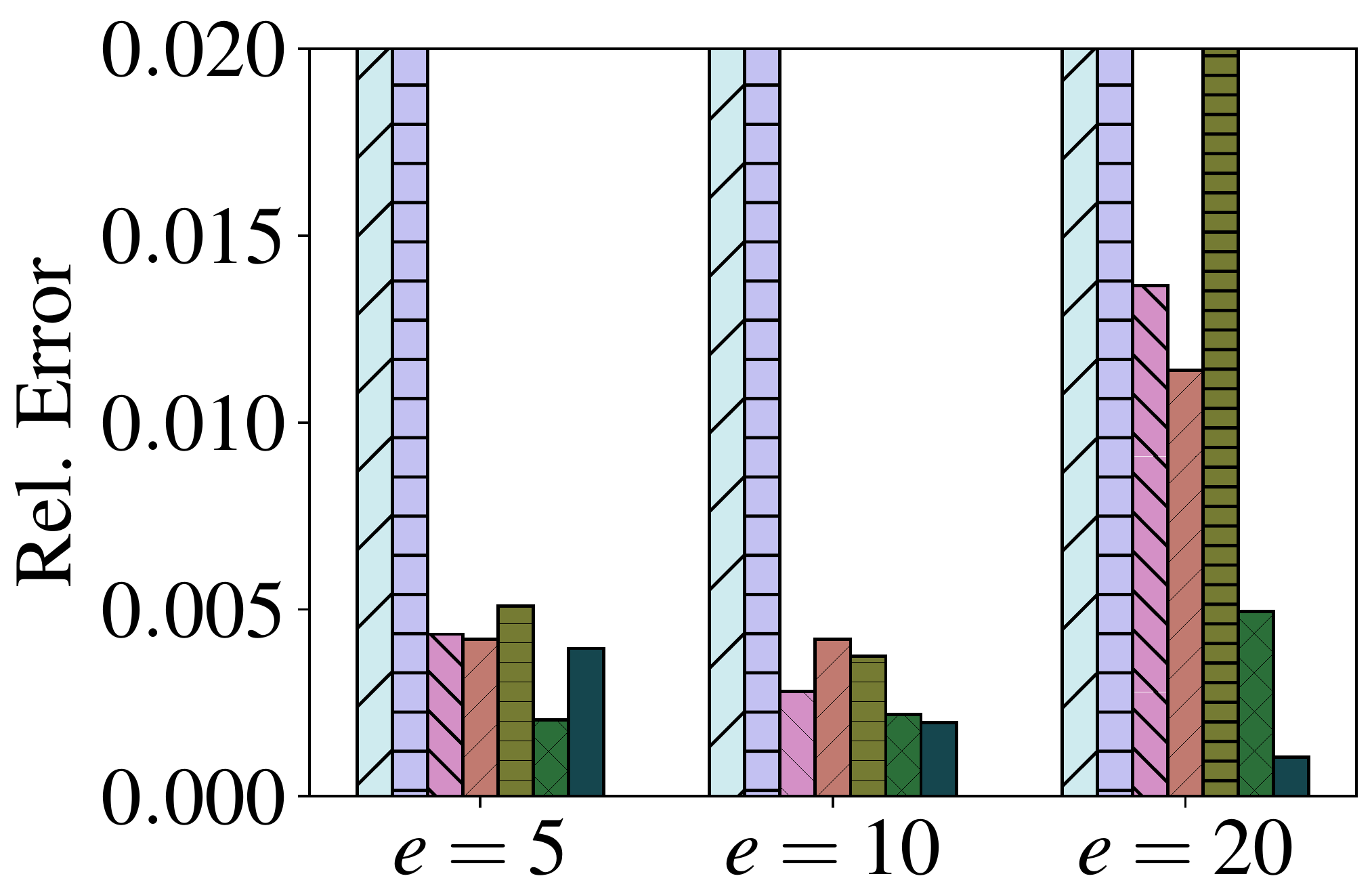}
        \vskip -5pt
        \caption{$\rho_{\rm noise}^2=0.1$}
    \end{subfigure}
    \begin{subfigure}{0.35\linewidth}
        \includegraphics[width=0.95\linewidth]{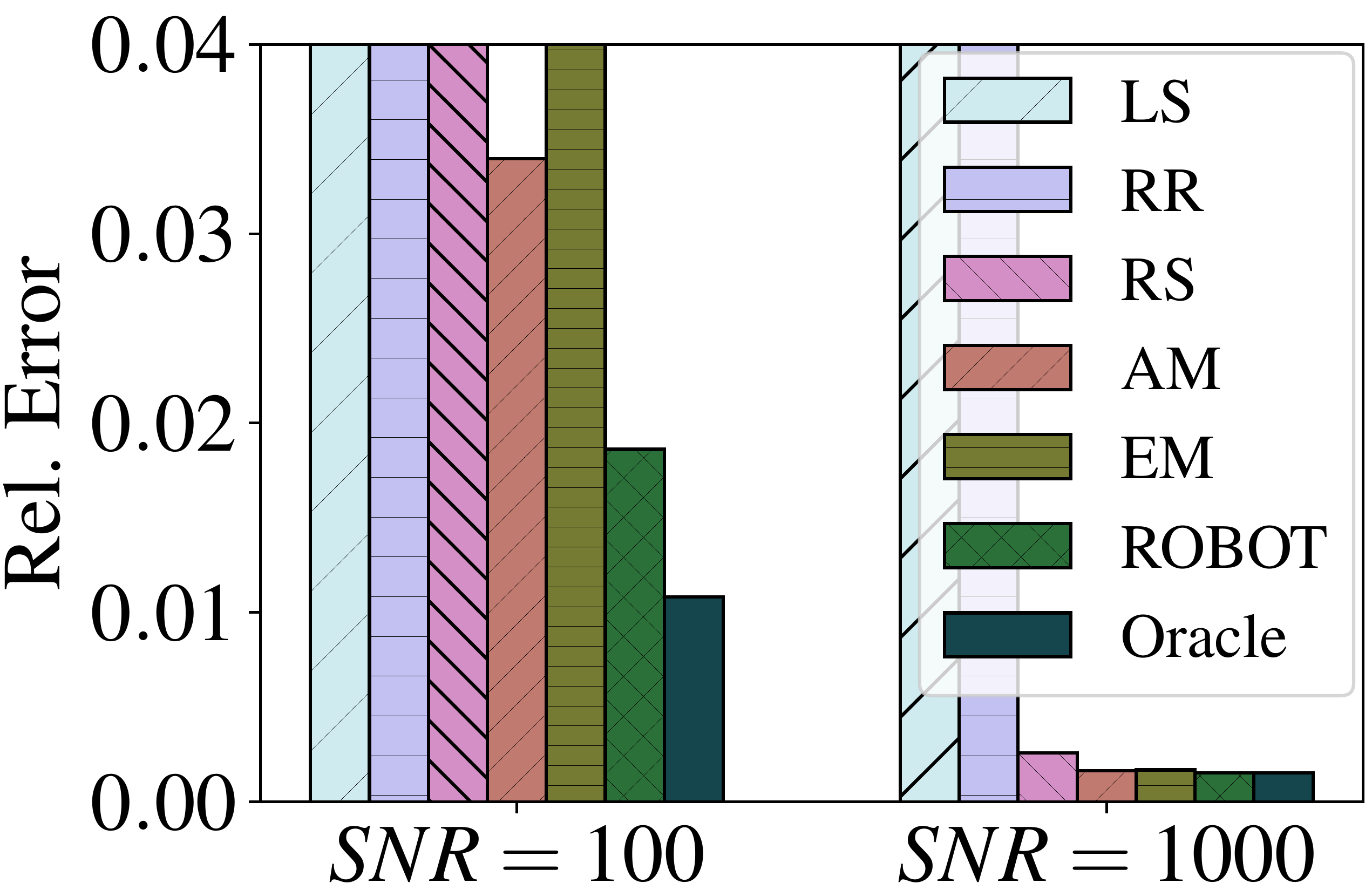}
        \vskip -5pt
        \caption{$e=10$}
    \end{subfigure}
%    \vspace{-5pt}
    \caption{\textit{Unlabeled sensing. Results are the mean over 10 runs.} \texttt{SNR}$=\norm{w}^2_2/\rho_{\rm noise}^2$ \textit{is the signal-to-noise ratio.}}
    \label{fig:synthetic_sensing_partial}
    %\vspace{5pt}
\end{figure}

% \textbf{Evaluation}. We evaluate the $\textrm{RSS}/\textrm{TSS}$ $=1-R^2$ on the test dataset, where $\textrm{RSS}$ is the residual sum of the squares, and $\textrm{TSS}$ is the total sum of the squares.

% \textbf{Results}. We visualize the fitting error (\textrm{RSS}/\textrm{TSS}) in Figure \ref{fig:synthetic_sensing_partial}, where each result is averaged from $10$ generated datasets. We compare to 5 baseline methods and 1 oracle method under different $d_2$ and the Signal to Noise Ratio \texttt{SNR}$=\norm{w}^2_2/\rho_{\rm noise}^2$. In all experiments, ROBOT achieves significantly better result than the baselines. 
\vspace{10pt}
\noindent \textbf{Results}. We visualize the results in Figure \ref{fig:synthetic_sensing_partial}. In all the experiments, ROBOT achieves better results than the baselines. Note that the relative error is larger for all methods except Oracle as the dimension and the noise increase. For low dimensional data, e.g., $e=5$, our model achieves even better performance than Oracle.  We include more discussions on using RS as initializations in  Section \ref{sec:diss}.

% \begin{figure}[h!]
%     \centering
%     \begin{subfigure}{0.23\linewidth}
%         \includegraphics[width=\linewidth]{figures/synthetic_sensing/sensing d1.png}
%         \caption{$n=1000,~\rho_{\rm noise}^2=0.1$.}
%     \end{subfigure}
%      \begin{subfigure}{0.23\linewidth}
%         \includegraphics[width=\linewidth]{figures/synthetic_sensing/sensing n.png}
%         \caption{$d_1=10,~\rho_{\rm noise}^2=0.1$.}
%     \end{subfigure}
%     \begin{subfigure}{0.23\linewidth}
%         \includegraphics[width=\linewidth]{figures/synthetic_sensing/sensing noise.png}
%         \caption{$n=1000,~d_1=10$.}
%     \end{subfigure}
%     % \vfill
%     \caption{RSS/TSS error with regressor 1. The three experiments test the influence of $d_2$,  $n$, and $\rho_{\rm noise}^2$, respectively.}
%     \label{fig:synthetic_sensing}
% \end{figure}

%\vspace{-0.13in}
\subsection{Nonlinear Regression}
%\vspace{-0.08in}
\label{sec:sec52}
% \begin{figure}[t!]
% \vspace{-10pt}
%     \centering
%     \begin{subfigure}{0.2\linewidth}
%         \includegraphics[width=\linewidth]{figures/synthetic_nonlinear/nonlinear_d1.pdf}
%         % \caption{ $n=1000,\\d_2=2,~\rho_{\rm noise}^2=0.1$}
%     \end{subfigure}
%     \begin{subfigure}{0.2\linewidth}
%         \includegraphics[width=\linewidth]{figures/synthetic_nonlinear/nonlinear_d2.pdf}
%         % \caption{$n=1000,\\d_1=2,~\rho_{\rm noise}^2=0.1$}
%     \end{subfigure}
%     % \vfill
%     \begin{subfigure}{0.2\linewidth}
%         \includegraphics[width=\linewidth]{figures/synthetic_nonlinear/nonlinear_noise.pdf}
%         % \caption{ $n=1000,\\d_1=2,~d_2=3$}
%     \end{subfigure}
%     \begin{subfigure}{0.2\linewidth}
%         \includegraphics[width=\linewidth]{figures/synthetic_nonlinear/nonlinear_n.pdf}
%         % \caption{ $d_1=2,~d_2=3,\\ \rho_{\rm noise}^2=0.1$}
%     \end{subfigure}
%     \begin{subfigure}[t]{0.16\linewidth}
%     \vspace{-30pt}
%         \includegraphics[width=\linewidth]{figures/synthetic_nonlinear/nonlinear_legend.pdf}
%     \end{subfigure}
%     \vspace{-5pt}
%     % \caption{Fitting error with nonlinear regression model.}
%     \caption{\textit{Nonlinear regression. We use $n=1000$, $d_1=2$, $d_2=3$, $\rho_{\rm noise}^2=0.1$ as defaults.}}
%     \label{fig:synthetic_nonlinear}
%     \vspace{-10pt}
% \end{figure}

\begin{figure}[htb!]
% \vspace{-10pt}
    \centering
    \begin{subfigure}{0.24\linewidth}
        \includegraphics[width=\linewidth]{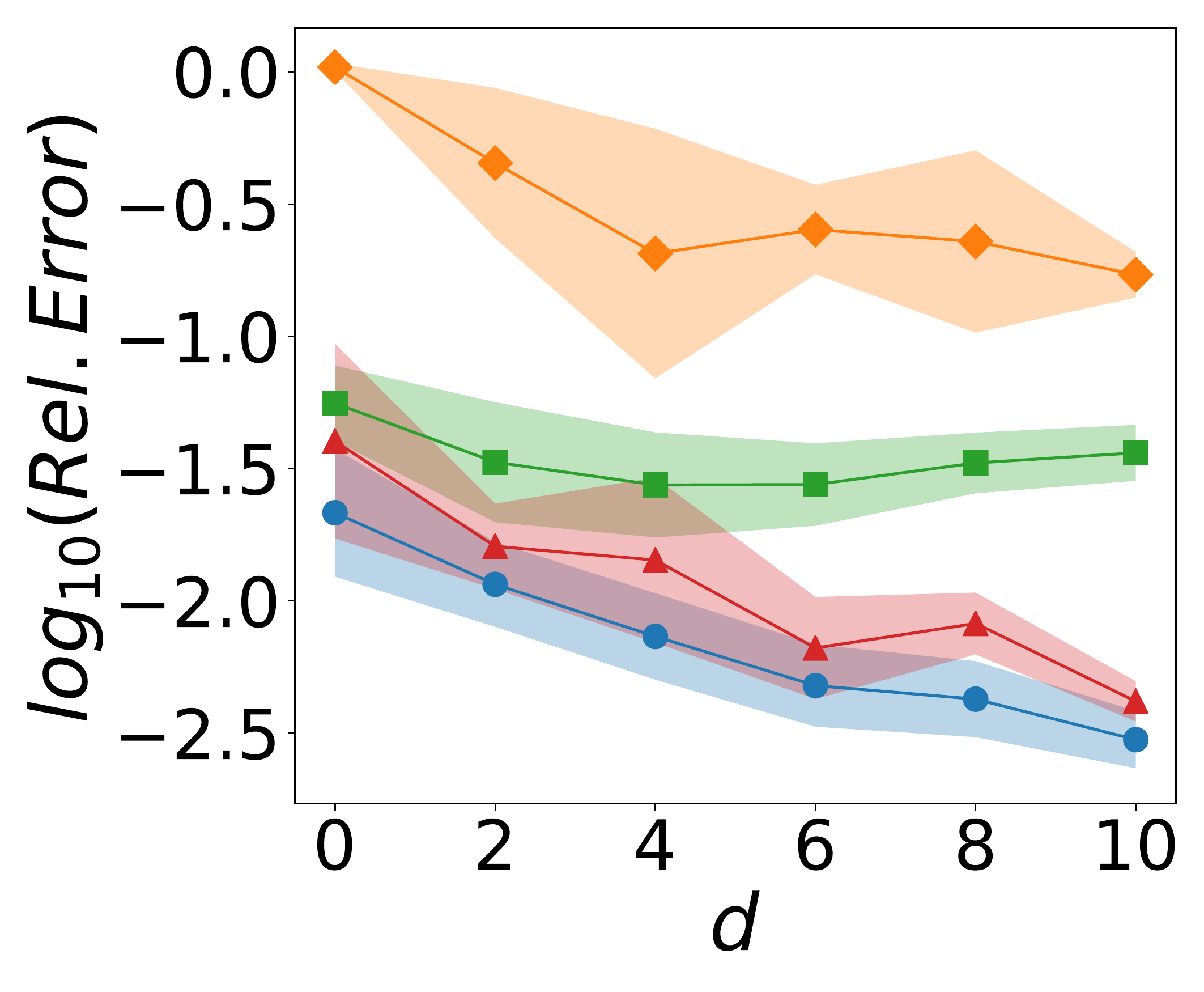}
        % \caption{ $n=1000,\\d_2=2,~\rho_{\rm noise}^2=0.1$}
    \end{subfigure}
    \begin{subfigure}{0.24\linewidth}
        \includegraphics[width=\linewidth]{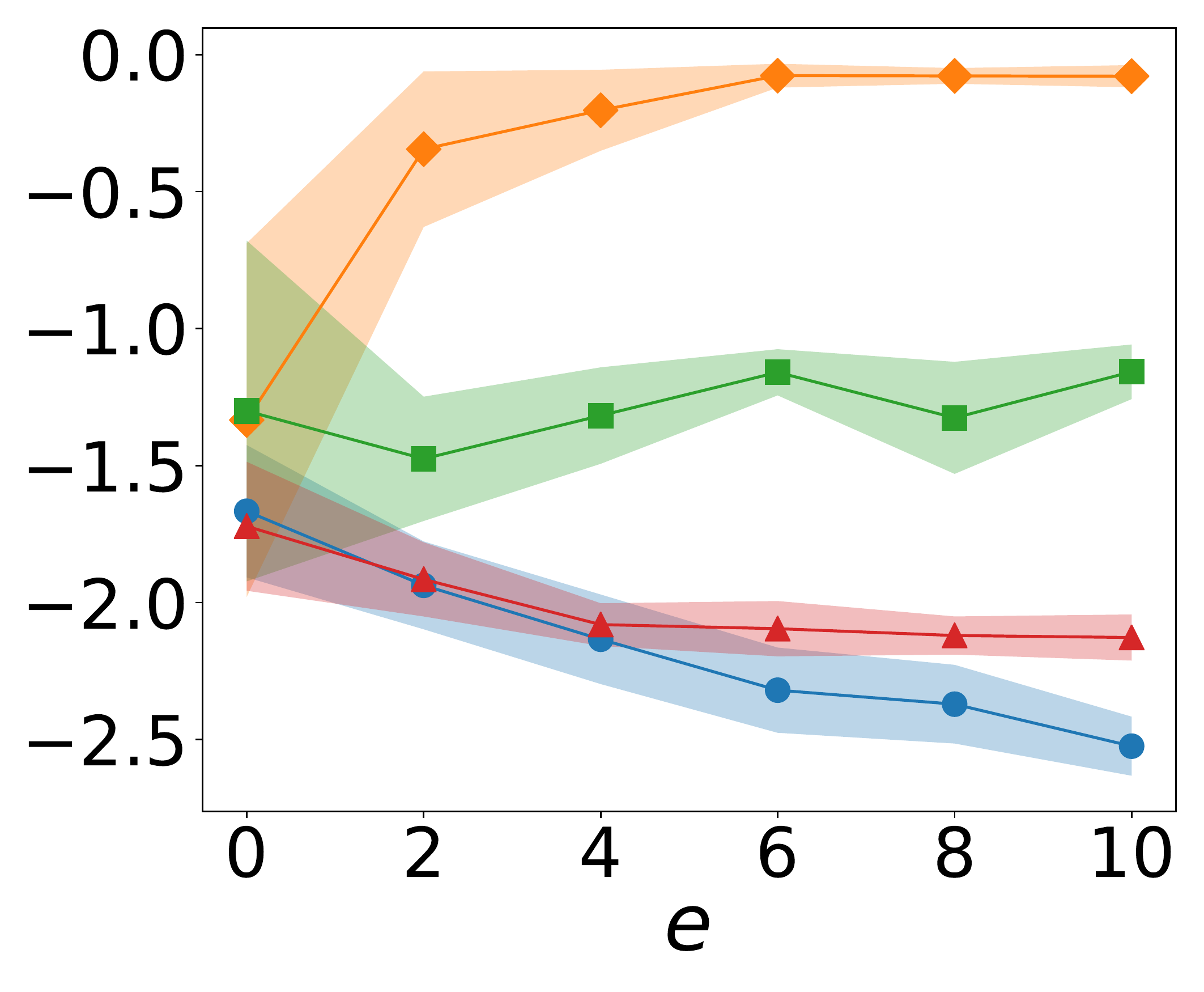}
        % \caption{$n=1000,\\d_1=2,~\rho_{\rm noise}^2=0.1$}
    \end{subfigure}
    % \vfill
    \begin{subfigure}{0.24\linewidth}
        \includegraphics[width=\linewidth]{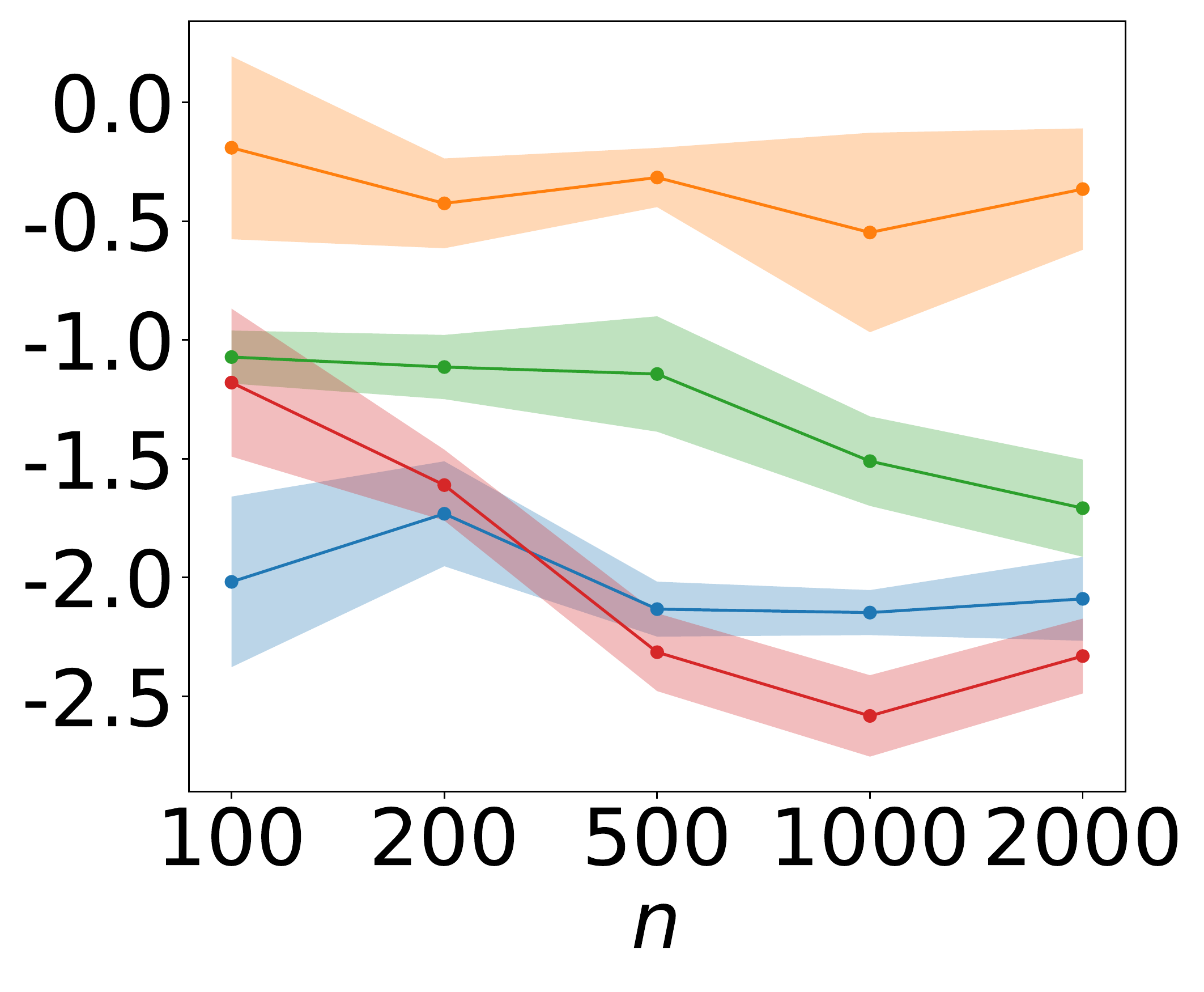}
        % \caption{ $n=1000,\\d_1=2,~d_2=3$}
    \end{subfigure}
    \begin{subfigure}{0.24\linewidth}
        \includegraphics[width=\linewidth]{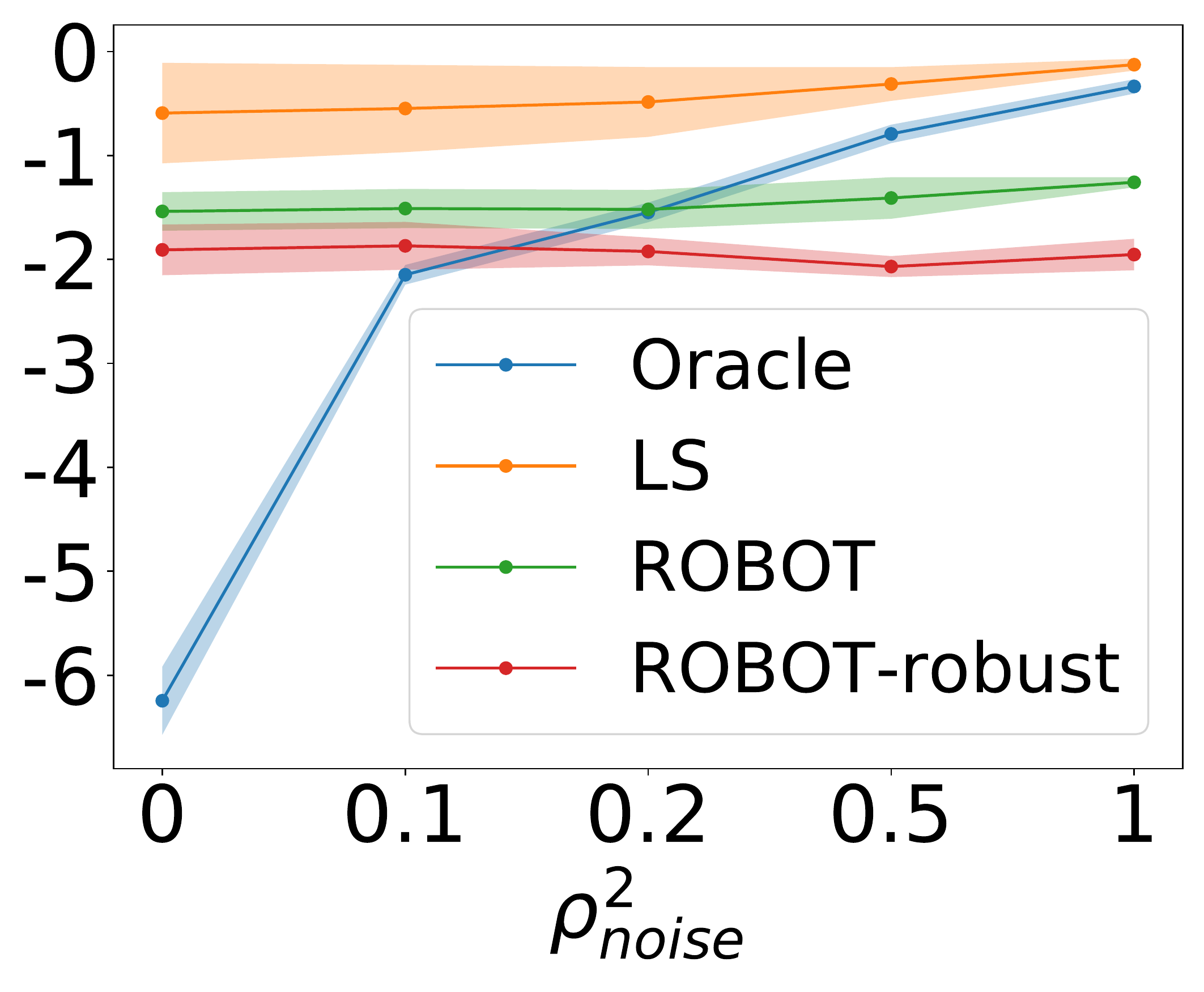}
        % \caption{ $d_1=2,~d_2=3,\\ \rho_{\rm noise}^2=0.1$}
    \end{subfigure}
    % \vspace{-5pt}
    \vskip -10pt
    % \caption{Fitting error with nonlinear regression model.}
    \caption{\textit{Nonlinear regression. We use $n=1000$, $d=2$, $e=3$, $\rho_{\rm noise}^2=0.1$ as defaults.}}
    \label{fig:synthetic_nonlinear}
    \vspace{-10pt}
\end{figure}

\noindent \textbf{Data Generation}. We mimic the scenario where the dataset is collected from different platforms.
Specifically, we generate $n$ data points $\{(y_i, [x_i, z_i])\}_{i=1}^n$, where $x_i \in \mathbb{R}^{d}$ and $z_i \in \mathbb{R}^{e}$. 
We first generate $x_i\sim \cN(\bm{0}_{d}, \bm{I}_{d})$, $z_i \sim \cN(\bm{0}_{e}, \bm{I}_{e})$, $w\sim \cN(\bm{0}_{d+e}, \bm{I}_{d+e})$, and $\varepsilon_i \sim \cN(0, \rho^2_{\rm noise})$. Then we compute $y_i = f([x_i, z_i]; w)+\varepsilon_i$.
Next, we randomly permute the order of  $\{z_i\}$ so that we lose the data correspondence. Here, $\cD_1=\{(x_i, y_i)\}$ and $\cD_2=\{z_j\}$ mimic two parts of data collected from two separate platforms. Since we are interested in the response on platform one, we treat all data from platform two, i.e., $\cD_2$, as well as $80\%$ of data in $\cD_1$ as the training data. The remaining data from $\cD_1$ are the test data. Notice that we have different number of data on $\cD_1$ and $\cD_2$, i.e., the correspondence is not exactly one-to-one.

\vspace{10pt}
\noindent \textbf{Baselines and Training}. We consider a nonlinear function $f(X, Z; w) = \sum_{k=1}^{d} \sin{([X, Z]_k w_k)}$. In this case, we consider only two baselines --- Oracle and LS, since the other baselines in the previous section are designed for linear models.
We evaluate the regression models by the transport cost divided by $\sum_i (y_i-\bar{y})^2$ on the test set.

\vspace{10pt}
\noindent \textbf{Results}. 
% The methods compared here are \textit{Oracle}, \textit{LS}, \textit{ROBOT}, and \textit{ROBOT-robust}. 
As shown in Figure \ref{fig:synthetic_nonlinear}, ROBOT-robust consistently outperforms ROBOT and LS, demonstrating the effectiveness of our robust formulation.
Moreover, ROBOT-robust achieves better performance than Oracle when the number of training data is large or when the noise level is high.

\begin{figure}[htb!]
\centering
%\vspace{-5pt}
\begin{subfigure}{0.15\linewidth}
\includegraphics[width=0.95\linewidth]{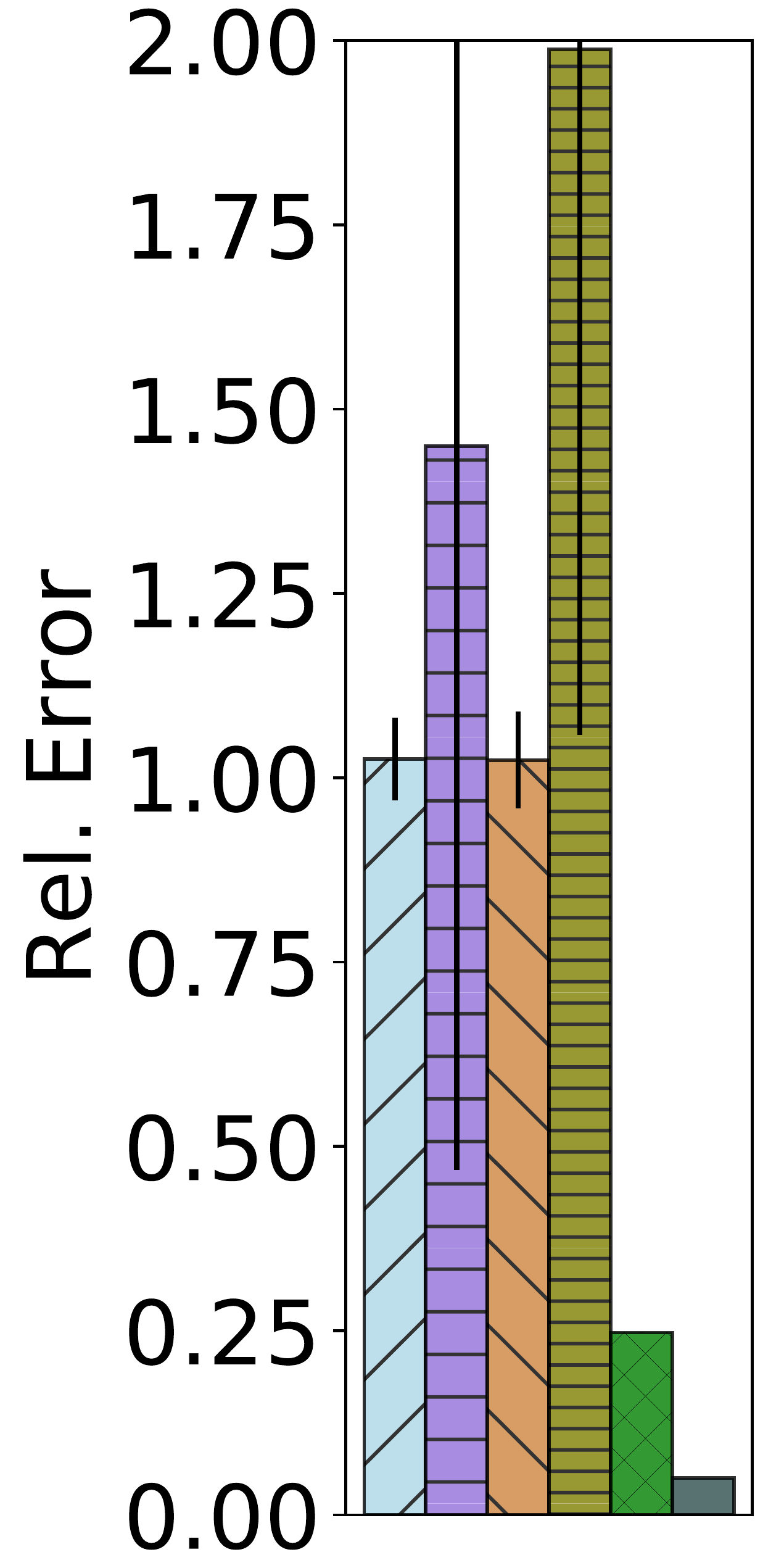}
%\vspace{-3pt}
\caption{{\textit{FC}}}
\end{subfigure}
\begin{subfigure}{0.31\linewidth}
\vspace{2pt}
\includegraphics[width=0.95\linewidth]{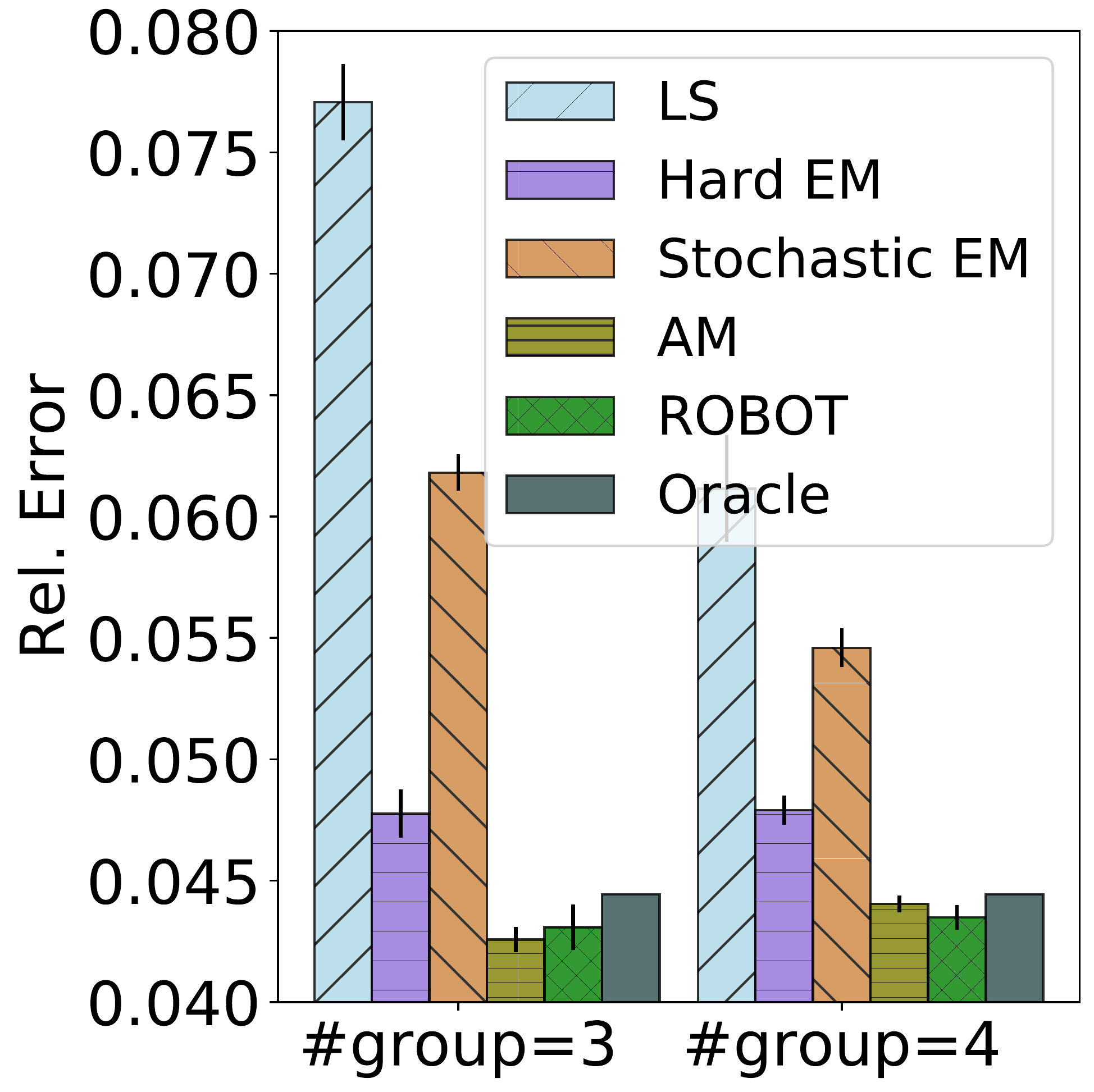} 
%\vspace{-3pt}
\caption{{\textit{GFC}}}
\end{subfigure}
%\vspace{-5pt}
\caption{\label{fig:flow} \textit{Relative error of different methods.}}
%\vspace{-20pt}
\end{figure}

%\vspace{-0.13in}
\subsection{Flow Cytometry }
%\vspace{-0.08in}
\label{sec:sec53}
% In this section we apply ROBOT to Flow Cytometry (FC) and Gated Flow Cytometry (GFC).
In flow cytometry (FC), a sample containing particles is suspended in a fluid and injected into the flow cytometer, but the measuring instruments are unable to preserve the correspondence between the particles and the measurements. Different from FC, gated flow cytometry (GFC) uses “gates” to sort the particles into one of many bins, which provides partial ordering information since the measurements are provided individually for each bin. In practice, there are usually $3$ or $4$ bins.
% Following \citet{abid2018stochastic}, here we want to characterize affinity of aptamers to a particular target, and to determine the dependence of affinity on the nucleotide motifs present in aptamers. 

% Flow cytometry was introduced in Section I. In this subsection, we consider a modification referred to as gated flow cytometry: First, particles are suspended in fluid and flowed through a cytometer; then, the cytometer analyzes the properties of each particle, and uses “gates” to sort the particles into one of many bins. This sorting provides partial ordering information, as it provides the analyst a list of labels for all of the particles in each bin. With enough bins, it would be possible to determine the ordering information completely, but generally, it is only practical to set up 3-4 bins.
% Gated flow cytometry has many use cases, but the par- ticular application we consider is measurement of aptamer affinity. Aptamers are short DNA sequences that bind to target molecules, and it is of interest to characterize affinity of aptamers to a particular target and to determine the dependence of affinity on the nucleotide motifs (e.g. “AGG” or “CC”) present in aptamers. 

% \textbf{Settings}. We adopt dataset from \citet{knight2009array}, and perform experiment with the same settings as \citet{abid2018stochastic}. 
\vspace{10pt}
\noindent \textbf{Settings}. We adopt the dataset from \citet{knight2009array}. Following \citet{abid2017linear},
the outputs $y_i$'s are normalized, and we select the top $20$ significant features by a linear regression on the top $1400$ items in the dataset.
% we lower the dimension of the dataset by choosing the $20$ most significant features computed from an initial run of standard linear regression and restrict the dataset to the top $1400$ items, to create a more homogeneous and linear dataset.
% We then normalize the output $y^{(i)}$.
We use $90\%$ of the data as the training data, and the remaining as test data. For ordinary FC, we randomly shuffle all the labels in the training set.
For GFC, the training set is first sorted by the labels, and then divided into equal-sized groups, mimicking the sorting by gates process. The labels in each group are then randomly shuffled. To simulate gating error, $1\%$ of the data are shuffled across the groups. 
We compare ROBOT with \text{Oracle}, \text{LS}, \text{Hard EM} (a variant of \text{Stochastic EM} proposed in \citet{abid2018stochastic}), \text{Stochastic EM}, and \text{AM}.
We use relative error on the test set as the evaluation metric.
% For evaluation of the regression models, we use the regression coefficients to make affinity predictions in the test set and record the RSS/TSS error.

\vspace{10pt}
\noindent \textbf{Results}. As shown in Figure \ref{fig:flow}, while AM achieves good performance on GFC when the number of groups is 3, it behaves poorly on the FC task. ROBOT, on the other hand, is efficient on both tasks.
\subsection{Multi-Object Tracking}
%\vspace{-0.08in}
\label{sec:sec54}

% \begin{wrapfigure}{r}{5.5cm}
% \centering
% \vspace{-10pt}
% \includegraphics[width=.85\linewidth]{figures/real/mot20.png}  
% \caption{\label{fig:illu_most20} One frame in \texttt{MOT20}  with detected bounding boxes in yellow.}
% \vspace{-15pt}
% \end{wrapfigure}

% \begin{figure}[!t]
%     % \centering
%     \begin{minipage}[t]{0.22\textwidth}
%     \vspace{10pt}
%     \includegraphics[width=.99\linewidth]{figures/real/mot20.png}  
% \caption{\label{fig:illu_most20} \textit{One frame in \texttt{MOT20}  with detected bounding boxes in yellow.}}
% \end{minipage}
% \quad
% \begin{minipage}[c]{0.75\textwidth}
%     \vspace{-20pt}
\begin{table}[t]
\centering
\caption{\label{tab:mot}\textit{Experiment results on MOT. Here, $\uparrow$ suggests the larger the better, and $\downarrow$ suggests the smaller the better.}}
% \resizebox{0.75\textwidth}{!}{%
\begin{tabular}{
@{\hspace{1pt}}l
@{\hspace{8pt}}l
@{\hspace{8pt}}l
@{\hspace{3pt}}l
@{\hspace{3pt}}l
@{\hspace{5pt}}l
@{\hspace{5pt}}l
@{\hspace{5pt}}l
@{\hspace{5pt}}l
@{\hspace{5pt}}l
@{\hspace{3pt}}}
\hline
Data & Method        &MOTA$\uparrow$ & MOTP$\uparrow$ & IDF1$\uparrow$ & MT$\uparrow$ & ML$\downarrow$& FP$\downarrow$ & FN$\downarrow$ & IDS$\downarrow$ \\ \hline
\multirow{2}{*}{\texttt{MOT17} (train)}  & ROBOT &  \textbf{48.3} & 82.6 & 55.3 & 407   & 553 &  22,443  &  149,988  & 1,811    \\
& w/o ROBOT    & 44.0 &81.3  &49.9 & 404 & 550   & 36,187   &  149,131 & 3,204  \\
 \hline
\multirow{3}{*}{\texttt{MOT17} (dev)} &  ROBOT & \textbf{48.2} & 76.6 &  43.4    & 455   &904& 29,419 &  259,714  &  3,228   \\
 & w/o ROBOT    & 42.1 &75.0  &36.8 & 414 & 890   & 61,210   &  259,318 & 6,138  \\
  &  SORT & 43.1 & 77.8 &  39.8    & 295   &997& 28,398 &  287,582  &  4,852    \\
% &  SiamRPN+ROBOT+train &      &      &      &    &    &    &    &     \\
 \hline
\multirow{2}{*}{\texttt{MOT20} (train)} &  ROBOT &  \textbf{56.2} & 84.9 &  47.6    & 805   &  288  &  113,752  &  377,247  & 5,888    \\
& w/o ROBOT   & 48.8 &81.5  & 40.2 & 769 &  290  & 186,245   &  384,562 & 10,153  \\
 \hline
\multirow{3}{*}{\texttt{MOT20} (dev)} &  ROBOT & \textbf{45.0} & 76.9 & 34.0 & 394  &257 & 70,416 & 210,425   &  3,683 \\
& w/o ROBOT   &  38.5 &75.1  & 27.0 & 383 & 233  & 104,958  & 207,627 & 5,696  \\
&  SORT & 42.7 & 78.5 & 45.1 & 208  &326 & 27,521 & 264,694   &  4,470 \\
% &  SiamRPN+ROBOT+train &      &      &      &    &    &    &    &     \\
\hline 
\end{tabular}
% }
%\vspace{-10pt}
\end{table}
% \end{minipage}
%     \vspace{-10pt}
% \end{figure}

\begin{figure}[htb!]
\centering
%\vspace{-10pt}
\includegraphics[width=.7\linewidth]{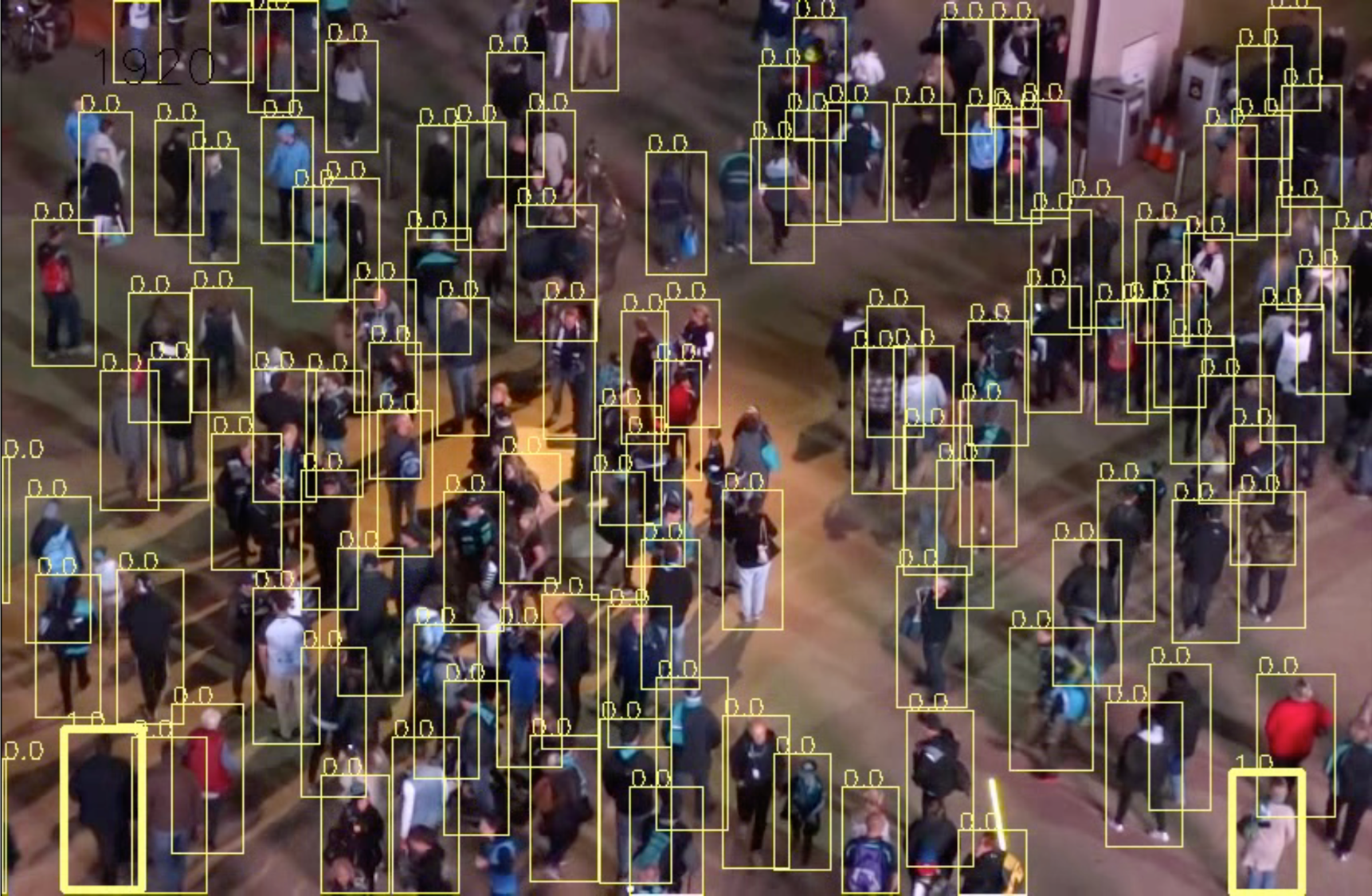}  
%\vspace{-2pt}
\caption{\label{fig:illu_most20} \textit{One frame in \texttt{MOT20}  with detected bounding boxes in yellow.}}
%\vspace{-13pt}
\end{figure}
In this section we extend our method to vision-based Multi-Object Tracking (MOT), a task with broad applications in mobile robotics and autonomous driving. Given a video and the current frame, the goal of MOT is to predict the locations of the objects in the next frame. Specifically, object detectors \citep{felzenszwalb2009object, ren2015faster} first provide us the potential locations of the objects by their bounding boxes. Then, MOT aims to assign the bounding boxes to trajectories that describe the path of individual objects over time. Here, we formulate the current frame and the objects' locations in the current frame as $\cD_2=\{z_j\}$, while we treat the next frame and the locations in the next frame as $\cD_1=\{(x_i, y_i)\}$.

Existing deep learning based MOT algorithms require large amounts of annotated data, i.e., the ground truth of the correspondence, during training. Different from them, our algorithm does not require the correspondence between $\cD_1$ and $\cD_2$, and all we need is the video. This task is referred to as \textit{unsupervised MOT} \citep{he2019tracking}.

\vspace{10pt}
\noindent \textbf{Related Works}. To the best of our knowledge, the only method that accomplishes unsupervised end-to-end learning of MOT is \citet{he2019tracking}. However, it targets tracking with low densities, e.g., \texttt{Sprites-MOT}, which is different from our focus.

\vspace{10pt}
\noindent \textbf{Settings}. We adopt the \texttt{MOT17} \citep{MOT16} and the \texttt{MOT20} \citep{MOTChallenge20} datasets. Scene densities of the two datasets are $31.8$ and $170.9$, respectively, which means the scenes are pretty crowded as we illustrated in Figure \ref{fig:illu_most20}.
% We use the training data of \texttt{MOT17} to pick the hyper-parameters.
We adopt the DPM detector \citep{felzenszwalb2009object} on \texttt{MOT17} and the Faster-RCNN detector \citep{ren2015faster} on \texttt{MOT20} to provide us the bounding boxes. Inspired by \citet{xu2019deepmot}, the cost matrix is computed as the average of the Euclidean center-point distance and the Jaccard distance between the bounding boxes,
\begin{align*}
    C_{ij}(w) = \frac{1}{2}\left( \frac{\norm{c(f(z_j; w))-c(y_i)}_2}{\sqrt{H^2+W^2}} + \cJ(f(z_j; w), y_i)\right), 
\end{align*}
where $c(\cdot)$ is the location of the box center, $H$ and $W$ are the height and the width of the video frame, and $\cJ(\cdot, \cdot)$ is the Jaccard distance defined as $1$-IoU (Intersection-over-Union). We utilize the single-object tracking model SiamRPN\footnote{The initial weights of $f$ are obtained from \texttt{https://github.com/foolwood/DaSiamRPN}.} \citep{li2018high} as our regression model $f$. We apply ROBOT-robust with $\rho_1=\rho_2=10^{-3}$.
% To track the birth and death of the tracks, we adapt the inference code in \citet{xu2019deepmot}.
See Appendix \ref{sec:appendix_exp} for more detailed settings.

\vspace{10pt}
\noindent \textbf{Results}. We demonstrate the experiment results in Table \ref{tab:mot}, where the evaluation metrics follow \citet{ristani2016performance}. In the table,  $\uparrow$ represents the higher the better, and $\downarrow$ represents the lower the better.
ROBOT signifies the model trained by ROBOT-robust, and w/o ROBOT means the pretrained model in \citet{li2018high}.  The scores are improved significantly after training with ROBOT-robust. 

We also include the scores of the SORT model \citep{bewley2016simple} obtained from the dataset platform. Different from SiamRPN and SiamRPN+ROBOT, SORT is a supervised learning model. As shown, our unsupervised training framework achieves comparable or even better performance.

% \subsection{Possible Datasets}

% Synthetic dataset

% High dim synthetic dataset: Face pose estimation (Biwi), facial landmark (FLD).

% Real applications, tabular dataset:

% \cite{varol2019robust}, C. elegans fluorescence imaging dataset,  \texttt{http://dx.doi.org/10.21227/H2901H}

% \cite{abid2017linear, abid2018stochastic}, aptamer evolution, public but link not found yet

% \cite{abid2018stochastic}, Partially Anonymized Housing Prices

% Survey on Household Income and Wealth 

% NC voter registration data

% Data from the Syrian conflict

% Real applications, larger dataset: simultaneous location and mapping (SLAM)? assigning observations to targets in multi-target tracking problems (\texttt{http://cs.binghamton.edu/~mrldata/pets2009})? mobile sensing scheme? clock jitter?

% \citep{david2004softposit}

%!TEX root = ../robot_tech.tex
%\vspace{-0.15in}
\section{Discussion}
%\vspace{-0.1in}
\label{sec:diss}

% \begin{figure}[!t]
%     % \centering
%     % \vspace{-0.25in}
% \begin{minipage}[b]{0.26\textwidth}
%     \centering
%     \hspace{-10pt}
%      \includegraphics[width=.99\linewidth]{figures/synthetic_sensing/partial_init_compare.pdf}  
%     \caption{\label{fig:illu_init} \textit{Errors with different training seed.}}
% \end{minipage}
% \quad
% \begin{minipage}[b]{0.39\textwidth}
%     \centering
%     % \vspace{10pt}
%      \begin{subfigure}{0.49\linewidth}
%         \includegraphics[width=\linewidth]{figures/synthetic/loss_demo.png}
%         \caption{$\cL(w)$\label{fig:illu_lossa}}
%     \end{subfigure}
%     \begin{subfigure}{0.49\linewidth}
%         \includegraphics[width=\linewidth]{figures/synthetic/loss_eps_demo.png}
%         \caption{$\cL_{\epsilon}(w)$\label{fig:illu_lossb}}
%     \end{subfigure}
%     \caption{\label{fig:illu_loss} \textit{How entropy helps escape local optima. Here, $\epsilon=10^{-2}$.}}
% \end{minipage}
% \quad
% \begin{minipage}[b]{0.28\textwidth}
%     % \vspace{10pt}
%     \centering
%      \includegraphics[width=.85\linewidth]{figures/synthetic/robust_demo_kl.png}
%     \caption{\label{fig:illu_rot_kl} \textit{Illustration with KL divergence.}}
% \end{minipage}
% \vspace{-20pt}
% \end{figure}

% \begin{wrapfigure}{r}{3.5cm}
% \centering
% \vspace{-15pt}
% \includegraphics[width=.5\linewidth]{figures/synthetic_sensing/partial_init_compare.pdf}  
% \caption{\label{fig:illu_init} Errors with different training seed.}
% \vspace{-20pt}
% \end{wrapfigure}

\noindent $\bullet$ \textbf{Sensitivity to initialization}. As stated in \citet{pananjady2017linear}, obtaining the global optima of (\ref{eq:ls}) is in general an NP-hard problem. Some ``global'' methods methods use global optimization techniques and have exponential complexity, e.g., \citet{Elhami2017unlabeled}, which is not applicable to large data. The other ``local'' methods only guarantee converge to local optima, and the convergence is very sensitive to initialization. Compared with existing ``local'' methods, our method is computationally efficient and greatly reduces the sensitivity to initialization.
% As most of the existing works adopt \eqref{eq:ls} as the objective function, the works with global optima guarantees generally have exponential complexity, e.g., \citet{Elhami2017unlabeled}.
% Similar to many existing works \citep{abid2017linear, abid2018stochastic}, our algorithm is not guaranteed to converge to the global optima. Therefore, it is usually used with multiple start in practice. Nevertheless, the entropy regularization in \eqref{eq:sinkhorn} and \eqref{eq:robust_rwoc}, can sometimes help escape the local optima. 

\begin{figure}[htb!]
    \centering
%    \vspace{-15pt}
    \begin{subfigure}{0.3\linewidth}
        \includegraphics[height=1.\linewidth]{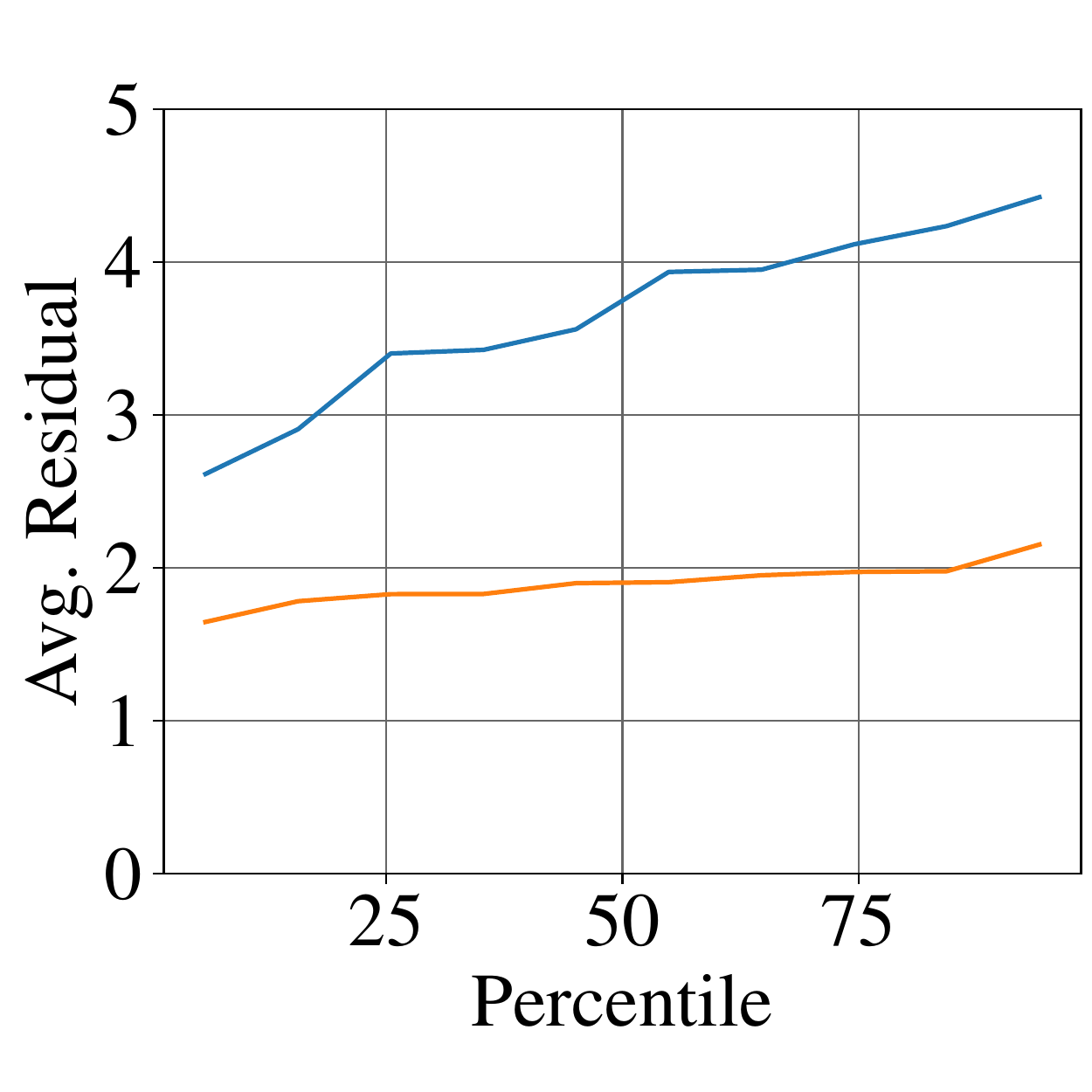}
        \vskip -5pt
        \caption{Training residual}
    \end{subfigure}
    \begin{subfigure}{0.3\linewidth}
        \includegraphics[height=1.\linewidth]{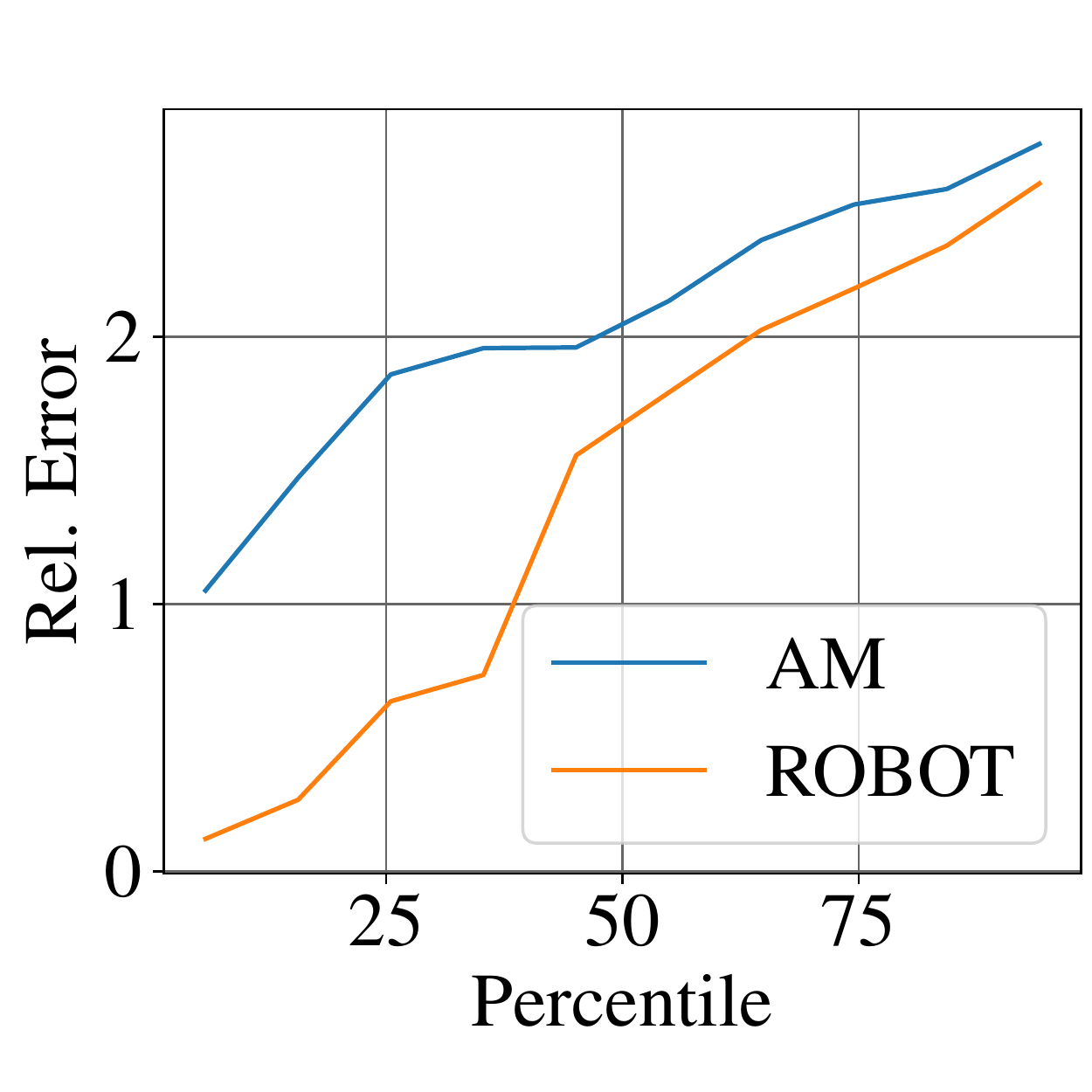}
        \vskip -5pt
        \caption{Test error}
    \end{subfigure}
    \vspace{-5pt}
    \caption{ROBOT and AM with \textit{different initial solutions.}}
    \label{fig:illu_init}
%    \vspace{-10pt}
\end{figure}

To demonstrate such an advantage, we run \text{AM} and \text{ROBOT} with $10$ different initial solutions, and then we sort the results based on (a) the averaged residual on the training set, and (b) the relative prediction error on the test set. We plot the percentiles in Figure \ref{fig:illu_init}.
Here we use fully shuffled data under the unlabeled sensing setting, and we set 
$n=1000$, $d=5$, $\rho_{\rm noise}^2=0.1$, and $\epsilon=10^{-2}$.
We can see that ROBOT is able to find ``good'' solutions in 30\% of the cases (The relative prediction error is smaller than $1$), but AM is more sensitive to the initialization and cannot find ``good'' solutions.

\vspace{10pt}

\begin{figure}[htb!]
    \centering
    \begin{subfigure}[b]{0.37\linewidth}
        \centering
        \includegraphics[width=0.99\linewidth]{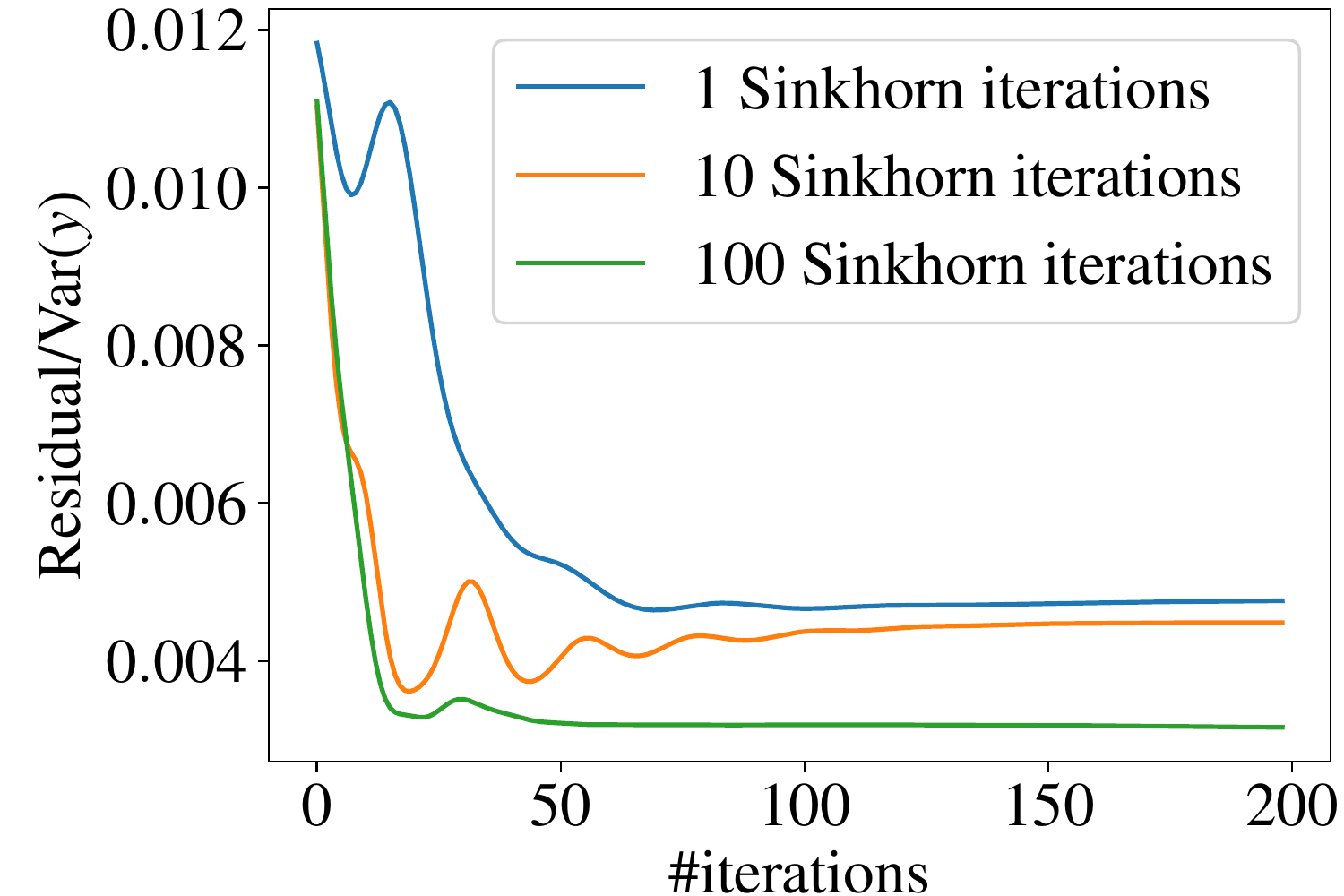}
        \vspace{-10pt}
        \caption{}
    \end{subfigure}
    \begin{subfigure}[b]{0.29\linewidth}
    \centering
        \includegraphics[height=0.99\linewidth]{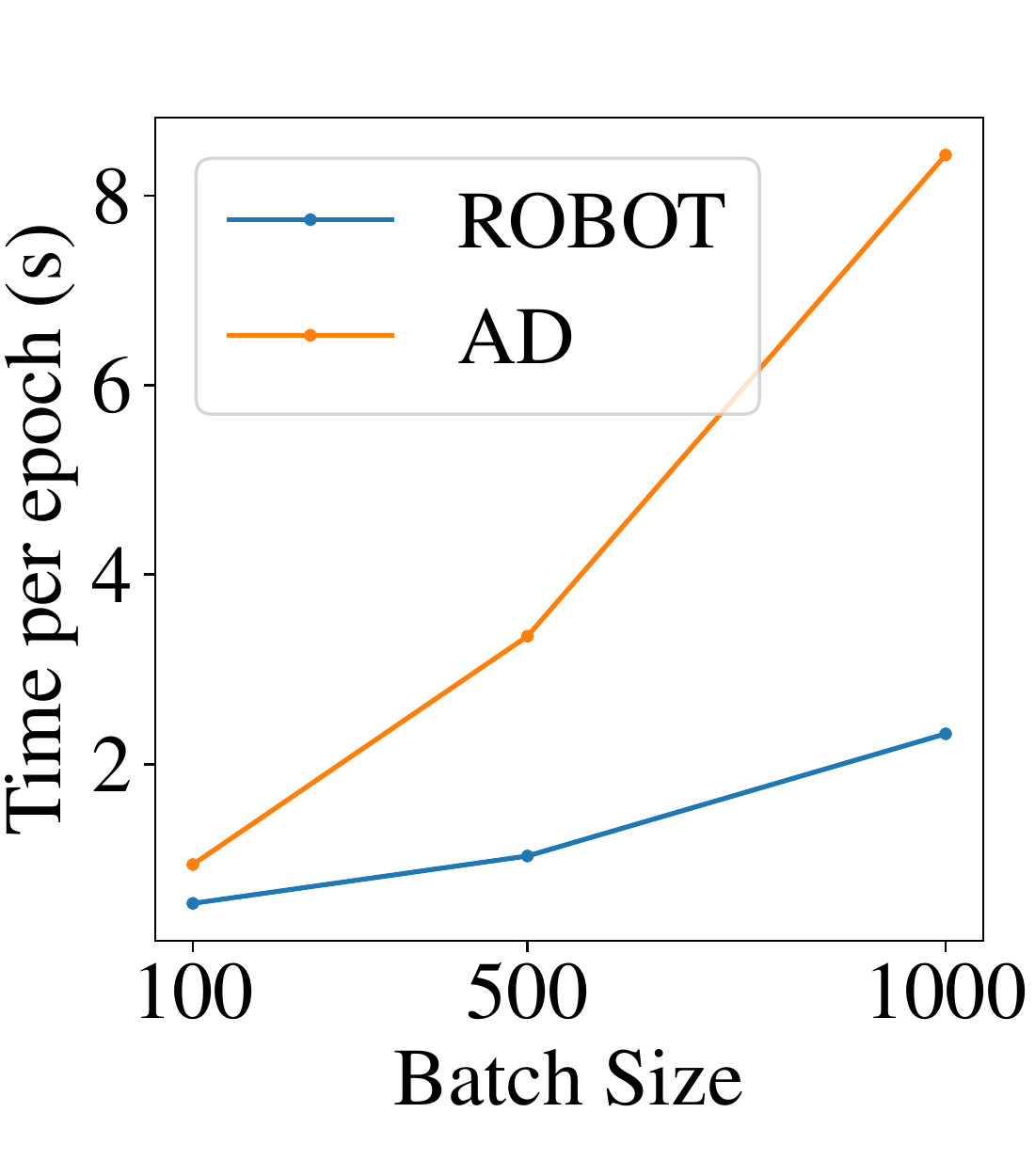}
        \vskip -5pt
        \caption{}
    \end{subfigure}
    \begin{subfigure}[b]{0.29\linewidth}
    \centering
        \includegraphics[height=0.99\linewidth]{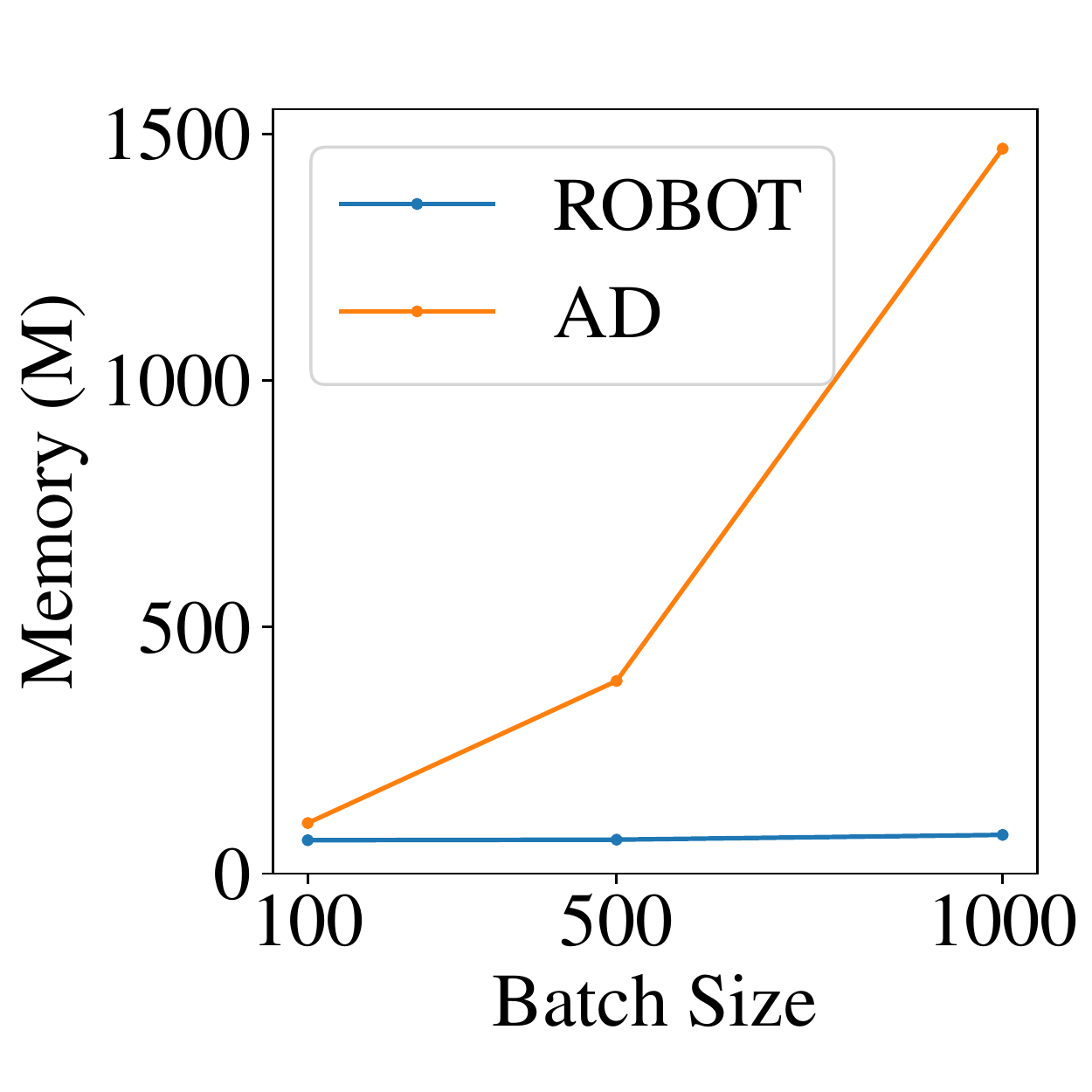}
        \vskip -5pt
        \caption{}
    \end{subfigure}
    \vspace{-5pt}
    \caption{\textit{The comparisons to AD. (a) Convergence under different number of Sinkhorn iterations of AD. (b) Time comparison. (c) Memory comparison.}}
    \label{fig:synthetic_efficiency}
\end{figure}

\noindent $\bullet$ \textbf{ROBOT v.s. Automatic Differentiation (AD).} 
Our algorithm computes the Jacobian matrix directly based on the KKT condition of the lower problem (\ref{eq:inner_sinkhorn}). An alternative approach to approximate the Jacobian is the automatic differentiation through the Sinkhorn iterations for updating $S$ when solving (\ref{eq:inner_sinkhorn}). As suggested by Figure \ref{fig:synthetic_efficiency} (a), running Sinkhorn iterations until convergence ($200$ Sinkhorn iterations) can lead to a better solution\footnote{We remark that running one iteration sometimes cannot converge.}. In order to apply AD, we need to store all the intermediate updates of all the Sinkhorn iterations. This require the memory usage to be proportional to the number of iterations, which is not necessarily affordable. In contrast, applying our explicit expression for the backward pass is memory-efficient. Moreover, we also observe that AD is much more time-consuming than our method. The timing performance and memory usage are shown in Figure \ref{fig:synthetic_efficiency} (b)(c), where we set $n=1000$. %Note that for robust RWOC, AD is not applicable to compute the Jacobian, since the forward pass is not necessarily differentiable. 

\begin{figure}[t]
    \centering
        \includegraphics[width=0.3\linewidth]{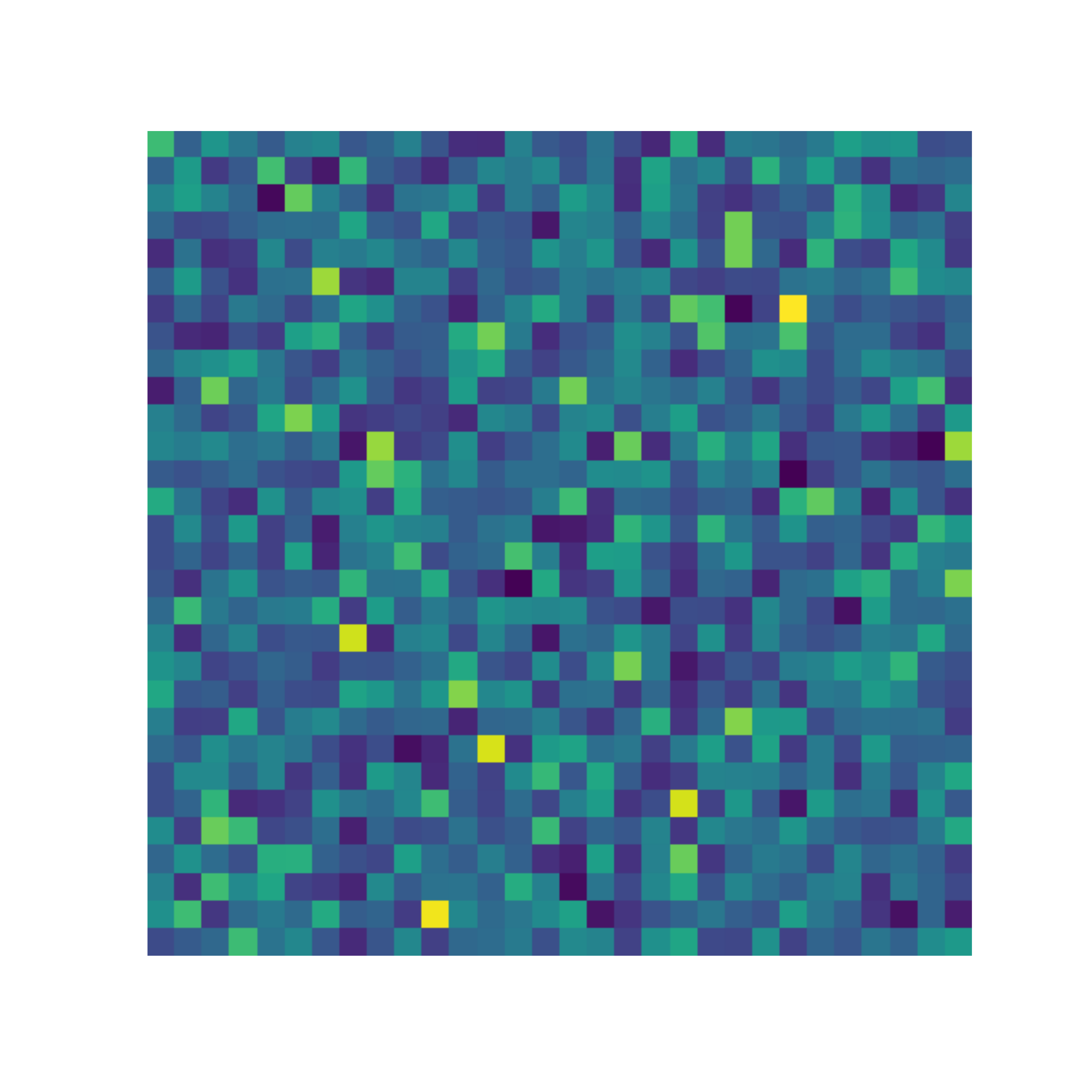}
        \vspace{-5pt}
        \caption{\textit{Expected correspondence in EM.}}
    \label{fig:em_permutation}
    \vspace{-10pt}
\end{figure}

\vspace{10pt}

\noindent $\bullet$ \textbf{Connection to EM.} \citet{abid2018stochastic} adopt an Expectation Maximization (EM) method for RWOC, where $S$ is modeled as a latent random variable. Then in the M-step, one maximizes the expected likelihood of the data over $S$. This method shares the same spirit as ours: We avoid updating $w$ using one single permutation matrix like AM. However, this method is very dependent on a good initialization. Specifically, if we randomly initialize $w$, the posterior distribution of $S$ in this iteration would be close to its prior, which is a uniform distribution. In this way, the follow-up update for $w$ is not informative. Therefore, the solution of EM would quickly converge to an undesired stationary point. Figure \ref{fig:em_permutation} illustrates an example of converged correspondence, where we adopt $n=30, o=e=1, d=0$.
For this reason, we initialize EM with good initial points, either by RS or AM throughout all experiments. 

\begin{table}[htb!]
\caption{\textit{Pairwise comparisons between RS alone and the combination of RS and ROBOT. The relative error ratio is the ratio of the relative errors of RS alone and RS+ROBOT combination. Ratios larger than $1$ suggest that RS performs worse than RS+ROBOT combination. }}\label{tab:ransac}
\vspace{-10pt}
\centering
\begin{tabular}{cccc}\\\toprule  
Proportion & $25\%$ & $50\%$ & $75\%$ \\\midrule
Rel. error ratio & $1.04\pm 0.20$ & $1.29\pm 0.32$ & $1.27\pm 0.34$ \\  \midrule
\end{tabular}
\vspace{-5pt}
\end{table} 

\noindent $\bullet$ \textbf{Combination with RS}.
 As suggested in Figure \ref{fig:synthetic_sensing_partial}, although RS cannot perform well itself, retraining the output of RS using our algorithms increases the performance by a large margin. To show that combining RS and ROBOT can achieve better results than RS alone, we compare the following two cases: i). Subsample $2\times10^5$ times using RS; ii). Subsample $10^5$ times using RS followed by ROBOT for $50$ training steps. The result is shown in Table \ref{tab:ransac}. For a larger permutation proportion, RS alone cannot perform as well as RS+ROBOT combination. Here, we have 10 runs for each proportion. We adopt \texttt{SNR}$=100$, $d=5$ for data, and $\epsilon=10^{-4}$, learning rate $10^{-4}$ for ROBOT training.

\vspace{10pt}

\noindent $\bullet$ \textbf{Related works with additional constraints.} There is another line of research which improves the computational efficiency by solving variants of RWOC with additional constraints. Specifically, \cite{haghighatshoar2017signal, Rigollet2018Uncoupled} assume an isotonic function (note that such an assumption may not hold in practice), and \cite{Shi2018Spherical, Slawski2019Linear, Slawski2019Two, slawski2019sparse, varol2019robust} assume only a small fraction of the correspondence is missing. Our method is also applicable to these problems, as long as the additional constraints can be adapted to the implicit differentiation step.

\vspace{10pt}

\noindent $\bullet$ \textbf{More applications of RWOC.} RWOC problems generally appear for two reasons. First, the measuring instruments are unable to preserve the correspondence. In addition to GFC and MOT, we list a few more examples: 
SLAM tracking \citep{thrun2007simultaneous},
archaeological measurements \citep{robinson1951method},
large sensor networks \citep{keller2009identity}, 
 pose and correspondence estimation \citep{david2004softposit}, and
the genome assembly problem from shotgun reads \citep{huang1999cap3}. Second, the data correspondence is masked for privacy reasons. For example, we want to build a recommender system for a new platform, borrowing user data from a mature platform.

% {\color{red} TO DO:  double check the labels}
% \input{tex_files/broader}

\bibliography{reference}
\bibliographystyle{ims}
\newpage
\appendix
\section{Connection between OT and RWOC}
\label{sec:appendix_connection}

\textbf{Theorem 1.} Denote $\Pi(a, b)=\{S\in\RR^{n\times m}: S\bm{1}_m = a, S^\top\bm{1}_n = b, S_{ij} \geq 0\}$ for any $a\in \RR^n$ and $b\in \RR^m$. Then at least one of the optimal solutions of the following problem lies in $\cP$.
\begin{align} \label{eq:app_convex_relax}
    \sideset{}{_{S\in\RR^{n\times n}}}\min \langle C(w),S\rangle,  ~~\textrm{s.t.}~S\in\Pi(\bm{1}_n, \bm{1}_n).
\end{align}

\begin{proof}
Denote the optimal solution of (\ref{eq:app_convex_relax}) as $Z^*$.
As we mentioned earlier, this is a direct corollary of Birkhoff–von Neumann theorem \citep{birkhoff1946three, von1953certain}. Specifically, Birkhoff–von Neumann theorem claims that the polytope $\Pi(\bm{1}_n, \bm{1}_n)$ is the convex hull of the set of $n\times n$ permutation matrices, and furthermore that the vertices of $\Pi(\bm{1}_n, \bm{1}_n)$ are precisely the permutation matrices.

On the other hand, (\ref{eq:app_convex_relax}) is a linear optimization problem. There would be at least one optimal solutions lies at the vertices given the problem is feasible. As a result, there would be at least one $Z^*$ being a permutation matrix.
\end{proof}

\section{Proof of Proposition \ref{better-cond}}
\label{sec:fast_convergence}

The bilevel optimization formulation has a better gradient descent iteration complexity than alternating minimization. To see this, consider a quadratic function $F(a_1,a_2)= a^\top P a + b^\top a$, where $a_1\in \RR^{d_1}$, $a_2\in \RR^{d_2}$, $a=[a_1^\top, a_2^\top]^\top\in \RR^{(d_1+d_2)}$, $P \in \RR^{(d_1+d_2)\times (d_1+d_2)}$, $b \in \RR^{(d_1+d_2)}$. To further simplify the discussion, we assume $P = \rho \bm{1}_{(d_1+d_2)} \bm{1}_{(d_1+d_2)}^\top + (1-\rho) I_{d_1+d_2}$, where $I_{d_1+d_2}$ is the identity matrix. Then we have the following proposition. 

\noindent \textbf{Proposition 1.} 
Given $F$ defined in (\ref{qp-bi-level}), we have
\begin{align*}
\frac{\lambda_{\max}(\nabla^2 F(a_1))}{\lambda_{\min}(\nabla^2 F(a_1))} = 1+\frac{1-\rho+\lambda}{d_2 \rho -\rho+\lambda +1}\cdot\frac{d_1 \rho}{1-\rho}\quad\textrm{and}\quad \frac{\lambda_{\max}(\nabla^2_{a_1a_1} L(a_1,a_2))}{\lambda_{\min}(\nabla^2_{a_1a_1} L(a_1,a_2))} = 1+\frac{d_1 \rho}{1-\rho}.
\end{align*}

\begin{proof}
For alternating minimization, the Hessian for $a_1$ is a submatrix of $P$, i.e.,
\begin{align*}
    H_{\rm AM} = \rho \bm{1}_{d_1} \bm{1}_{d_1}^\top + (1-\rho) I_{d_1},
\end{align*}
whose condition number is 
\begin{align*}
    C_{\rm AM} = 1+\frac{d_1\rho}{1-\rho}.
\end{align*}
We now compute the condition number for ROBOT. Denote 
\begin{align*}
    P = \begin{bmatrix}
     P_{11} & P_{12} \\
     P_{21} & P_{22}
    \end{bmatrix}, \quad
    b = \begin{bmatrix}
     b_1 \\
     b_2
    \end{bmatrix}, 
\end{align*}
where $P_{11} \in \RR^{d_1\times d_1}$, $P_{12} \in \RR^{d_1\times d_2}$, $P_{21} \in \RR^{d_2\times d_1}$,  $P_{22} \in \RR^{d_2\times d_2}$, and $b_1\in \RR^{d_1}$, $b_2\in \RR^{d_2}$. ROBOT first minimize over $a_2$,
\begin{align*}
    a^*_2(a_1) = \argmin_{a_2} F(a_1,a_2) = - (P_{22}+\lambda I_{d_2})^{-1} (P_{21}a_1 + b_2/2).
\end{align*}
Substituting $a^*_2(a_1)$ into $F(a_1,a_2)$, we can obtain the Hessian for $a_1$ is
\begin{align*}
    H_{\rm ROBOT} = P_{11} - P_{12}(P_{22}+\lambda I_{d_2})^{-1} P_{21}.
\end{align*}
Using Sherman--Morrison formula, we can explicitly express $P_{22}^{-1}$ as 
\begin{align*}
    P_{22}^{-1} = \frac{1}{1-\rho+\lambda} I_{d_2} - \frac{\rho}{(1-\rho+\lambda)(1-\rho +\lambda + \rho d_2)} \bm{1}_{d_2} \bm{1}^\top_{d_2}.
\end{align*}
Substituting it into $H_{\rm ROBOT}$, 
\begin{align*}
    H_{\rm ROBOT} = P_{11} - P_{12}P_{22}^{-1} P_{21} = (1-\rho) I_{d_1} + \left(\rho - \frac{d_2 \rho^2}{d_2 \rho -\rho+\lambda  +1}\right)\bm{1}_{d_1} \bm{1}^\top_{d_1}.
\end{align*}
Therefore, the condition number is
\begin{align*}
    C_{\rm ROBOT} = 1+ \frac{1-\rho+\lambda}{1-\rho}\frac{d_1 \rho}{d_2 \rho -\rho+\lambda +1}.
\end{align*}
\end{proof}
Note that $C_{\rm AM}$ increases linearly with respect to $d_1$. Therefore, the optimization problem inevitably becomes ill-conditioned as dimension increase. In contrast, $C_{\rm ROBOT}$ can stay in the same order of magnitude when $d_1$ and $d_2$ increase simultaneously. 

Since the iteration complexity of gradient descent is proportional to the condition number \citep{bottou2018optimization}, ROBOT needs fewer iterations to converge than AM.

\section{Differentiability}
\label{sec:appendix_proof}

% We first provide more discussion on that $S^*(w)=\argmin_{S^*\in \Pi(\bm{1}_n/n, \bm{1}_n/n)} \langle C(w), S \rangle$ is not differentiable with respect to $w$. {\color{red}When $w$ changes, }

\textbf{Theorem 2.}
For any $\epsilon>0$, $S^*_\epsilon(w)$ is differentiable, as long as the cost $C(w)$ is differentiable with respect to $w$. As a result, the objective $\cL_{\epsilon}(w)=\langle C(w), S^*_\epsilon(w) \rangle$ is also differentiable. 

\begin{proof} 
The proof is analogous to \citet{xie2020differentiable}.

We first prove the differentiability of $S^*_\epsilon(w)$. This part of proof mirrors the proof in \citet{luise2018differential}.
By Sinkhorn's scaling theorem \citep{sinkhorn1967concerning}, 
\begin{align*}
        S^*_\epsilon(w) = {\rm diag}(e^{\frac{\xi^*(w)}{\epsilon}})e^{-\frac{C(w)}{\epsilon}}{\rm diag}(e^{\frac{\zeta^*(w)}{\epsilon}}).
    \end{align*}
Therefore, since $C_{ij}(w)$ is differentiable, $\Gamma^{*, \epsilon}$ is differentiable if
$(\xi^*(w), \zeta^*(w))$ is differentiable as a function of $w$.

Let us set 
\begin{align*}
        \mathcal{L}(\xi, \zeta; \mu, \nu, C) = \xi^T \mu + \zeta^T \nu -\epsilon \sum_{i,j=1}^{n,m} e^{-\frac{C_{ij}-\xi_i-\zeta_j}{\epsilon}}.
\end{align*} 
and recall that $(\xi^*, \zeta^*) = \argmax_{\xi, \zeta} L(\xi, \zeta; \mu, \nu, C)$. The differentiability of $(\xi^*, \zeta^*)$ is proved using the Implicit Function theorem and follows from the differentiability and strict convexity in $(\xi^*, \zeta^*)$ of the function $\mathcal{L}$. 
\end{proof}

\noindent \textbf{Theorem 3.} 
The gradient of $\cF_{\epsilon}$ with respect to $w$ is
\begin{align}
    \nabla_{w}{\cF_{\epsilon}}  = \frac{1}{\epsilon} \sum_{i,j=1}^{n,n} \left( (1- C_{ij})S^*_{\epsilon,ij} + \sum_{h,\ell=1}^{n,n} C_{h\ell}S^*_{\epsilon,h\ell}\frac{d\xi^*_h}{dC_{ij}} + \sum_{h,\ell=1}^{n,n} C_{h\ell}S^*_{\epsilon,h\ell}\frac{d\zeta^*_\ell}{dC_{ij}} \right)\nabla_{w}{ C_{ij}}, \label{eq:update_w_app}
\end{align}
where $\displaystyle \begin{bmatrix}
    \nabla_C \xi^*\\
    \nabla_C\zeta^*
\end{bmatrix} = 
\begin{bmatrix}
 - H^{-1} D \\
 \bm{0}
\end{bmatrix} \quad {\rm with} \quad - H^{-1} D\in \mathbb{R}^{(2n-1)\times n \times n}, \bm{0}\in \mathbb{R}^{1\times n \times n},$
{
\begin{align*}
    & D_{\ell ij} = \dfrac{1}{n\epsilon} \begin{cases}
    \delta_{\ell i} S^*_{\epsilon,ij},\quad  \ell=1, \cdots, n; \\
    \delta_{\ell j} S^*_{\epsilon,ij},\quad  \ell=n+1, \cdots, 2n-1,
    \end{cases} 
    H^{-1} = -{\epsilon}{n} \begin{bmatrix} 
    I_n +  \bar{S^*_{\epsilon}} \mathcal{K}^{-1} \bar{S^*_{\epsilon}}^T  & -  \bar{S^*_{\epsilon}} \mathcal{K}^{-1} \\
    -\mathcal{K}^{-1} \bar{S^*_{\epsilon}} ^T  & \mathcal{K}^{-1}
    \end{bmatrix}, \\
    &\textrm{and}\quad \mathcal{K} = I_{n-1}- \bar{S^*_{\epsilon}}^T  \bar{S^*_{\epsilon}}, \quad \bar{S^*_{\epsilon}} = S^*_{\epsilon, 1:n,1:n-1}.
\end{align*}
}

\begin{proof}
This result is straightforward combining the Sinkhorn's scaling theorem and Theorem 3 in \citet{xie2020differentiable}.
Specifically, notice the similarity between the lower-level optimization and (\ref{eq:kanto}),
    \begin{align*}
        \Gamma = \argmin_{\Gamma\in \Pi(\mu, \nu)} \langle C(w), \Gamma \rangle + \epsilon \sum_{i,j} \Gamma_{ij} \ln \Gamma_{ij}.
    \end{align*}
We will first derive $\nabla_C \Gamma$, then derive $\nabla_w\cF_\epsilon$ using the chain rule. To avoid possible confusion, we will derive the case where $\mu\in\RR^n$ and $\nu\in\RR^m$, and then take $\mu=\nu=\bm{1}_n/n$. The dual problem of the above optimization problem is
    \begin{align*}
        \xi^*, \zeta^* = \argmax_{\xi, \zeta} \mathcal{L}(\xi, \zeta; C),
    \end{align*}
    where
    \begin{align*}
        \mathcal{L}(\xi, \zeta; C) = \xi^\top \mu + \zeta^\top \nu -\epsilon \sum_{i,j=1}^{n,m} e^{-\frac{C_{ij}-\xi_i-\zeta_j}{\epsilon}}.
    \end{align*}
    And it is connected to the prime form by 
    \begin{align*}
        \Gamma^{*, \epsilon} = {\rm diag}(e^{\frac{\xi^*}{\epsilon}})e^{-\frac{C}{\epsilon}}{\rm diag}(e^{\frac{\zeta^*}{\epsilon}}).
    \end{align*}

Notice that there is one redundant dual variable, since $\mu \bm{1}_n = \nu \bm{1}_m=1$. Therefore, we can rewrite $\mathcal{L}(\xi, \zeta; C)$ as
\begin{align*}
        \mathcal{L}(\xi,\bar{\zeta}; C) = \xi^T \mu + \bar{\zeta}^T \bar{\nu} -\epsilon \sum_{i,j=1}^{n,m-1} e^{\frac{-C_{ij}+\xi_i+\zeta_j}{\epsilon}}-\epsilon \sum_{i=1}^{n} e^{\frac{-C_{im}+\xi_i}{\epsilon}}.
\end{align*}
Denote
\begin{align}
    &\phi(\xi, \bar{\zeta},C) = \frac{d \mathcal{L}(\xi, \bar{\zeta};C)}{d \xi}
     = \mu - F \bm{1}_m, \label{eq:phi} \\
   & \psi(\xi, \bar{\zeta},C) = \frac{d \mathcal{L}(\xi, \bar{\zeta};C)}{d \bar{\zeta}}
     = \bar{\nu} - \bar{F}^\top \bm{1}_n, \label{eq:psi} 
\end{align}
where 
\begin{align}
     & F_{ij}  = e^{\frac{-C_{ij}+\xi_i+\zeta_j}{\epsilon}}, \quad\forall i=1, \cdots, n, \quad j=1, \cdots, m-1 \nonumber \\
     & F_{im} = e^{\frac{-C_{im}+\xi_i}{\epsilon}}, \quad\forall i=1, \cdots, n, \nonumber\\
     & \bar{F}  = F_{:,:-1}. \nonumber
\end{align}
Since $(\xi^*, \bar{\zeta}^*)$ is a maximum of $\mathcal{L}(\xi,\bar{\zeta}; C)$, we have
\begin{align*}
    & \phi(\xi^*, \bar{\zeta}^*,C)=0, \\
    & \psi(\xi^*, \bar{\zeta}^*,C)=0 .
\end{align*}
Therefore,
\begin{align*}
    & \frac{d \phi(\xi^*, \bar{\zeta}^*,C)}{d C} = \frac{\partial \phi(\xi^*, \bar{\zeta}^*,C)}{\partial C} + \frac{\partial \phi(\xi^*, \bar{\zeta}^*,C)}{\partial \xi^*} \frac{d \xi^*}{d C}+ \frac{\partial \phi(\xi^*, \bar{\zeta}^*,C)}{\partial \bar{\zeta}^*} \frac{d \bar{\zeta}^*}{d C} = 0, \\
    & \frac{d \psi(\xi^*, \bar{\zeta}^*,C)}{d C} = \frac{\partial \psi(\xi^*, \bar{\zeta}^*,C)}{\partial C} + \frac{\partial \psi(\xi^*, \bar{\zeta}^*,C)}{\partial \xi^*} \frac{d \xi^*}{d C}+ \frac{\partial \psi(\xi^*, \bar{\zeta}^*,C)}{\partial \bar{\zeta}^*} \frac{d \bar{\zeta}^*}{d C} = 0.
\end{align*}
Therefore,
\begin{align*}
    \begin{bmatrix}
    \frac{d \xi^*}{d C} \\
    \frac{d \bar{\zeta}^*}{d C}
    \end{bmatrix} & = -
    \begin{bmatrix}
    \frac{\partial \phi(\xi^*, \bar{\zeta}^*,C)}{\partial \xi^*} & \frac{\partial \phi(\xi^*, \bar{\zeta}^*,C)}{\partial \bar{\zeta}^*}  \\
    \frac{\partial \psi(\xi^*, \bar{\zeta}^*,C)}{\partial \xi^*} & \frac{\partial \psi(\xi^*, \bar{\zeta}^*,C)}{\partial \bar{\zeta}^*}
    \end{bmatrix}^{-1}  
    \begin{bmatrix}
    \frac{\partial \phi(\xi^*, \bar{\zeta}^*,C)}{\partial C} \\
    \frac{\partial \psi(\xi^*, \bar{\zeta}^*,C)}{\partial C}
    \end{bmatrix} \\
    & \triangleq -
    H^{-1}  
    \begin{bmatrix}
    D^{(1)} \\
    D^{(2)}
    \end{bmatrix} \\
    & \triangleq - H^{-1} D.
\end{align*}
After some derivations, we have
\begin{align*}
    H = -\frac{1}{\epsilon}\begin{bmatrix}
        {\rm diag}(\mu) & \bar{\Gamma} \\
        \bar{\Gamma}^T & {\rm diag}(\bar{\nu})
    \end{bmatrix}.
\end{align*}
Following the formula for inverse of block matrices,
\begin{align*}
    \begin{bmatrix}
    \mathbf{A} & \mathbf{B} \\
    \mathbf{C} & \mathbf{D}
  \end{bmatrix}^{-1} = \begin{bmatrix}
     \mathbf{A}^{-1} + \mathbf{A}^{-1}\mathbf{B}(\mathbf{D} - \mathbf{CA}^{-1}\mathbf{B})^{-1}\mathbf{CA}^{-1} &
      -\mathbf{A}^{-1}\mathbf{B}(\mathbf{D} - \mathbf{CA}^{-1}\mathbf{B})^{-1} \\
    -(\mathbf{D}-\mathbf{CA}^{-1}\mathbf{B})^{-1}\mathbf{CA}^{-1} &
       (\mathbf{D} - \mathbf{CA}^{-1}\mathbf{B})^{-1}
  \end{bmatrix},
\end{align*}
denote 
\begin{align*}
    \mathcal{K} = {\rm diag}(\bar{\nu}) - \bar{\Gamma}^T ({\rm diag}(\mu))^{-1} \bar{\Gamma}.
\end{align*}
Finally we have
\begin{align*}
    H^{-1} = -\epsilon \begin{bmatrix} 
    ({\rm diag}(\mu))^{-1} + ({\rm diag}(\mu))^{-1} \bar{\Gamma} \mathcal{K}^{-1} \bar{\Gamma}^T ({\rm diag}(\mu))^{-1} & - ({\rm diag}(\mu))^{-1} \bar{\Gamma} \mathcal{K}^{-1} \\
    -\mathcal{K}^{-1} \bar{\Gamma} ^T ({\rm diag}(\mu))^{-1} & \mathcal{K}^{-1}
    \end{bmatrix}.
\end{align*}
And also
\begin{align*}
    & D^{(1)}_{hij} = \frac{1}{\epsilon} \delta_{hi}\Gamma_{ij} \\
    & D^{(2)}_{\ell ij} = \frac{1}{\epsilon} \delta_{\ell j}\Gamma_{ij}.
\end{align*}
The above derivation can actually be viewed as we explicitly force $\zeta_m=0$, i.e., no matter how $C$ changes, $\zeta_m$ does not change. Therefore, we can treat $\frac{d\zeta_m}{dC}=\bm{0}_{n\times m}$. 

After we obtain $\frac{d\xi^*}{dC}$ and $\frac{d\zeta^*}{dC}$, we can now compute $\frac{d\Gamma}{dC}$.
\begin{align*}
    \frac{d\Gamma_{h\ell}}{d C_{ij}} = \frac{d}{dC_{ij}} e^{\frac{-C_{h\ell}+\xi^*_h+\zeta^*_\ell}{\epsilon}} = \frac{1}{\epsilon} \left( -\Gamma_{h\ell}\delta_{ih}\delta_{j\ell} + \Gamma_{h\ell}\frac{d\xi^*_h}{dC_{ij}} + \Gamma_{h\ell}\frac{d\zeta^*_\ell}{dC_{ij}} \right).
\end{align*}
Note that $\cF_\epsilon(w) = \langle C(w), n\Gamma\rangle$.
Substituting $\frac{d\Gamma_{h\ell}}{d C_{ij}}$ into the expression of $\nabla_w \cF_\epsilon(w)$, we get the equation in the theorem.

\end{proof}

\section{Algorithm of the Forward Pass for ROBOT-robust}
\label{sec:appendix_robust_foward}

For better numerical stability, in practice we add two more regularization terms,
\begin{align}
     S^*_{\rm r}(w),\bar{{\mu}}^*, \bar{{\nu}}^*  & = \sideset{}{_{S\in \Pi(\bar{{\mu}}, \bar{{\nu}}),~\bar{{\mu}}, \bar{{\nu}}\in \Delta_n}}\argmin
    \langle C(w), S \rangle + \epsilon H(S) +\epsilon_1 h(\bar{\mu})+\epsilon_2 h(\bar{\nu}), \label{eq:robust_all}\\
    & {\rm s.t.~} \cF(\bar{{\mu}}, {\mu}) \leq \rho_1, ~\cF(\bar{{\nu}}, {\nu}) \leq \rho_2, \nonumber
\end{align}
where $h(\bar{\mu})=\sum_i \bar{\mu}_i \log \bar{\mu}_i$ is the entropy function for vectors. This can avoid the entries of $\bar{\mu}$ and $\bar{\nu}$ shrink to zeros when updated by gradient descent. We remark that since we have entropy term $H(S)$, the entries of $S$ would not be exactly zeros. Furthermore, we have $\bar{\mu}=S\bm{1}$ and $\bar{\mu}=S\bm{1}$. Therefore, theoretically the entries of $\bar{\mu}$ and $\bar{\nu}$ will not be zeros. We only add the two more entropy terms for numerical consideration.
The detailed algorithm is in Algorithm \ref{alg:robust_forward}. Although the algorithm is not guaranteed to converge to a feasible solution, in practice it usually converges to a good solution \citep{wang2015random}. 

\begin{algorithm}
    \caption{Solving $S^*_{\rm r}$ for robust matching}
    \label{alg:robust_forward}
    \begin{algorithmic}
        \REQUIRE $C \in \mathbb{R}^{m\times n},{\mu}, {\nu}, K, \epsilon, L, \eta$
        \STATE  $ G_{ij} = e^{-\frac{C_{ij}}{\epsilon}}$
        \STATE $\bar{\mu}=\mu, \bar{\nu}=\nu$
        \STATE $b = \bm{1}_n$
        \FOR{$l = 1, \cdots, L$}
        \STATE $a = \bar{\mu}/(Gb), b = \bar{\nu}/(G^T a) $
        \STATE $\bar{\mu} = \bar{\mu} - \eta (e^{\frac{a}{\epsilon}}+\epsilon_1*\log\bar{\mu} ),\bar{\nu} = \bar{\nu} - \eta (e^{\frac{b}{\epsilon}}+\epsilon_2*\log\bar{\nu} )$
        \STATE $\bar{\mu} = \max \{\bar{\mu}, 0\},\bar{\nu} = \max \{ \bar{\nu}, 0\}$
        \STATE $\bar{\mu} = \bar{\mu}/(\bar{\mu}^\top \bm{1}),\bar{\nu} =\bar{\nu}/(\bar{\nu}^\top \bm{1})$
        \IF{$\norm{\bar{\mu}-\mu}^2_2 > \rho_1$}
        \STATE $\bar{\mu} = \mu + \sqrt{\rho_1}\frac{\bar{\mu}-\mu}{\norm{\bar{\mu}-\mu}_2}$
        % \STATE $\bar{\mu} = \bar{\mu}/(\bar{\mu}^\top \bm{1})$
        \ENDIF
        \IF{$\norm{\bar{\nu}-\nu}^2_2 > \rho_2$}
        \STATE $\bar{\nu} = \nu + \sqrt{\rho_2}\frac{\bar{\nu}-\nu}{\norm{\bar{\nu}-\nu}_2}$
        % \STATE $\bar{\nu} = \bar{\nu}/(\bar{\nu}^\top \bm{1})$
        \ENDIF
        \ENDFOR
        \STATE $S = {\rm diag}(a)\odot G \odot {\rm diag}(b)$
    \end{algorithmic}
\end{algorithm}

\section{Algorithm of the Backward Pass for ROBOT-robust}
\label{sec:appendix_robust}

We first summarize the outline of the derivation, then provide the detailed derivation.

\subsection{Summary}

Given $\bar{{\mu}}^*, \bar{{\nu}}^*, S^*_{\rm r}(w)$, we compute the Jacobian matrix $d S^*_{\rm r}(w)/d w$ using implicit differentiation and differentiable programming techinques. Specifically, the Lagrangian function of Problem (\ref{eq:robust_all}) is
\begin{align*}
    \mathcal{L} = & \langle C, S \rangle+ \epsilon H(S)+ \epsilon_1 h(\bar{{\mu}}) + \epsilon_2 h(\bar{{\nu}}) - \xi^\top (\Gamma \bm{1}_m - \mu) - \zeta^\top (\Gamma^\top \bm{1}_n - \nu) \\
    &+ \lambda_1( \bar{\mu}^\top \bm{1}_n-1)+ \lambda_2( \bar{\nu}^\top \bm{1}_m - 1) + \lambda_3( \norm{\bar{\mu} - \mu}_2^2 - \rho_1)+ \lambda_4 ( \norm{\bar{\nu} - \nu}_2^2 - \rho_2).
\end{align*}
where $\xi$ and $\zeta$ are dual variables.
The KKT conditions (Stationarity condition) imply that the optimal solution $\Gamma^{*, \epsilon}$ can be formulated using the optimal dual variables $\xi^*$ and $\zeta^*$ as, 
\begin{align} \label{eq:pri_dual}
    S^{*}_{\rm r} = {\rm diag}(e^{\frac{\xi^*}{\epsilon}})e^{-\frac{C}{\epsilon}}{\rm diag}(e^{\frac{\zeta^*}{\epsilon}}).
\end{align}
By the chain rule, we have 
\begin{align*}
    \frac{d S^*_{\rm r}}{dw} = \frac{d S^*_{\rm r}}{dC} \frac{dC}{dw} = \left( \frac{\partial S^*_{\rm r}}{\partial C}+\frac{\partial S^*_{\rm r}}{\partial \xi^*}\frac{d \xi^*}{dC}+\frac{\partial S^*_{\rm r}}{\partial \zeta^*}\frac{d \zeta^*}{dC} \right)\frac{dC}{dw}.
\end{align*}
Therefore, we can compute  $d S^*_{\rm r}(w)/d w$ if we obtain  $\frac{d \xi^*}{dC}$ and $\frac{d \zeta^*}{dC}$.

Substituting (\ref{eq:pri_dual}) into the Lagrangian function, at the optimal solutions we obtain 
\begin{align*}
    \mathcal{L} = \mathcal{L}(\xi^*,\zeta^*, \bar{\mu}^*, \bar{\nu}^*, \lambda_1^*, \lambda_2^*, \lambda_3^*, \lambda_4^*; C).
\end{align*}

Denote $r^* = [(\xi^*)^\top, (\zeta^*)^\top, (\bar{\mu})^\top, (\bar{\nu})^\top, \lambda_1^*, \lambda_2^*, \lambda_3^*, \lambda_4^*]^\top$, and $\phi(r^*; C) = {\partial \mathcal{L}(r^*; C)}/{\partial r^*}$.
At the optimal dual variable $r^*$, the KKT condition immediately yields
$\phi(r^*; C) \equiv 0$.
By the chain rule, we have
\begin{align} \label{eq:chain}
    \frac{d \phi(r^*; C)}{d C} = \frac{\partial \phi(r^*; C)}{\partial C} + \frac{\partial \phi(r^*; C)}{\partial r^*} \frac{dr^*}{d C} = 0.
\end{align}
Rerranging terms, we obtain 
\begin{align} \label{eq:omega_C}
    \frac{dr^*}{d C} = - \left(\frac{\partial \phi(r^*; C)}{\partial r^*}\right)^{-1} \frac{\partial \phi(r^*; C)}{\partial C}.
\end{align}
Combining (\ref{eq:pri_dual}), (\ref{eq:chain}), and (\ref{eq:omega_C}), we can then obtain $d S^*_{\rm r}(w)/d w$.

\subsection{Details}

Now we provide the detailed derivation for computing $dS^*_{\rm r}/dw$. 

Since $S^*_{\rm r}$ is the optimal solution of an optimization problem, we can follow the implicit function theorem to solve for the closed-form expression of the gradient. Specifically, we adopt $\cF(\bar{\mu}, \nu) = \sum_i (\bar{\mu}_i - \mu_i)^2$, and rewrite the optimization problem as
\begin{align*}
      & \min_{\bar{\mu}, \bar{\nu}, S}
    \langle C, S \rangle + \epsilon \sum_{ij} S_{ij}(\log S_{ij}-1)+ \epsilon_1 \sum_{i} \bar{\mu}_{i}(\log \bar{\mu}_{i}-1)+ \epsilon_2 \sum_{j} \bar{\nu}_{j}(\log \bar{\nu}_{j}-1), \\
     {\rm s.t.,~}  & \sum_j S_{ij} = \bar{\mu}_i, \quad \sum_i S_{ij} = \bar{\nu}_j, \\
    & \sum_i \bar{\mu}_i = 1, \quad \sum_j \bar{\nu}_j = 1, \\
    & \sum_i (\bar{\mu}_i - \mu_i)^2 \leq \rho_1, \quad \sum_j (\bar{\nu}_j - \nu_j)^2 \leq \rho_2.
\end{align*}
The Language of the above problem is
\begin{align*}
     & \cL(C, S,\bar{\mu}, \bar{\nu}, \xi, \zeta, \lambda_1, \lambda_2, \lambda_3, \lambda_4) \\
    & = \langle C, S \rangle + \epsilon \sum_{ij} S_{ij}(\log S_{ij}-1)+ \epsilon_1 \sum_{i} \bar{\mu}_{i}(\log \bar{\mu}_{i}-1)+ \epsilon_2 \sum_{j} \bar{\nu}_{j}(\log \bar{\nu}_{j}-1) \\
    & - \xi^\top (S\bm{1}_m-\bar{\mu}) - \zeta^\top (S^\top\bm{1}_n-\bar{\nu}) \\
    &+ \lambda_1(\sum_i \bar{\mu}_i-1)+ \lambda_2( \sum_j \bar{\nu}_j - 1) + \lambda_3( \sum_i (\bar{\mu}_i - \mu_i)^2 - \rho_1)+ \lambda_4 ( \sum_j (\bar{\nu}_j - \nu_j)^2 - \rho_2).
\end{align*}
Easy to see that the Slater's condition holds. Denote 
\begin{align*}
    \cL^* = \cL(C, S^*_{\rm r},\bar{\mu}^*, \bar{\nu}^*, \xi^*, \zeta^*, \lambda^*_1, \lambda^*_2, \lambda^*_3, \lambda^*_4).
\end{align*}
Following the KKT conditions,
\begin{align*}
    \frac{d \cL^*}{d S^*_{{\rm r}, ij}} = C_{ij} + \epsilon \log S^*_{{\rm r}, ij} -\xi_i^* -\zeta_j^* = 0.
\end{align*}
Therefore, $S^*_{{\rm r},ij} =  e^{\frac{\xi^*_i+\zeta^*_j-C_{ij}}{\epsilon}}$. Then we have
\begin{align*}
    \frac{d S^*_{\rm r}}{d w} =  (\frac{\partial S^*_{\rm r}}{\partial C} + \frac{\partial S^*_{\rm r}}{\partial \xi^*} \frac{d \xi^*}{dC}+ \frac{\partial S^*_{\rm r}}{\partial \zeta^*}\frac{d \zeta^*}{dC})\frac{dC}{dw}.
\end{align*}
So all we need to do is to compute $\frac{d \xi^*}{dC}$ and $\frac{d \zeta^*}{dC}$. Denote $F_{ij} = e^{\frac{\xi_i+\zeta_j-C_{ij}}{\epsilon}}$. Denote
\begin{align*}
    &\phi = \frac{d \mathcal{L}}{d \xi}
     = \bar{\mu} - F \bm{1}_m, \\
   & \psi = \frac{d \mathcal{L}}{d {\zeta}}
     = \bar{\nu} - {F}^\top \bm{1}_n, \\
     &p = \frac{d \mathcal{L}}{d \bar{\mu}}
     = \xi + \lambda_1 \bm{1}_n +2\lambda_3 (\bar{\mu}-\mu) +\epsilon_1 \log \bar{\mu}, \\
     &q = \frac{d \mathcal{L}}{d \bar{\nu}}
     = \zeta + \lambda_2 \bm{1}_m +2\lambda_4 (\bar{\nu}-\nu) +\epsilon_2 \log \bar{\nu},  \\
     & \chi_1 = \frac{d \mathcal{L}}{d \lambda_1} = \bar{\mu}^\top \bm{1}_n -1, \\
     & \chi_2 = \frac{d \mathcal{L}}{d \lambda_2} = \bar{\nu}^\top \bm{1}_m -1, \\
     & \chi_3 = \lambda_3(\norm{\bar{\mu}-\mu}_2^2-\rho_1), \\
     & \chi_4 = \lambda_4(\norm{\bar{\nu}-\nu}_2^2-\rho_2).
\end{align*}
Denote $\chi=[\chi_1, \chi_2, \chi_3, \chi_4]$, and $\lambda = [\lambda_1, \lambda_2,\lambda_3, \lambda_4]$. Following the KKT conditions, we have
\begin{align*}
     \phi=0, \psi=0, p=0, q=0, \chi=0,
\end{align*}
at the optimal solutions.  Therefore, for the optimal solutions we have
\begin{align*}
    & \frac{d \phi}{d C} = \frac{\partial \phi}{\partial C} + \frac{\partial \phi}{\partial \xi^*} \frac{d \xi^*}{d C}+ \frac{\partial \phi}{\partial {\zeta}^*} \frac{d {\zeta}^*}{d C} + \frac{\partial \phi}{\partial \bar{\mu}^*} \frac{d \bar{\mu}^*}{d C}+ \frac{\partial \phi}{\partial \bar{\nu}^*} \frac{d \bar{\nu}^*}{d C}+ \frac{\partial \phi}{\partial \lambda^*} \frac{d \lambda^*}{d C} = 0, \\
    & \frac{d \psi}{d C} = \frac{\partial \psi}{\partial C} + \frac{\partial \psi}{\partial \xi^*} \frac{d \xi^*}{d C}+ \frac{\partial \psi}{\partial {\zeta}^*} \frac{d {\zeta}^*}{d C} + \frac{\partial \psi}{\partial \bar{\mu}^*} \frac{d \bar{\mu}^*}{d C}+ \frac{\partial \psi}{\partial \bar{\nu}^*} \frac{d \bar{\nu}^*}{d C}+ \frac{\partial \psi}{\partial \lambda^*} \frac{d \lambda^*}{d C}= 0, \\
    & \frac{d p}{d C} = \frac{\partial p}{\partial C} + \frac{\partial p}{\partial \xi^*} \frac{d \xi^*}{d C}+ \frac{\partial p}{\partial {\zeta}^*} \frac{d {\zeta}^*}{d C} + \frac{\partial p}{\partial \bar{\mu}^*} \frac{d \bar{\mu}^*}{d C}+ \frac{\partial p}{\partial \bar{\nu}^*} \frac{d \bar{\nu}^*}{d C}+ \frac{\partial p}{\partial \lambda^*} \frac{d \lambda^*}{d C}= 0, \\
    & \frac{d q}{d C} = \frac{\partial q}{\partial C} + \frac{\partial q}{\partial \xi^*} \frac{d \xi^*}{d C}+ \frac{\partial q}{\partial {\zeta}^*} \frac{d {\zeta}^*}{d C} + \frac{\partial q}{\partial \bar{\mu}^*} \frac{d \bar{\mu}^*}{d C}+ \frac{\partial q}{\partial \bar{\nu}^*} \frac{d \bar{\nu}^*}{d C}+ \frac{\partial q}{\partial \lambda^*} \frac{d \lambda^*}{d C}= 0 \\
    & \frac{d \chi}{d C} = \frac{\partial \chi}{\partial C} + \frac{\partial \chi}{\partial \xi^*} \frac{d \xi^*}{d C}+ \frac{\partial \chi}{\partial {\zeta}^*} \frac{d {\zeta}^*}{d C} + \frac{\partial \chi}{\partial \bar{\mu}^*} \frac{d \bar{\mu}^*}{d C}+ \frac{\partial \chi}{\partial \bar{\nu}^*} \frac{d \bar{\nu}^*}{d C}+ \frac{\partial \chi}{\partial \lambda^*} \frac{d \lambda^*}{d C}= 0.
\end{align*}
Therefore, we have
\begin{align*}
    \begin{bmatrix}
    \dfrac{d \xi^*}{d C} \\
    \dfrac{d {\zeta}^*}{d C} \\
    \dfrac{d \bar{\mu}^*}{d C} \\
    \dfrac{d \bar{\nu}^*}{d C} \\
    \dfrac{d \lambda^*}{d C}
    \end{bmatrix} & = -
    \begin{bmatrix}
    \dfrac{\partial \phi}{\partial \xi^*} & \dfrac{\partial \phi}{\partial {\zeta}^*} & \dfrac{\partial \phi}{\partial \bar{\mu}^*} & \dfrac{\partial \phi}{\partial \bar{\nu}^*} &
    \dfrac{\partial \phi}{\partial \lambda^*}\\
    \dfrac{\partial \psi}{\partial \xi^*} & \dfrac{\partial \psi}{\partial {\zeta}^*} & \dfrac{\partial \psi}{\partial \bar{\mu}^*} & \dfrac{\partial \psi}{\partial \bar{\nu}^*} &
    \dfrac{\partial \psi}{\partial \lambda^*}\\
    \dfrac{\partial p}{\partial \xi^*} & \dfrac{\partial p}{\partial {\zeta}^*} & \dfrac{\partial p}{\partial \bar{\mu}^*} & \dfrac{\partial p}{\partial \bar{\nu}^*} &
    \dfrac{\partial p}{\partial \lambda^*}\\
    \dfrac{\partial q}{\partial \xi^*} & \dfrac{\partial q}{\partial {\zeta}^*} & \dfrac{\partial q}{\partial \bar{\mu}^*} & \dfrac{\partial q}{\partial \bar{\nu}^*} & \dfrac{\partial q}{\partial \lambda^*} \\
    \dfrac{\partial \chi}{\partial \xi^*} & \dfrac{\partial \chi}{\partial {\zeta}^*} & \dfrac{\partial \chi}{\partial \bar{\mu}^*} & \dfrac{\partial \chi}{\partial \bar{\nu}^*} & \dfrac{\partial \chi}{\partial \lambda^*}
    \end{bmatrix}^{-1}  
    \begin{bmatrix}
    \dfrac{\partial \phi}{\partial C} \\
    \dfrac{\partial \psi}{\partial C} \\
    \dfrac{\partial p}{\partial C} \\
    \dfrac{\partial q}{\partial C} \\
    \dfrac{\partial \chi}{\partial C}
    \end{bmatrix} .
\end{align*}
After some derivation, we have
{\tiny
\begin{align*}
    \begin{bmatrix}
    \dfrac{d \xi^*}{d C} \\
    \dfrac{d {\zeta}^*}{d C} \\
    \dfrac{d \bar{\mu}^*}{d C} \\
    \dfrac{d \bar{\nu}^*}{d C} \\
    \dfrac{d \lambda_1^*}{d C} \\
    \dfrac{d \lambda_2^*}{d C} \\
    \dfrac{d \lambda_3^*}{d C} \\
    \dfrac{d \lambda_4^*}{d C}
    \end{bmatrix} & = -
    \begin{bmatrix}
    -\dfrac{1}{\epsilon} {\rm diag}(\bar{\mu}) & -\dfrac{1}{\epsilon} S^*_{\rm r} & \bm{I}_n & \bm{0}  & \bm{0} & \bm{0} & \bm{0} & \bm{0}\\
    -\dfrac{1}{\epsilon} (S^*_{\rm r})^\top & -\dfrac{1}{\epsilon} {\rm diag}(\bar{\nu}) & \bm{0} & \bm{I}_m  & \bm{0} & \bm{0} & \bm{0} & \bm{0}\\
    \bm{I}_n & \bm{0} & 2\lambda_3 \bm{I}_n + {\rm diag}(\frac{\epsilon_1}{\bar{\mu}}) & \bm{0}  & \bm{1}_n & \bm{0} & 2(\bar{\mu}-\mu) & \bm{0} \\
    \bm{0} & \bm{I}_m & \bm{0} & 2\lambda_4\bm{I}_m+ {\rm diag}(\frac{\epsilon_2}{\bar{\nu}})  & \bm{0} & \bm{1}_m & \bm{0} & 2(\bar{\nu}-\nu) \\
    \bm{0} & \bm{0} & \bm{1}_n^\top & \bm{0} & 0 & 0 & 0 & 0 \\
    \bm{0} & \bm{0} & \bm{0} & \bm{1}_m^\top & 0 & 0 & 0 & 0 \\
    \bm{0} & \bm{0} & 2\lambda_3(\bar{\mu}-\mu)^\top & \bm{0} & 0 &0 & \norm{\bar{\mu}-\mu}_2^2-\rho_1 & 0 \\
    \bm{0} & \bm{0} & \bm{0} & 2\lambda_4(\bar{\nu}-\nu)^\top & 0 & 0 & 0 & \norm{\bar{\nu}-\nu}_2^2-\rho_2 
    \end{bmatrix}^{-1}  
    \begin{bmatrix}
    \dfrac{\partial \phi}{\partial C} \\
    \dfrac{\partial \psi}{\partial C} \\
    \bm{0} \\
    \bm{0} \\
    0 \\
    0 \\
    0 \\
    0
    \end{bmatrix},
\end{align*}
}
and
\begin{align*}
    & \frac{\partial \phi_h}{\partial C_{ij}} =  \frac{1}{\epsilon} \delta_{hi} S_{ij}, \forall h=1, \cdots, n, \quad i=1, \cdots, n, \quad j=1, \cdots, m \\
    & \frac{\partial \psi_\ell}{\partial C_{ij}}  =  \frac{1}{\epsilon} \delta_{\ell j} S_{ij}, \forall \ell=1, \cdots, m-1, \quad i=1, \cdots, n, \quad j=1, \cdots, m.
\end{align*}
To efficiently solve for the inverse in the above equations, we denote
\begin{align*}
    A & =
    \begin{bmatrix}
    -\dfrac{1}{\epsilon} {\rm diag}(\bar{\mu}) & -\dfrac{1}{\epsilon} S^*_{\rm r} & \bm{I}_n & \bm{0}\\
    -\dfrac{1}{\epsilon} (S^*_{\rm r})^\top & -\dfrac{1}{\epsilon} {\rm diag}(\bar{\nu}) & \bm{0} & \bm{I}_m  \\
    \bm{I}_n & \bm{0} & 2\lambda_3 \bm{I}_n + {\rm diag}(\frac{\epsilon_1}{\bar{\mu}}) & \bm{0} \\
    \bm{0} & \bm{I}_m & \bm{0} & 2\lambda_4\bm{I}_m+ {\rm diag}(\frac{\epsilon_2}{\bar{\nu}}) 
    \end{bmatrix},
\end{align*}
\begin{align*}
    B_1 & =
    \begin{bmatrix}
     \bm{1}_n & \bm{0} & 2(\bar{\mu}-\mu) & \bm{0} \\
    \bm{0} & \bm{1}_m & \bm{0} & 2(\bar{\nu}-\nu) 
    \end{bmatrix},
\end{align*}

\begin{align*}
    C_1 & =
    \begin{bmatrix}
    \bm{1}_n^\top & \bm{0} \\
    \bm{0} & \bm{1}_m^\top \\
     2\lambda_3(\bar{\mu}-\mu)^\top & \bm{0} \\
    \bm{0} & 2\lambda_4(\bar{\nu}-\nu)^\top
    \end{bmatrix},
\end{align*}

\begin{align*}
    D & =
    \begin{bmatrix}
    0 & 0 & 0 & 0 \\
    0 & 0 & 0 & 0 \\
    0 &0 & \norm{\bar{\mu}-\mu}_2^2-\rho_1 & 0 \\
    0 & 0 & 0 & \norm{\bar{\nu}-\nu}_2^2-\rho_2 
    \end{bmatrix}.
\end{align*}
We first $A^{-1}$ using the rules for inverting a block matrix,
\begin{align*}
    A^{-1} = \begin{bmatrix}
    K & -KL \\
    -LK & L+LKL
    \end{bmatrix} =:
    \begin{bmatrix}
    A_1 & A_2 \\
    A_3 & A_4
    \end{bmatrix}
\end{align*}
where
\begin{align*}
    L = \begin{bmatrix}
     2\lambda_3 \bm{I}_n+ {\rm diag}(\frac{\epsilon_1}{\bar{\mu}}) & \bm{0}  \\
    \bm{0} & 2\lambda_4\bm{I}_m + {\rm diag}(\frac{\epsilon_1}{\bar{\nu}})
    \end{bmatrix}^{-1}, \quad
    K &  =  \left(\dfrac{1}{\epsilon}
    \begin{bmatrix}
     {\rm diag}(\bar{\mu}) &  S^*_{\rm r}   \\
     (S^*_{\rm r})^\top &  {\rm diag}(\bar{\nu}) \end{bmatrix} +   
    L\right)^{-1}.
\end{align*}
Then using the rules of inverting a block matrix again, we have
\begin{align*}
    \begin{bmatrix}
    \dfrac{d \xi^*}{d C} \\
    \dfrac{d {\zeta}^*}{d C}
    \end{bmatrix} = 
    (A_1 + A_2 B_1(D-C_1 A_4 B_1)^{-1}C_1 A_3)\begin{bmatrix}
    \dfrac{\partial \phi}{\partial C} \\
    \dfrac{\partial \psi}{\partial C} \end{bmatrix}.
\end{align*}
Therefore, the bottleneck of computation is the inverting step in computing $K$. Note $L$ is a diagonal matrix, we can further lower the computation cost by applying the rules for inverting a block matrix again. The value of $\lambda_3$ and $\lambda_4$ can be estimated from the fact $p=0, q=0$ . We detail the algorithm in Algorithm \ref{alg:w_grad}. 

%()

\begin{algorithm}
\caption{\label{alg:w_grad} Computing the gradient for $w$}
\begin{algorithmic} 
\REQUIRE $C \in \mathbb{R}^{m\times n}, {\mu}, {\nu}, \epsilon, \frac{d C}{d w}$
\STATE Run forward pass to get $S=S^*_{\rm r}, \bar{\mu}, \bar{\nu}, \xi, \zeta$
\STATE $x_1 = \sum_{i=1}^{\lceil n/2 \rceil} (\bar{\mu}_i-\mu_i), x_2 = \sum_{i=\lceil n/2 \rceil}^n (\bar{\mu}_i-\mu_i), b_1 =- \sum_{i=1}^{\lceil n/2 \rceil} \xi_i, b_2 = -\sum_{i=\lceil n/2 \rceil}^n \xi_i $ 
\STATE $[\lambda_1, \lambda_3]^\top = [\lceil n/2 \rceil, x_1; n-\lceil n/2 \rceil, x_2]^{-1}[b1, b2]^\top$
\STATE $x_1 = \sum_{j=1}^{\lceil m/2 \rceil} (\bar{\nu}_j-\nu_j), x_2 = \sum_{j=\lceil m/2 \rceil}^m (\bar{\nu}_j-\nu_j), b_1 =- \sum_{j=1}^{\lceil m/2 \rceil} \zeta_j, b_2 = -\sum_{j=\lceil m/2 \rceil}^m \zeta_j $ 
\STATE $[\lambda_2, \lambda_4]^\top = [\lceil m/2 \rceil, x_1; m-\lceil m/2 \rceil, x_2]^{-1}[b1, b2]^\top$
\STATE $\bar{\mu}=\bar{\mu}+\epsilon(2\lambda_3 \bm{1}_n+ \frac{\epsilon_1}{\bar{\mu}})^{-1}, \bar{\nu}=\bar{\nu}+\epsilon(2\lambda_4 \bm{1}_m+ \frac{\epsilon_2}{\bar{\nu}})^{-1}$
\STATE $\bar{\nu}' = \bar{\nu}[:-1], S' = S[:,:-1]$
\STATE $\mathcal{K} \leftarrow {\rm diag}(\bar{\nu}')  - (S')^T ({\rm diag}(\bar{\mu}))^{-1} S'$
\STATE $H_1 \leftarrow ({\rm diag}(\bar{\mu}))^{-1} + ({\rm diag}(\bar{\mu}))^{-1} S' \mathcal{K}^{-1} (S')^\top ({\rm diag}(\bar{\mu}))^{-1} $ 
\STATE $H_2 \leftarrow - ({\rm diag}(\bar{\mu}))^{-1} S' \mathcal{K}^{-1}$ 
\STATE $H_3 \leftarrow  (H_2)^\top $
\STATE $H_4 \leftarrow \mathcal{K}^{-1} $ 
\STATE Pad $H_2$ to be $[n, m]$ with value $0$
\STATE Pad $H_3$ to be $[m, n]$ with value $0$
\STATE Pad $H_4$ to be $[m, m]$ with value $0$
\STATE $L = {\rm diag}([\epsilon(2\lambda_3 \bm{1}_n+ \frac{\epsilon_1}{\bar{\mu}})^{-1}, \epsilon(2\lambda_4 \bm{1}_m+ \frac{\epsilon_2}{\bar{\nu}})^{-1}])$
\STATE $A1 = [H_1, H_2; H_3, H_4]$
\STATE $A_2 = -A_1 \cdot L$
\STATE $A_3 = A_2^\top$
\STATE $A_4 = L + L\cdot A_1\cdot L$
\STATE $E = A_1 + A_2\cdot B1(D-C\cdot A_4\cdot B)^{-1}C\cdot A_3$, where $B1, C_1, D$ defined above
\STATE $[J_1, J_2; J_3, J_4] = E$, where $J_1\in \RR^{n\times n}, J_2\in \RR^{n\times m}, J_3\in \RR^{m\times n}, J_4\in \RR^{m\times m}$
\STATE $[\frac{d \xi^*}{dC}]_{nij} \leftarrow  [J_1]_{ni}S_{ij}+[J_2]_{nj}S_{ij}$ 
\STATE $[\frac{d \zeta^*}{dC}]_{mij} \leftarrow  [J_3]_{mi}S_{ij}+[J_4]_{mj}S_{ij}$ 
\STATE Pad $\frac{d \zeta^*}{dC}$ to be $[m, n, m]$ with value $0$
\STATE $[\frac{d \mathcal{L}}{dC}]_{ij} \leftarrow \frac{1}{\epsilon}(-C_{ij}S_{ij} + \sum_{n,m}C_{nm}S_{nm}[\frac{d a^*}{dC}]_{nij} + \sum_{n,m}C_{nm}S_{nm}[\frac{d b^*}{dC}]_{mij} ) + S_{ij}$
\RETURN $\dfrac{d \mathcal{L}}{dC}\dfrac{dC}{dw}$
\end{algorithmic}
\end{algorithm}

\section{Details on Experiments}
\label{sec:appendix_exp}

\subsection{Unlabeled Sensing}

We now provide more training details for experiments in Section \ref{sec:sec51}. Here, AM and ROBOT is trained with batch size $500$ and learning rate $10^{-4}$ for $2,000$ iterations. For the Sinkhorn algorithm in ROBOT we set $\epsilon=10^{-4}$. We run RS for $2\times 10^{5}$ iterations with inlier threshold as $10^{-2}$. Other settings for the hyper-parameters in the baselines follows the default settings of their corresponding papers.

\subsection{Nonlinear Regression}
For the nonlinear regression experiment in Section \ref{sec:sec52}, ROBOT and ROBOT-robust is trained with learning rate $10^{-4}$ for $80$ iterations. For $n=100,200,500,1000,2000$, we set batch size $10,30,50,100,300$, respectively.We set  $\epsilon=10^{-4}$ for the Sinkhorn algorithm in ROBOT. For Oracle and LS, we perform ordinary regression model and ensure convergence, i.e., learning rate $5\times10^{-2}$ for $100$ iterations.
\subsection{Flow Cytometry}
We provide more details for the Flow Cytometry experiment in Section \ref{sec:sec53}. In the FC seting, ROBOT is trained with batch size $1260$ and learning rate $10^{-4}$  for $80$ iterations. In the GFC seting, ROBOT is trained with batch size $1260$ and learning rate $6\times10^{-4}$ for $60$ iterations. We set  $\epsilon=10^{-4}$ for the Sinkhorn algorithm in ROBOT. Other settings for the hyper-parameters in the baselines follows the default settings of their corresponding papers. EM is initialized by AM.

%\begin{wrapfigure}{r}{7cm}
%    \centering
%    \vspace{-20pt}
%        \includegraphics[width=0.95\linewidth]{figures/synthetic/ransac.pdf}
%%        \vspace{-5pt}
%        \caption{\textit{Pairwise comparisons between RS alone and the combination of RS and ROBOT. The relative error ratio is the ratio of the relative error of RS alone and RS+ROBOT combination.}}
%    \label{fig:ransac}
%    \vspace{-40pt}
%\end{wrapfigure}

\subsection{Multi-Object Tracking}

For the MOT experiments in Section \ref{sec:sec54}, the reported results of \texttt{MOT17} (train) and \texttt{MOT17} (dev) is trained on \texttt{MOT17} (train), and the reported results of \texttt{MOT20} (train) and \texttt{MOT20} (dev) is trained on \texttt{MOT20} (train). Each model is trained for $1$ epoch. We adopt Adam optimizer with learning rate$=10^{-5}$, $\epsilon=10^{-4}$, and $\eta=10^{-3}$.  
To track the birth and death of the tracks, we adapt the inference code of \citet{xu2019deepmot}.

\subsection{The Effect of $\rho_1$ and $\rho_2$}
\begin{figure}[ht]
\centering
\begin{subfigure}{.32\textwidth}
  \centering
  % include first image
  \includegraphics[width=1.2\linewidth]{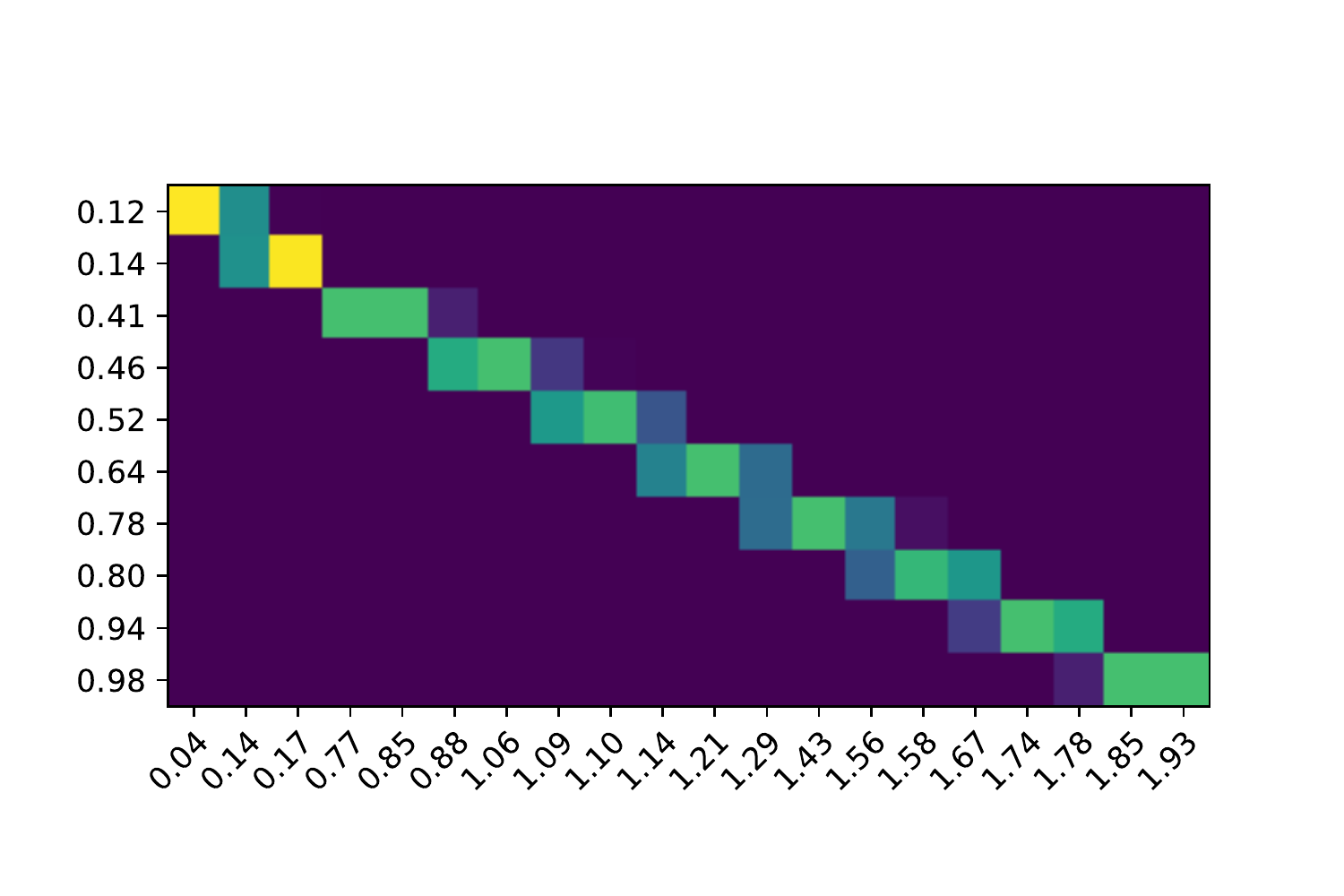}  
  \caption{$\rho_1=0, \rho_2=0$}
  \label{fig:sub-first}
\end{subfigure}
\begin{subfigure}{.32\textwidth}
  \centering
  % include first image
  \includegraphics[width=1.2\linewidth]{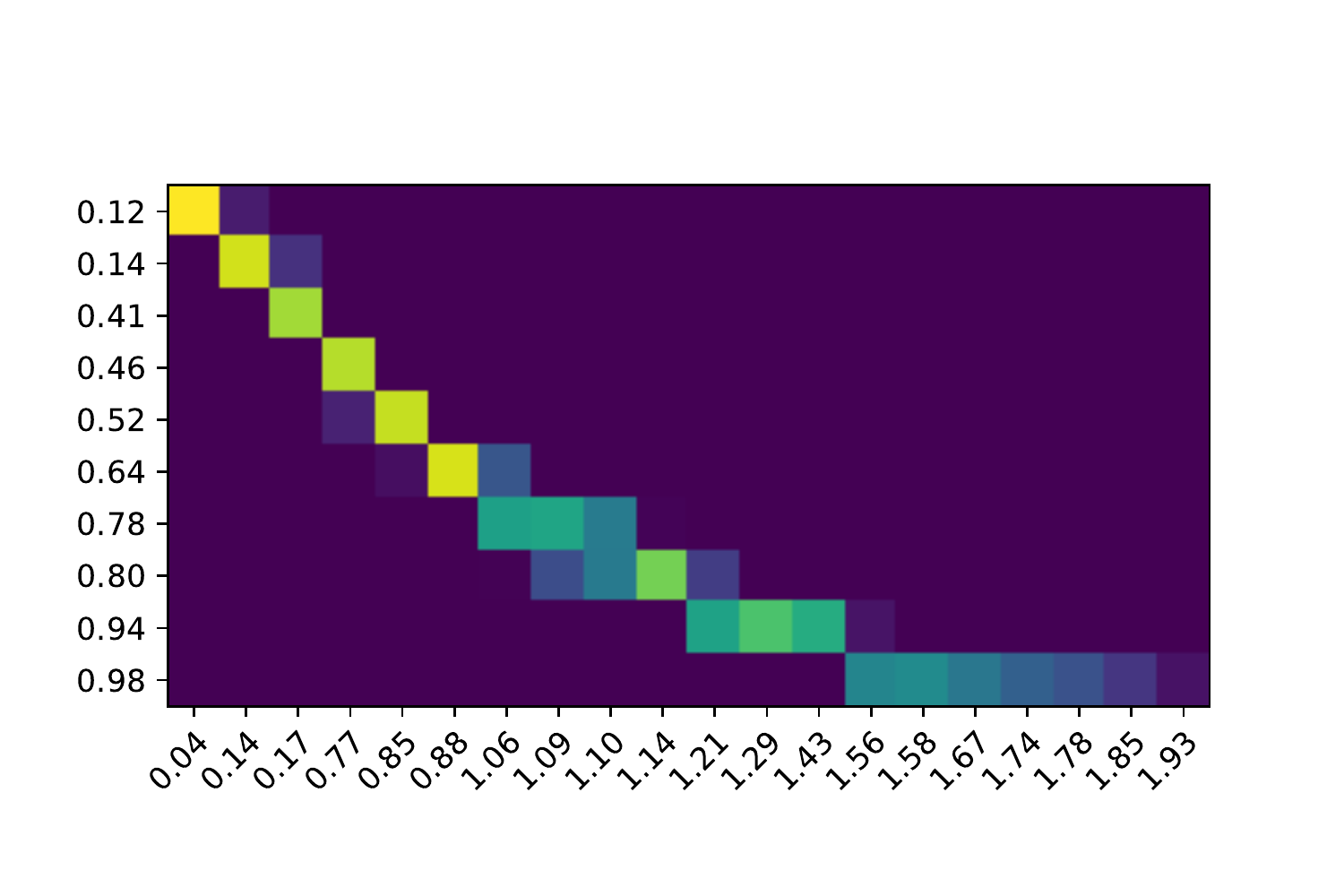}  
  \caption{$\rho_1=0.1, \rho_2=0.1$}
  \label{fig:sub-first}
\end{subfigure}
\begin{subfigure}{.32\textwidth}
  \centering
  % include first image
  \includegraphics[width=1.2\linewidth]{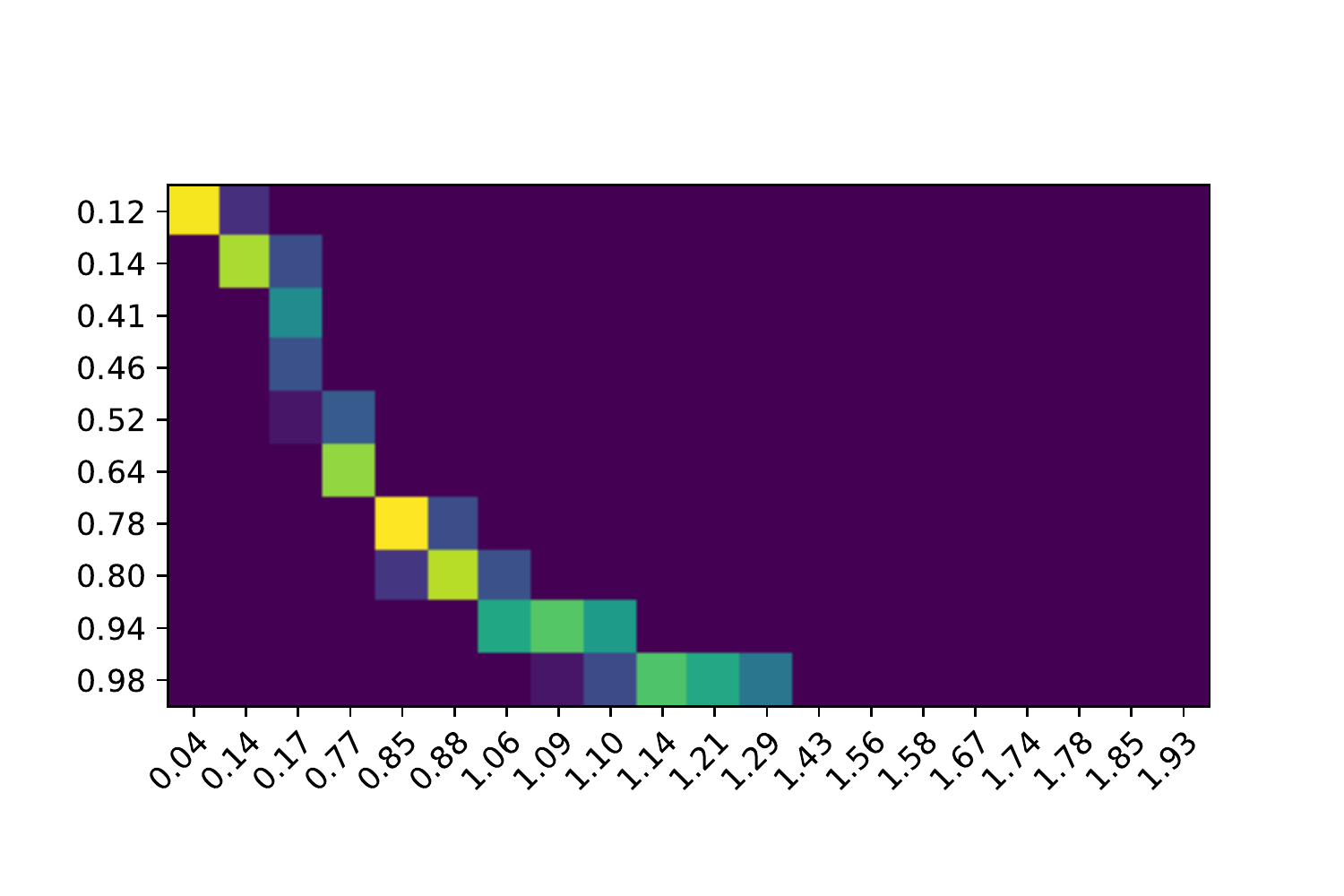}  
  \caption{$\rho_1=0.2, \rho_2=0.2$}
  \label{fig:sub-first}
\end{subfigure}
\caption{Computed $S^*$ for robust optimal transport problem.}
\label{fig:synthetic_rot_forward}
\end{figure}

% \textbf{Robust optimal transport}: 
We visualize $S^*_{\rm r}$ computed from the robust optimal transport problem in Figure \ref{fig:synthetic_rot_forward}. The two input distributions are Unif($0,2$) and Unif($0,1$). We can see that with large enough $\rho_1$ and $\rho_2$, Unif($0,1$) would be aligned with the first half of Unif($0,2$).

% % \section{Variants of OT}
% % \label{sec:var_ot}

% \begin{figure}[htb!]
%     \centering
%     \begin{subfigure}{0.28\linewidth}
%         \includegraphics[height=0.6\linewidth,valign=t]{figures/synthetic_sensing/sensing_sin_d2.pdf}
%         \caption{{ $n=1000,~\rho_{\rm noise}^2=10^{-3}$}}
%     \end{subfigure}
%     \begin{subfigure}{0.28\linewidth}
%         \includegraphics[height=0.6\linewidth,valign=t]{figures/synthetic_sensing/sensing_sin_data.pdf}
%         \caption{$~\rho_{\rm noise}^2=10^{-3},~d_2=3$.}
%     \end{subfigure}
%     % \vfill
%     \begin{subfigure}{0.28\linewidth}
%         \includegraphics[height=0.6\linewidth,valign=t]{figures/synthetic_sensing/sensing_sin_noise.pdf}
%         \caption{$n=1000, ~d_2=3$.}
%     \end{subfigure}
%     \begin{subfigure}{0.1\linewidth}
%         \includegraphics[height=1.5\linewidth,valign=t]{figures/synthetic_sensing/legend.pdf}
        
%     \end{subfigure}
%     \caption{RSS/TSS error with regressor 2.}
%     \label{fig:synthetic_sensing_sin}
% \end{figure}

\subsection{Comparison of Residuals in Linear Regression}

\textbf{Settings}. We generate $n$ data points $\{(y_i, [x_i, z_i])\}_{i=1}^n$, where $x_i \in \mathbb{R}^{d}$ and $z_i \in \mathbb{R}^{e}$.
We first generate $x_i\sim \cN(\bm{0}_{d}, \bm{I}_{d})$, $z_i\sim \cN(\bm{0}_{e}, \bm{I}_{e})$, $w\sim \cN(\bm{0}_{d+e}, \bm{I}_{d+e})$, and $\varepsilon_i \sim \cN(0, \rho^2_{\rm noise})$. Then we compute $y_i = f([x_i, z_i]; w)+\varepsilon_i$.
Next, we randomly permute the order of  $\{z_i\}$ so that we lose the data correspondence. Here, $\cD_1=\{(x_i, y_i)\}$ and $\cD_2=\{z_j\}$ mimic two parts of data collected from two separate platforms. 

We adopt a linear model $f(x;w)=x^\top w$. To evaluate model performance, we use error$=\sum_i (\hat{y}_i-y_i)^2 / \sum_i (y_i-\bar{y})^2$, where $\hat{y}_i$ is the predicted label, and $\bar{y}$ is the mean of $\{y_i\}$.

% {\bf Settings}. 
% We generate $n$ data samples $\{(y_i, [x_1_i, x_2_i])\}_{i=1}^n$, where $x_1_i\in \mathbb{R}^{d_1}$, $x_2_i\in \mathbb{R}^{d_2}$, $y_i\in \mathbb{R}$ as follows. First, we sample $x_1_i, x_2_i$ and $w$ from a standard multivariate  Gaussian distribution. Then we generate $y_i = f(x_i; w)$. Next, we randomly permute the order of  $\{x_2_i\}$. We randomly split $\{(y_i,x_1_i)\}_{i=1}^n$ into the training data and the test data. 
% $y_i$ as $x_iw$ plus a Gaussian noise with zero mean and variance $\rho^2_{\rm noise}$. 

% \textbf{Evaluation}. 

% which is a nonlinear function of $x$. Likewise, we randomly shuffle part of the input data and randomly sample out some data points to be the held-out validation set.
\vspace{10pt}
\noindent \textbf{Baselines}. We use Oracle, LS, Stochastic-EM as the baselines. Notice that without a proper initialization, Stochastic-EM performs well in partially permuted cases, but not in fully shuffled cases. For better visualization, we only include this baseline in one experiment. Furthermore, we adopt two new baselines: Sliced-GW \citep{vayer2019sliced} and Sinkhorn-GW \citep{xu2019gromov}, which can be used to align distributions and points sets.

\begin{figure}[h!]
\vskip -0.1in
    \centering
    \begin{subfigure}{0.4\linewidth}
        \includegraphics[width=\linewidth]{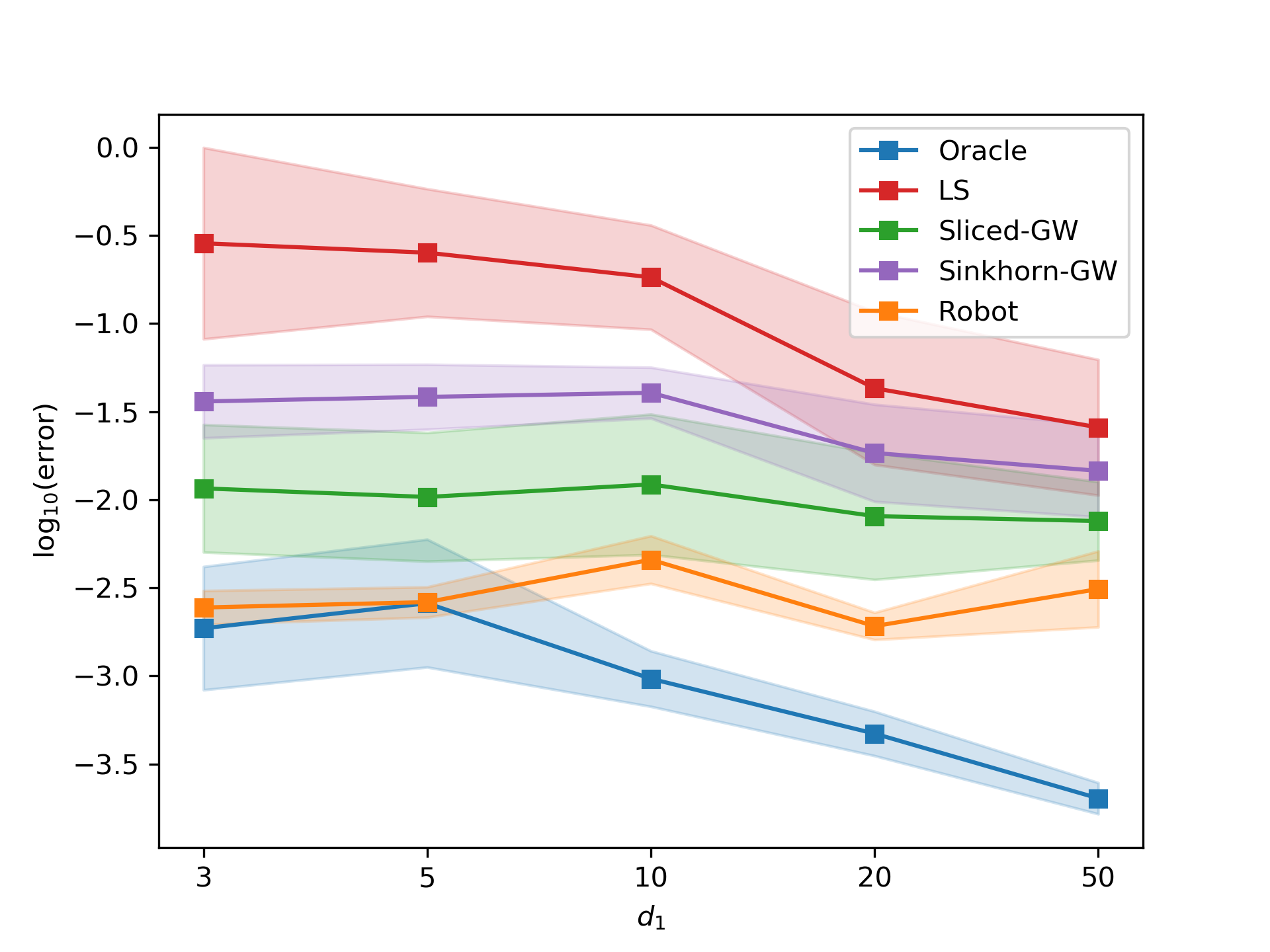}
    \end{subfigure}
    \begin{subfigure}{0.4\linewidth}
        \includegraphics[width=\linewidth]{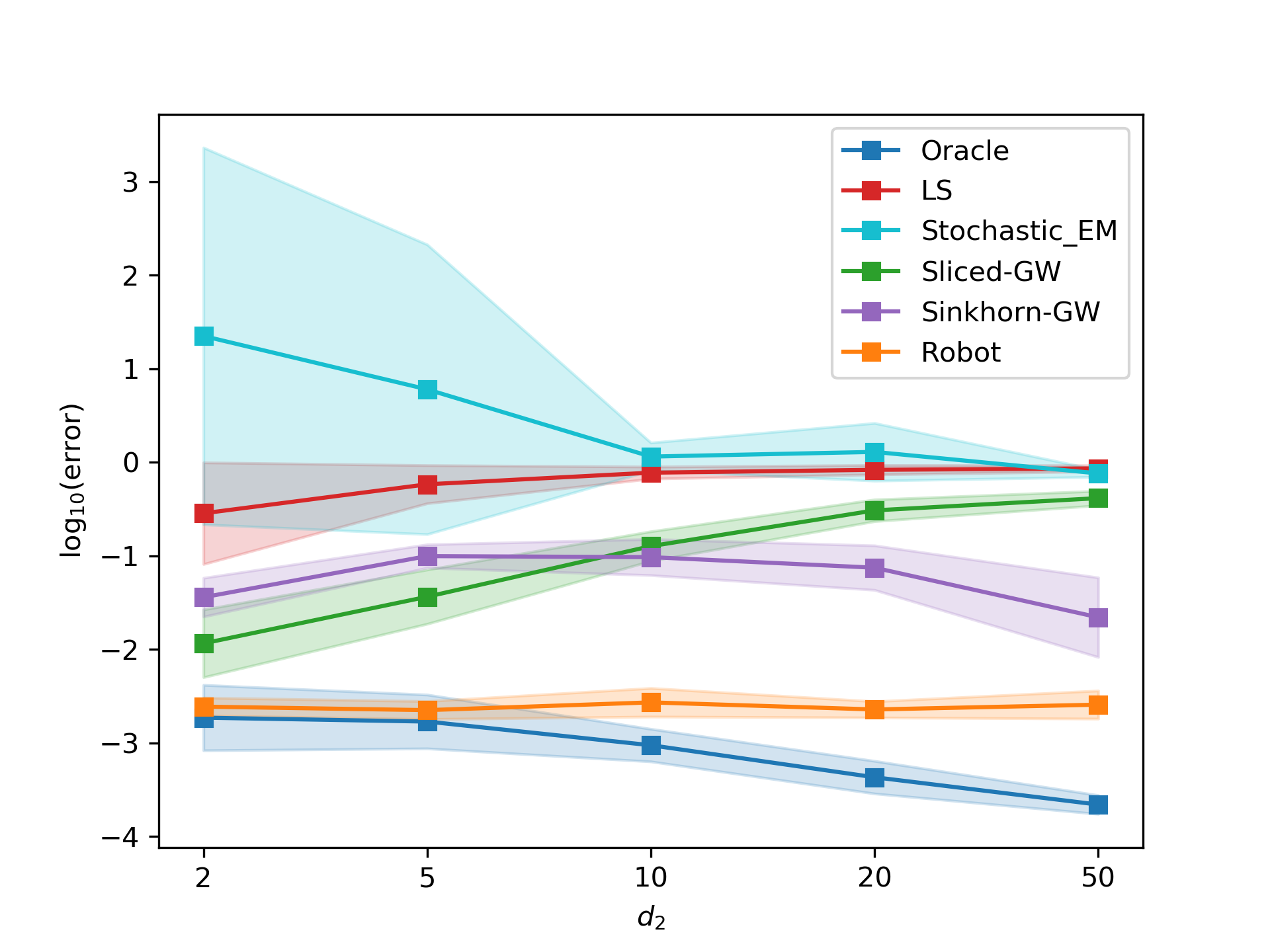}
    \end{subfigure}
    \begin{subfigure}{0.4\linewidth}
        \includegraphics[width=\linewidth]{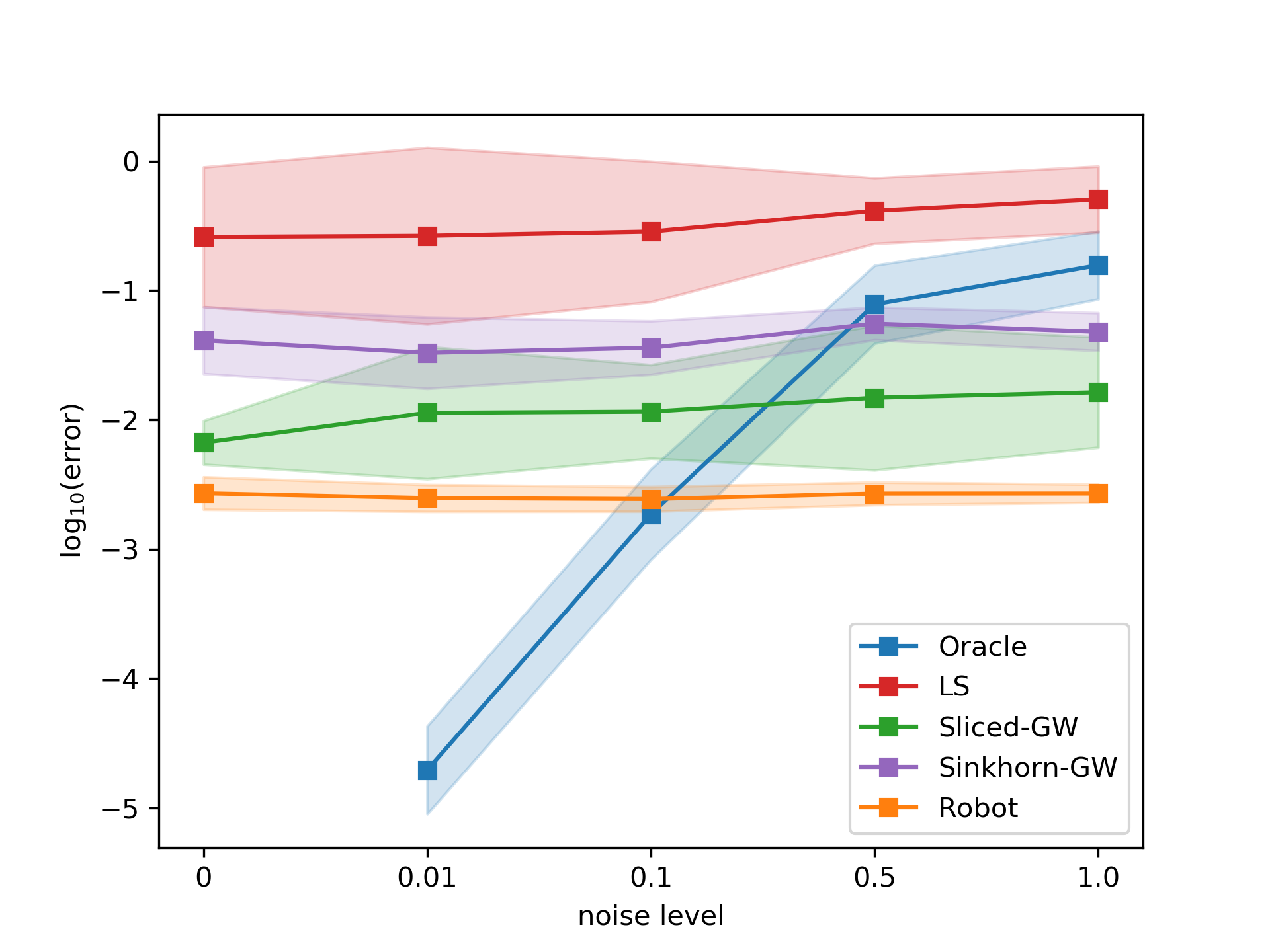}
    \end{subfigure}
    \begin{subfigure}{0.4\linewidth}
        \includegraphics[width=\linewidth]{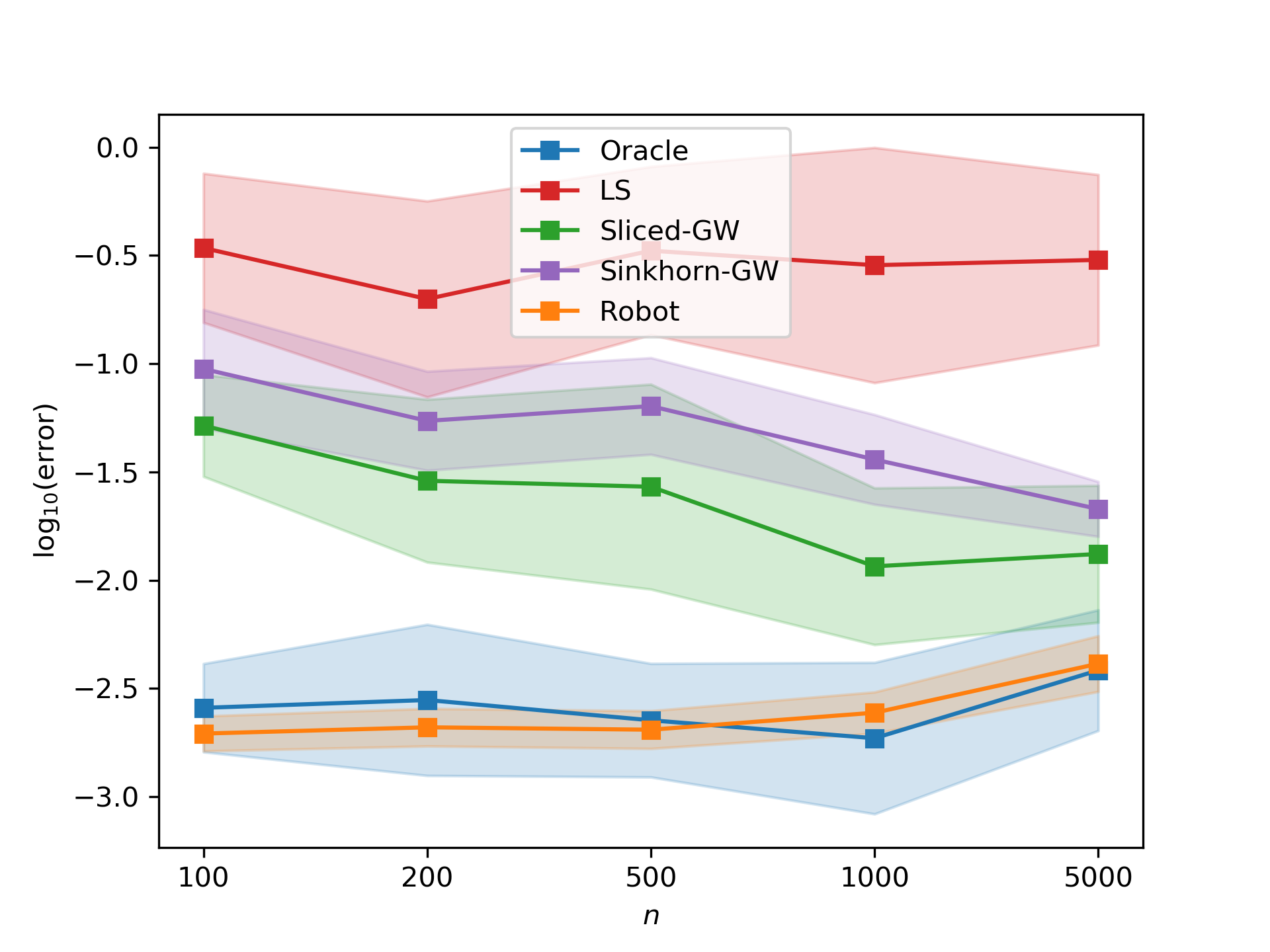}
    \end{subfigure}
    \caption{Linear regression. We use $n=1000$, $d=2$, $e=3$, $\rho_{\rm noise}^2=0.1$ as defaults.}
    \label{fig:synthetic_linear}
\end{figure}

\vspace{10pt}
\noindent  \textbf{Results}. We visualize the fitting error of regression models in Figure \ref{fig:synthetic_linear}. We can see that ROBOT outperforms all the baselines except Oracle. Also, our model can beat the Oracle model when the dimension is low or when the noise is large.
% Here error is calculated as $\textrm{RSS}/\textrm{TSS}=1-R^2$. 
% Yujia: I move the following part to discussion
%In some previous papers, the fitting error is defined by relative error of the inferenced parameters, i.e., $\| w-w^* \| / \| w^* \|$, where $w^*$ is the ground truth. However, in the setting of shuffled regression, multiple regression parameters $w$ can generate satisfactory fitting results, depending on the inferenced permutations. Also, since we are targeting for a reliable predictive model, using goodness-of-fit statistics, i.e., $R^2$, to evaluate our model is more reasonable.

% It is worth mentioning that we notice that \textit{Stochastic-EM} performs well in partially permuted cases, which is not applicable to our experiments
% poorly on low dimension data, i.e., . For better visualization, we only include this baseline in one experiment.
% In the experiments, we have two platforms. In the first platform, we have data points $x$ with dimension $d_1$ and their corresponding labels $y$. While in the second platform, we only have the data points $x_2$ without label information.

% (A set of good parameters for robust matching: $n=10^3, d_1=3, d_2=2, \rho^2_{\rm noise}=0.1$, train iter=200, bs=100, lr=5e-7, $\epsilon=10^{-4}$, $\rho_1=0.1, \rho_2=0.1, \eta=10^{-3}$. Can get better result than oracle.)

\end{document}